%% file: main.tex
\pdfoutput=1

\documentclass[11pt]{article}

\usepackage{ACL2023}

\usepackage{times}
\usepackage{latexsym}

\usepackage[T1]{fontenc}

\usepackage[utf8]{inputenc}

\usepackage{microtype}

\usepackage{inconsolata}

\usepackage{amsmath}

\usepackage[noabbrev,capitalize]{cleveref}
\usepackage{multirow}
\usepackage{graphicx}
\usepackage{colortbl}
\usepackage[normalem]{ulem}

\usepackage{nicematrix}
\usepackage{tikz}
\usetikzlibrary{patterns,shadings}

\usepackage{arydshln}
\usepackage[export]{adjustbox}
\def\hideXXX#1{}

\def\yesXXXifaccepted#1{#1}%

\def\hidePV#1{}

\def\footurl#1{\footnote{\url{#1}}}

\def\inparcite#1{\citealp{#1}} 
\def\furl#1{\footnote{\url{#1}}}

\newlength{\Oldarrayrulewidth}

\usepackage{array}
\newcolumntype{?}{!{\vrule width 1.5pt}}

\usepackage{lipsum,xcolor}
\newcommand\dunderline[3][-1pt]{{%
  \sbox0{#3}%
  \ooalign{\copy0\cr\rule[\dimexpr#1-#2\relax]{\wd0}{#2}}}}

\title{Robustness of Multi-Source MT to Transcription Errors}

\include{authors.tex}

\begin{document}
\maketitle
\begin{abstract}

Automatic speech translation is sensitive to speech recognition errors, but in a multilingual scenario, the same content may be available in various languages via simultaneous interpreting, dubbing or subtitling. In this paper, we hypothesize that leveraging multiple sources will improve translation quality if the sources complement one another in terms of correct information they contain. To this end, we first show that on a 10-hour ESIC corpus, the ASR errors in the original English speech and its simultaneous interpreting into German and Czech are mutually independent. We then use two sources, English and German, in a multi-source setting for translation into Czech to establish its robustness to ASR errors. Furthermore, we observe this robustness when translating both noisy sources together in a simultaneous translation setting. Our results show that multi-source neural machine translation has the potential to be useful in a real-time simultaneous translation setting, thereby motivating further investigation in this area.
\end{abstract}

\section{Introduction}


Speech translation (ST) suffers from automatic speech recognition (ASR) errors, especially in challenging conditions such as non-native language speakers, background noise, named entities and specialized vocabulary usage \cite{antrecorp:2019,gaido-etal-2021-moby,gaido-etal-2022-talking,anastasopoulos-etal-2022-findings}. ASR errors negatively impact translation quality, via the compounding of speech recognition and translation errors \cite{ruiz17_interspeech,sperber-paulik-2020-speech}, thereby limiting the application of automatic speech translation in realistic settings. Fortunately, there are multilingual settings where a source is simultaneously or consecutively interpreted into multiple languages. Many documents are also dubbed or subtitled in offline mode. This simultaneous interpretation takes place either via human interpreters, in the form of dubbing or subtitling.
In a situation where the same sentence is available in multiple languages, multi-source machine translation (MT) significantly improves translation quality, especially when the two sources used separately do not yield high quality translations 
\cite{RajDabre2018,zoph-knight-2016-multi,nishimura-etal-2018-multisource}.

Although not yet clearly verified, multi-source MT could be useful in settings where the sources complement each other. In other words, challenges in translation posed by using each source should be independent of one another. Given that ASR is noisy and that multiple sources can help overcome limitations of individual sources, this paper asks the following question: ``Can multi-source MT be leveraged for speech translation in a multilingual setting where an original source transcription and its simultaneously interpreted transcription are available?'' We decompose this question into three parts where we hypothesize that (a) ASR errors in the original and interpreted transcripts are independent, which makes them complementary, (b) multi-source MT is robust to transcription errors present in individual sources, and (c) the robustness of multi-source MT continues to hold in a simultaneous translation setting. We address each question in Sections~\ref{sec:asr-dep}, \ref{sec:multisrc-st}, and \ref{sec:simul-mtmain}, respectively.

To prove our hypotheses, firstly, we verify on the Europarl Simultaneous Interpreting Corpus (ESIC, \inparcite{machacek21_interspeech}) that original speech ASR and interpreted speech ASR are indeed complementary in terms of errors. Secondly, we simulate transcription errors in a full sentence multi-source MT setting for English and German to Czech translation. We clearly show that when both sources are noisy, using them together leads to significant improvements when compared to using them individually, in contrast to drops in quality when at least one of the sources is clean.  For example, on ESIC test set with 15\% WER noise in English source and 10\% WER noise in German source, multi-sourcing performs 0.9 BLEU score higher than English source.
Finally, we use both sources in a simultaneous translation setting and show that multi-source MT continues to be robust to transcription errors.


Our findings show that multi-source MT has strong potential in a simultaneous translation setting where multiple sources are available via ASR or interpreted ASR. 
We note that our current analysis is limited to the case where the multiple sources are aligned and hence available at the same time. This is a setting of, e.g., dubbed and subtitled videos where we would want to consider additional target languages. In simultaneous settings where one source is available with a delay, the synchronization of the sources would be a considerable problem, which we leave for future research.

\section{Related Work}
This paper mainly focuses on ASR errors, multilingual multi-source translation and simultaneous translation.

\paragraph{ASR errors} often propagate to MT in cascaded ST systems and in a real time translation setting where ASR systems are used. It is a major issue that affects translation quality. \citet{martucci21_interspeech} propose a method to tune the MT on the training data with artificial noise that mimics ASR errors, via a unigram ``lexical noise model'' learned on automatic--gold transcript pairs.
Other authors propose similar methods for training \cite{sperber-etal-2017-toward,di-gangi-etal-2019-robust,xue-etal-2020-robust,serai-2022}. However, in this work, rather than mimicking ASR noise during training, 
we complement a noisy source with another whose ASR errors are provably independent of the first. Specifically, we use the lexical noising of \citet{martucci21_interspeech} to simulate noise in multiple source languages and show the robustness of MT models, especially in a multi-source setting. 


\paragraph{Multilingualism} \cite{10.1145/3406095} has been shown to improve translation quality in a variety of situations. In particular, multi-source machine translation has high potential for improving translation quality, but has been relatively underexplored. In the context of multi-source text-to-text translation, \citet{zoph-knight-2016-multi} and \citet{RajDabre2018} showed that leveraging the same sentence in different languages improves translation quality as the two sources are expected to complement hard to translate phenomena in the other source. Although this approach requires multi-parallel sentence-aligned data, the missing sources can be obtained by MT \cite{nishimura-etal-2018-multisource} or via simultaneous interpretation. Rather than training a multi-source model, \citet{firat-etal-2016-zero} propose the ``late averaging'' which needs multilingual models trained on pairwise bilingual data, which we also focus on when evaluating multi-source models. Late averaging is akin to ensembling via logits averaging, but with source sentences in different languages. These works do not consider transcription errors which are ubiquitous in speech translation, an aspect this paper focuses on. 

\paragraph{Multi-sourcing} in simultaneous translation settings has not been extensively explored. \citet{DBLP:journals/corr/abs-2104-07410} have explored simultaneous multi-pivot translation where a source is translated into a target language via multiple pivot languages, where the pivot languages are translated using multi-source translation. Unlike them, we consider only one pivot language which is interpreted from the source and then use it together with the source to show that the translation quality into the target language improves. Additionally, they do not consider the effect of transcription noise on translation, which we do. Simultaneous translation approaches such as wait-$k$ \cite{ma-etal-2019-stacl} and Local Agreement (LA-$n$, \inparcite{polak-etal-2022-cuni}) are commonly used, and we use the latter for our experiment.

\yesXXXifaccepted{%
An alternative multi-sourcing approach is to select and use only one source. \citet{machacek21_interspeech} provide an analysis of using either the original, or its simultaneously interpreted equivalent as a source for simultaneous ST. \yesXXXifaccepted{Interpreting is delayed, but shorter and simpler than translationese. Interpreters also segment their speech to sentences differently than the original speakers, so it is not easy to align segments. In any case, selecting sources will involve additional effort and thus we consider using multiple sources together to be a more effective approach.} In this regard, multiple language sources, both as text and speech streams, could be used in ASR \cite{paulik-etal-ste-asr-2005,Soky2022LeveragingST} as well as in pre-neural MT and ST \cite{och-ney-2001-statistical,paulik-waibel-2008,khadivi-ney-2008}. \citet{miranda-etal-punctuation-2013} use them for punctuation restoration.
\citet{kocmi-book-2021} provide a broad analysis of benefits of multilingual MT.}
\section{Parallel Source-Interpreted ASRs are Independent}
\label{sec:asr-dep}

We assume a multi-source setting with the original speech and its simultaneously interpreted equivalent as the two sources will improve robustness to ASR errors if the errors in the two source streams complement each other. This is not obvious because, on the one hand, the ASRs work independently where they are deployed for different languages, trained on different data, and the processing is fully independent. On the other hand, the content of the speeches is identical.
Interpreters' speech pacing also depends on the original speaker, and it may influence the quality of both ASRs the same way. Therefore, in this section, we analyze the dependency of ASR errors in the source and interpreter, on 10-hour ESIC corpus \cite{machacek21_interspeech} to prove that the ASR errors are indeed independent.

\paragraph{Methodology}
First, we processed ASR for English original speakers and interpreters into Czech and German.
For English, we used the low-latency neural ASR by \citet{nguyen21c_interspeech}. 
For German, we used an older hybrid HMM-DNN model trained using the Janus Recognition Toolkit, which features a single-pass decoder \cite{kit-german-lectures}.
For Czech, we used Kaldi \cite{Povey11thekaldi} HMM-DNN model trained on Czech Parliament data \cite{FIXkratochvil-etal-2020-large}. \cref{tab:esic-asrs-wer} summarizes the transcription quality on ESIC showing that the quality is low, but to the best of our knowledge it is the best one available for this domain.

\begin{table}[]
    \centering
    \footnotesize
    \resizebox{\columnwidth}{!}{%
    \begin{tabular}{l|c|c|c}
    \textbf{subset} & \textbf{Cs interp.} & \textbf{De interp.} & \textbf{En original}\\
    \hline
    \textbf{dev}  & 14.84 & 25.14 & 13.63 \\
    \textbf{test} & 14.04 & 23.79 & 14.71 \\
    \end{tabular}}
    \caption{Transcription WER on ESIC. There are 191 and 179 documents in dev and test subsets. The scores are weighted by number of words in gold transcripts.}
    \label{tab:esic-asrs-wer}
\end{table}


We then re-used the word alignments of gold transcripts between the original and interpretation as described in \citet{machacek21_interspeech}. 
38\% of tokens were aligned between English and Czech interpretations, and 40\% between English and German, see \cref{tab:aligned-tokens}.
It may be caused by the characteristics of the language pair (e.g.\ compound words in German vs multi-word expressions in English), features of interpreting (non-verbatim translation, shortening) and by errors in automatic alignment. We only analyzed the aligned tokens further. Since there are many tokens left in two 5-hour subsets of the corpus, we consider further analysis as valid.

\begin{table}[]
    \centering
    \resizebox{\columnwidth}{!}{%
    \begin{tabular}{c@{~}|r@{~}|r@{~}r@{~}|r@{~}r}
     & \textbf{En tokens} & \multicolumn{2}{@{~}c@{~}|}{\textbf{En-Cs aligned}} & \multicolumn{2}{@{~}c}{\textbf{En-De aligned}} \\
    \hline
    \textbf{dev}      & 44,494 & 16,962 & (38.12\%) & 17,809 & (40.03\%) \\
    \textbf{test}     & 46,151 & 17,623 & (38.19\%) & 19,280 & (41.78\%) \\
    \end{tabular}}
    \caption{Number and percentage of aligned tokens in gold transcripts between the original source (English [En]) and its interpretations (German [De] and Czech [Cs]).}
    \label{tab:aligned-tokens}
\end{table}

Finally, we aligned gold and automatic transcripts using Le\-venshtein edit distance.\furl{https://pypi.org/project/edlib/} We classified each token in the ASR transcript as transcribed correctly or not, both for source and interpretations. 

\paragraph{Results}
We made a contingency table (\cref{tab:confusion-matrix}) and ran a $\chi^2$ test \cite{Pearson1900} of statistical independence. The results show that the \textbf{parallel source and interpretation ASRs make errors independently} of each other with $p<0.01$, for both pairs, English-Czech and English-German, for both dev and test subsets. 

\begin{table}[]
    \centering
    \footnotesize
    \begin{tabular}{c@{~}c|r@{~}r|r@{~}rrrr}
 \multicolumn{2}{c|}{\multirow{2}{*}{\textbf{En orig.}}}   & \multicolumn{2}{c|}{\textbf{Cs int.}} &  \multicolumn{2}{c}{\textbf{De int.}} \\
   &       & \textbf{corr.}  & \textbf{incorr.} & \textbf{corr.} & \textbf{incorr.} \\
         \hline
\multirow{2}{*}{\rotatebox[origin=c]{0}{\textbf{dev}}} & \textbf{corr.}       & 13815 & 1497 & 7192 & 1561 \\
 &  \textbf{incorr.}  & 1228  & 422 & 633 & 307 \\
 
 \hline
 
\multirow{2}{*}{\textbf{test}} & \textbf{corr.} & 14204 & 1655 & 7895 & 1638 \\
                      & \textbf{incorr.} & 1344 & 420 & 692 & 336 \\
    \end{tabular}
    
    \caption{Contingency table of correctly and incorrectly recognized aligned tokens in English source (in rows) and interpretation into Czech and German (in columns), in dev and test subset of ESIC corpus. According to the $\chi^2$ test of statistical independence, in all 4 cases, the parallel recognition is independent with $p<0.01$.}
    \label{tab:confusion-matrix}
\end{table}

We manually assessed the severity of the ASR errors and realized that most errors are only in spelling and fluency, and not in adequacy. We therefore conclude that our finding of independence of parallel ASRs may be valid only for ASRs of comparable quality to ours.










\section{Multi-Source Speech Translation}
\label{sec:multisrc-st}

Having established that ASR errors are independent, we now analyze whether multi-source neural machine translation (NMT) is robust to noisy sources.
We focus on NMT for individual sentences, with gold sentence alignment of the
sources and reference. 
It is a less realistic use-case than translating long speech documents
without any sentence segmentation and alignment of the sources, but proving the robustness of multi-sourcing in this setting paves the way for its application in long speech document translation.


\paragraph{Data}

For training, we use data from OPUS \cite{tiedemann-nygaard-2004-opus}, aiming at a multi-way
model with English and German on the source side and Czech as a target. We
download all the data from OPUS, remove all sentences from IWSLT, WMT, ESIC and other test sets, filter them by language identification, and then process with dual cross-entropy scoring
\cite{junczys-dowmunt-2018-dual} using the bilingual NMT models from \citet{tiedemann-thottingal-2020-opus}. We select the top 30 million sentences for each language pair as training data, to prevent overfitting for either. \yesXXXifaccepted{It is also
near the threshold that \citet{chen-etal-2021-university} showed as optimal.}

For NMT validation and evaluation, we use the ``revised transcript and
translations''
from ESIC \cite{machacek21_interspeech}. These are the texts that were originally uttered in the European Parliament, transcribed, revised and normalized for reading and publication on the website, and then translated. They are analogous, but not identical, to the gold transcripts of the original and interpretations that we used in \cref{sec:asr-dep}.
In addition to the version published
by \citet{machacek21_interspeech}, we properly align the sentences in all
the three languages. \yesXXXifaccepted{Two documents were removed because they missed
German translation. The corpus is of comparable size to a usual MT test
set.} See size statistics in \cref{tab:esic-size}.

\begin{table}[]
    \centering
    \footnotesize
    \begin{tabular}{@{~}l|r@{~}r|rrr@{~}}
         & sent. & doc. & En words & De w. & Cs w. \\
         \hline
 dev        & 2002 & 179 & 44866 & 43323 & 38347 \\
 test       & 1963 & 189 & 44273 & 42491 & 37695 \\
    \end{tabular}
    \caption{Size statistics of tri-parallel sentence-aligned ``revised translations'' of ESIC \cite{machacek21_interspeech}. English is original, German and Czech are translations.
    }
    \label{tab:esic-size}
\end{table}

For a contrastive evaluation, we use Newstest11 \cite{callison-burch-etal-2011-findings}. It contains 3003 sentences in 5 languages: English, German, Czech, French and Spanish, the same amount in each. Newstest11 has references that were translated directly, not through an intermediate language. We also use three additional Czech references of Newstest11 that were translated from German \cite{news11}.

\paragraph{Multi-Sourcing}

%
We convert Marian models to PyTorch to be used with the Hugging Face Transformers \cite{wolf-etal-2020-transformers} library, in which we implement late and early averaging.
For both single- and multi-sourcing, we use greedy decoding because beam search support is not implemented with multi-source. 

\paragraph{Training details}
We train a multi-way NMT model using Marian \cite{junczys-dowmunt-etal-2018-marian} with English and German as sources, with language identification tokens, and Czech as the target. We use two separate SentencePiece \cite{kudo-richardson-2018-sentencepiece} vocabularies, both sizes of 16\,000. The source vocabulary is joint for German and English, and the target is only for Czech. The model is a Transformer Base (6 layers, 512 embedding size, 8 self-attention heads, 2048 filter size) trained on 8  Quadro P5000 GPUs with 16 GB memory
for 17 days, until convergence. 


\paragraph{Checkpoint selection}
We validate all checkpoints (every 1000 training steps, 15 minutes) on two single sources (English and German) and two multi-sourcing options: early averaging, and late averaging of a single checkpoint with two sources. Furthermore, after the training has ended, we selected top 10 checkpoints that reached the highest BLEU scores for English and German single-source on the ESIC dev set. We evaluated all pairs of the top performing checkpoints in late averaging multi-sourcing setup.
The top performing model from all validation and grid search options was selected as a final model. It is late averaging with a pair of distinct checkpoints. We also use these two checkpoints for single source evaluation.


\def\chrf{chrF2}
\def\CometEn{En COMET}
\def\CometDe{De COMET}

\paragraph{Evaluation Metrics}
We estimate translation quality by BLEU \cite{papineni-etal-2002-bleu} and \chrf{} \cite{popovic-2016-chrf} calculated by sacreBLEU\footnote{Metric signatures:    \tiny{BLEU|nrefs:1|case:mixed|eff:no|\-tok:13a|\-smooth:exp|
version:2.2.1}, \tiny{chrF2|nrefs:1|case:mixed|eff:yes|nc:6|nw:0|space:no|version:2.2.1}} \cite{post-2018-call}. 
We also report the current state-of-the-art metric COMET\footnote{\texttt{wmt20-comet-da} model} \cite{rei-etal-2020-unbabels} that achieves the highest correlation with direct assessment as a kind of human judgements \cite{mathur-etal-2020-results}. However, COMET requires one source on the input and is not suitable for multi-source. Therefore, we report it twice (En/De COMET) with two single sources. Note that \CometEn{} scores assume English as source and Czech as target. Since ESIC is tri-parallel, even if the translation is obtained using German or English and German multi-source, we only use the English source as the input to the COMET model. \CometDe{} scores are computed similarly.

\paragraph{Results with clean inputs}

\Cref{tab:results-clean} shows the results of multi-sourcing with clean inputs, without any speech recognition noise.
One would be tempted to conclude that the translation from English is of a higher quality than the translation from German (e.g. 33 vs. 26 BLEU on ESIC dev set), but such a claim is risky. The metrics  measure the match of the candidate translation with the reference sentence (and, in case of COMET, also with the source), and it is conceivable that the English served as the source for the human reference translation. The Czech reference thus may very well exhibit more traits of the English source than of the German source. While the \chrf{} scores agree with BLEU, COMET scores seem to indicate that multi-sourcing is as good as, if not better than, using a single source. Since COMET is known to correlate with human judgements better than BLEU \cite{mathur-etal-2020-results} our results show that multi-sourcing is indeed a viable solution.

To further shed light on the impact of the source used for creating references, we evaluated the models with Newstest11 and computed the scores with three additional references that were translated only from German. The German single source achieves much higher BLEU than the English source (32.23 vs 16.62 BLEU), with multi-sourcing in between (22.47 BLEU). Similar trends are observed in \chrf{} and COMET scores. This is the opposite of ESIC scores, where the reference was obtained from English. It shows that the traits of the source language such as word order, structure of clauses and terms are remarkable in automatic metrics when the reference is constructed from that source, but these effects may be negligible in human evaluation. \Cref{sec:ref-src} contains more details.

Finally, we consider a ``balanced'' scenario where an equal number of references comes from both sources and this shows similar scores for both single sources (23.40 vs 22.85 BLEU) with multi-sourcing outperforming them by 0.6 and 1.1 BLEU. We therefore conclude that our multi-source model should be well-prepared for content originating in any of the source languages, but the automatic evaluation metrics may not always capture this. Moving forward, we only use BLEU for simplicity.



\def\pms{$\pm$}
\def\avgstddev#1#2{$\underset{\pm\text{#2}}{\text{#1}}$}
\def\boldavgstddev#1#2{$\underset{\pm\text{#2}}{\textbf{#1}}$}

\def\settworows#1#2{\multirow{4}{*}{\shortstack[c]{\textbf{#1}\\#2}}}
\def\settworowsBig#1#2{\multirow{6}{*}{\shortstack[c]{\textbf{#1}\\#2}}}
\def\sign#1{$^*$#1}
\def\nosign#1{$^\times$#1}
\def\to{$\rightarrow$}
\def\balanced{$\{$De,En,Fr,Es$\}$\to{}Cs}
\def\balancedcs{\balanced{},\\ Cs}
\def\enrefsrc{En\to{}Cs}
\def\refsrclan{ref. translation:}
\def\threetimesde{
    3$\times\{$De\to{}Cs$\}$%
    }
\begin{table}[t]
\footnotesize
    \centering
    \begin{tabular}{@{~}c@{~}|c@{~}||@{~}r@{~}r@{~}r@{~}}
\bf Set        & \multirow{2}{*}{\bf Metric}   & \multicolumn{3}{c}{\bf Model} \\
\refsrclan{} & &    En & De & De+En \\
  \hline
  \hline
\settworows{ESIC dev}{\enrefsrc{}} & BLEU      &
\sign{\textbf{33.31}} & 26.13 & \sign{31.90}  \\
                   & \chrf{}      &  \sign{\textbf{60.17}} & 54.00 & \sign{58.59} \\
                   &\CometEn{}& \nosign{\bf 0.920} & 0.860 & \sign{0.919}   \\
                   &\CometDe{}& \nosign{1.007} & 0.994 & \sign{\textbf{1.022}}    \\
                   \hline
\settworows{ESIC test}{\enrefsrc{}} & BLEU     & \sign{\textbf{33.63}} & 27.99 & \sign{32.57} \\
 & \chrf{}  & \sign{\textbf{59.58}} & 54.75 & \sign{58.63} \\
 &\CometEn{}& \sign{0.906} & 0.871 & \nosign{\textbf{0.912}} \\
 &\CometDe{}& 0.994 & \nosign{1.006} & \sign{\textbf{1.018}} \\
  \hline
  \hline
\settworowsBig{news11}{\threetimesde{}} & BLEU  &\avgstddev{16.62}{0.29} & \boldavgstddev{32.23}{0.53} & \avgstddev{22.47}{0.44}  \\
      & \chrf{}  & \avgstddev{44.84}{0.18} & \boldavgstddev{58.81}{0.38} & \avgstddev{49.72}{0.27} \\
      & \CometEn{} & \avgstddev{0.528}{0.002} & \boldavgstddev{0.823}{0.002} & \avgstddev{0.652}{0.003} \\
      & \CometDe{} & \avgstddev{0.600}{0.002}  & \boldavgstddev{0.967}{0.001} & \avgstddev{0.757}{0.003} \\
 \hline
      \settworows{news11}{\balancedcs{}} & BLEU &  \sign{23.40} & 22.85 & \sign{\textbf{23.96}}  \\  
 & \chrf{} & \nosign{\textbf{51.00}} & 50.27 & \sign{50.83} \\
  &\CometEn{}& 0.627 & \sign{\textbf{0.674}} & \sign{0.659} \\
  &\CometDe{}& 0.700 & \sign{\textbf{0.832}} & \sign{0.766} \\
    \end{tabular}
    \caption{Evaluation scores with clean inputs (no ASR noise), machine-translated into Czech with single-sourcing English (En) or German (De), or multi-sourcing (De+En), on ESIC and Newstest11 (news11). Newstest is evaluated on a balanced reference that has origin in 5 languages (\balanced{} translations and Cs original; 600 sentences each), and 3-times with additional references that were translated from German (``\threetimesde{}''). We report avg\pms{}stddev for them. 
    ``\CometEn{}'' and ``\CometDe{}'' are run with English and German source, respectively. Maximum scores are in bold.
    The symbol \sign{} means that there is statistically significant difference $(p<0.05)$ from all the lower scores in the same row, \nosign{} means no significance ($t$-test for COMET, paired bootstrap resampling for BLEU and \chrf{}).
    }
    \label{tab:results-clean}
\end{table}

\begin{table*}[ht]
\input{outertable.tex}

\caption{
BLEU (avg\plusm{}stddev) with transcription noise on ESIC dev set whose reference translations was English and on Newstest11 with balanced reference source language. Green-backgrounded area is where the \colorbox{colorone}{English} single-source outperforms \colorbox{colortwo}{German} single-source. \munderline{Black underlined} numbers indicate the area where multi-sourcing achieves higher score than both single-sourcing options. In \textbf{bold} is near maximum gap from single-source, more than 2.1 BLEU. \textcolor{colorlow}{Red-colored} numbers are where at least one single-source scores higher.
}
    \label{tab:wer-noise}
\end{table*}

\begin{table}[]
    \centering
    \resizebox{\columnwidth}{!}{%
    \begin{tabular}{@{~}l@{~}||r@{~}|r@{~}|r@{~}} 
\textbf{WER}    & \textbf{En} & \textbf{De} & \textbf{En+De} \\
    \hline
15\% En, 10\% De & 23.58\pms{}0.16 & 23.23\pms{}0.05 & \textbf{26.50\pms{}0.27} \\
    \end{tabular}}
    \caption{ESIC test multi-sourcing vs single-sourcing BLEU scores on the artificial WER noise level where multi-sourcing achieved the largest improvement.}
    \label{tab:test-set-results}
\end{table}

\subsection{Modeling Transcription Noise}
\label{sec:asr-noise}

Although multi-sourcing English and German is not very beneficial when both sources are clean, we hypothesize that it could show benefits with noisy sources. Averaging two noisy sources can lead to cancelling the noise.
Since ESIC contains tri-parallel sentence-aligned translations as texts and not speech, 
and since we want to evaluate different levels of ASR noise, and we do not have many ASRs, we generate the ASR errors artificially.


\paragraph{Custom WER noise model}
We adopt the lexical noise model by \citet{martucci21_interspeech} and modify it to create outputs with arbitrary WER. The lexical noise model modifies the source by applying insertion, deletion, substitution, or copy operations on each word with probabilities $p_I, p_D,$ and $p_S$, respectively. The probabilities are learned from the ASR and gold transcript pairs. It thus may learn to shuffle homonyms such as ``eight'' and ``ate''. 

In the original lexical noise model by \citet{martucci21_interspeech}, the target WER is bound to the performance of the given ASR system on which it is trained, and can not be changed. WER is defined as the number of incorrect words in the ASR transcript divided by the number of correct words in the gold transcript. The errors are either insertions, deletions, or substitutions. In the lexical noise model, insertion is applied independently on the other operations. 
Therefore, we can decompose WER to the sum of insertion rate and the rate of deletions or substitutions. 

In the lexical noise model, the insertion rate equals to the expected number of insertions for each gold word. Since the probability of not inserting is $1-p_I$, the expected number of repetitions before not inserting succeeds is $\frac{p_I}{1-p_I}$. It is also a mean of a geometric distribution with $p=1-p_I$. 

The rate of deletions and substitutions is $p_D + (1 - p_D) p_S$, where $p_D$ is the number of deletions. The words that were not deleted can be substituted, and there is $(1-p_D) p_S$ of them. In summary, the original model WER is 

\begin{equation}
    \text{WER} = \frac{p_I}{1-p_I} + p_D + (1 - p_D) p_S,
    \label{formula:wer}
\end{equation}


To get a custom target WER, we rescale the learned probabilities by a constant $c$:

\begin{equation}
    \text{WER}_\text{desired} = \frac{cp_I}{1-cp_I} + cp_D + (1 - cp_D) cp_S.
\end{equation}




We simplify the equation above to 

\begin{equation}
    \text{WER}_\text{desired} \approx cp_I + cp_D + (1 - cp_D) cp_S.
\end{equation}

It leads to a quadratic function where $c$ can be found easily. Since we work with probabilities, we select the smallest non-negative root as the solution.
We release our implementation online.\footnote{\url{https://github.com/pe-trik/asr-errors-simulator}}

\paragraph{Training the noise model}
For training the noise model, we utilize VoxPopuli \cite{wang-etal-2021-voxpopuli} to retrieve around 100,000 audio and gold transcript sentences in English and 60,000 in German. They are from the same domain as ESIC, both corpora are from the European Parliament.
We processed the audio with NVidia NeMo CTC ASRs\footnote{
stt\_de\_quartznet15x5 and stt\_en\_conformer\_ctc\_large from \url{https://catalog.ngc.nvidia.com/models}%
} \cite{kuchaiev2019nemo,gulati20_interspeech}. Then we trained the rules of the lexical noise model and applied them on source data. Since the result is deterministic on the random seed of the lexical noise model, we perform multi-sourcing using three different seeds and report average BLEU scores with standard deviation.




\paragraph{Results with transcription noise}

\cref{tab:wer-noise} summarizes the BLEU scores of two-source MT with different levels of transcription noise in each of the sources on two sets: ESIC dev with reference translated from English, and Newstest11 with balanced reference. \Cref{sec:chrf-appendix} contains the corresponding \chrf{} scores. Table~\ref{tab:test-set-results} shows the results on the ESIC test set for the settings where multi-source models achieved the highest improvement due to noisy inputs. 

In Table~\ref{tab:wer-noise}, on both sets, we observe that the less noisy single source achieves higher BLEU than the other single source. When the difference in noise levels between the sources is small (close to diagonal in the table), then multi-sourcing reaches slightly higher BLEU than single sources. In case of balanced Newstest11, this area matches the diagonal. In case of ESIC with English original source and reference translated from English, the area of multi-source outperforming single-source is shifted. This tendency is reflected in the test set results in Table~\ref{tab:test-set-results} as well. Only when the German source is less noisy than the English one, it does improve BLEU in multi-sourcing. We explain it by the discrepancy of source languages for MT and reference that affect BLEU the same way as in offline mode in \Cref{sec:multisrc-st}. On Newstest11, with the references translated from German, we expect the reverse.

We also observe expected behavior that the more noise, the lower BLEU in all setups. Compare e.g.\ 33.3 BLEU with zero noise and 12.1 with 40\% WER in both sources. With very large noise, it is possible that neither option would be usable. 
In ESIC dev, e.g.\ when English WER is 20\%, we observe large span, between 5 and 25\% WER in German, where multi-sourcing outperforms single source at least by several hundreths of BLEU. This span in Newstest11 is much more narrow, only 20 to 25\% WER in German. We hypothesize that it may be caused by the domain difference. The lexical noise model is trained on Europarl. In news domain, there may be fewer words for substitution, so the noise consists more of deletions and insertions, and it might be more harmful for MT in combination of two sources. However, multi-sourcing appears to be robust to ASR errors regardless of whether we have one or both sources as original.






\begin{figure*}[th]
    \centering
\input{multisimulgrid}
    \caption{
    Single-sourcing vs multi-sourcing with different level of artificial ASR noise of the sources (\% WER) in simultaneous mode on ESIC dev set. The results are depicted as quality (BLEU) and latency (AL) trade-off of the candidate systems. The plots highlighted by gray background show noise levels where multi-sourcing (En+De, blue line) outperforms both single sources in BLEU at least for AL$>$5.5. 
    }
    \label{fig:multisimul}
\end{figure*}
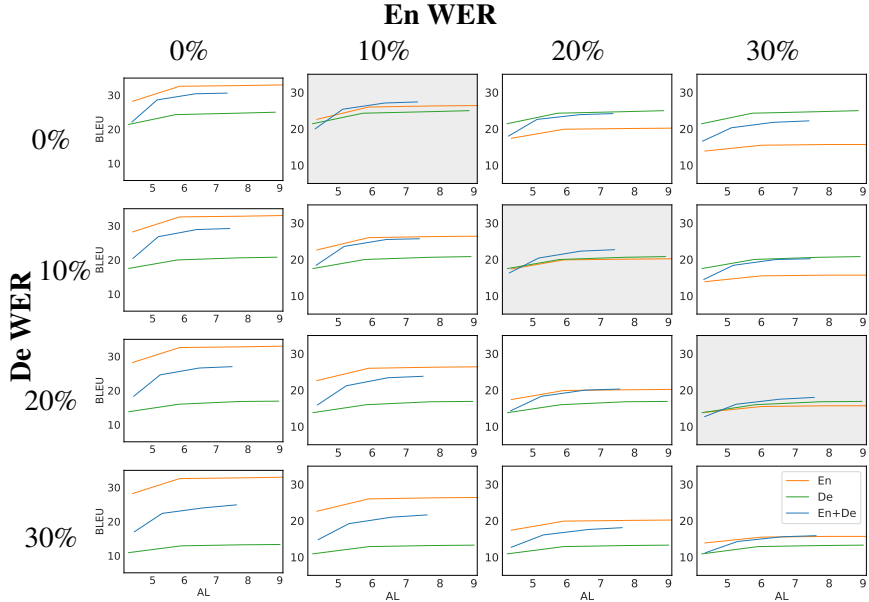

\section{Simultaneous Multi-Source}
\label{sec:simul-mtmain}
In the previous section, we experimented with offline translation with artificial ASR noise and showed that multi-source models are indeed robust to noise. However, one important use case of speech translation is in a real time setting where simultaneous MT is used. We therefore evaluate the robustness of multi-source models in a simultaneous setting.

\subsection{Simultaneous Machine Translation}
\label{sec:simulmt}


Simultaneous MT is a task that simulates one subtask of a technology that translates long-form monologue speech in real-time, or with the lowest possible latency.
There exist two main approaches to simultaneous MT: streaming and re-translating \cite{niehues18_interspeech,arivazhagan-etal-2020-translation}. Re-translating systems generate preliminary translation hypotheses that can be updated. Both approaches have complementary benefits and drawbacks. In this paper, we focus on streaming. 

We assume that simultaneous MT continuously receives an input text segmented to sentences, one token at a time, as produced by the speaker and upstream tasks.
After reading each input token, the system can either produce one or more target tokens, or decide to read the next input token, e.g.\ to have more context for translation.
The goal of simultaneous MT is to translate the input with high quality and low latency. Quality is measured on full sentences as in standard text-to-text MT, e.g.\ by BLEU.
The standard latency measure of simultaneous MT is Average Lagging (AL, \inparcite{ma-etal-2019-stacl}). It is an average number of tokens behind an ``optimal'' policy that generates the target proportionally with reading the source.


Simultaneous MT can be created from standard text-to-text NMT by applying any simultaneous decoding algorithm. However, it is recommendable first to adapt NMT, so it is inclined to translate consecutive sentence prefixes with the same target prefix.
We use Local Agreement (LA-$n$) as a decoding algorithm. It achieved good performance by the best performing system \cite{polak-etal-2022-cuni} in the most recent IWSLT competition \cite{anastasopoulos-etal-2022-findings}.
Local Agreement (LA-$n$) means that $n$ consecutive updates must agree on a target prefix to commit and write. The last committed prefix is then forced as a prefix to decoding the next units. 
Agreement size $n$ is a parameter that controls the latency.

\subsection{Creating Simultaneous MT Systems}
\label{sec:creating}

In \Cref{sec:multisrc-st}, we used multi-way models trained on full sentences, but in a simultaneous setting, these models will make mistakes when translating partial sentences using the LA-$n$ approach. Therefore, our multi-way models should first be adapted for partial sentence translation. To this end, we used the multi-way English and German to Czech MT model as a base for simultaneous MT. 
We fine-tuned the last trained model checkpoint for stable translation on 1:1 mix of incomplete sentence prefixes and full sentences as \citet{niehues18_interspeech}. For each source-target pair of the training data, we selected 5-times 1 to 90 \% of source and target characters and rounded them to full words. Then, we ran training for 1 day on 1 GPU. We validated BLEU score on ESIC dev and Normalized Erasure (NE, \inparcite{arivazhagan-etal-2020-translation}) on all prefixes of the first 65 sentences (around 1500 words) of ESIC dev set. We ran fine-tuning with multi-way data for English and German as source languages, and for bilingual English-Czech and German-Czech MT.


We stopped training after one day when there were no improvements in stability or quality. Then, we selected one checkpoint for English and one for German that reached acceptable quality and stability values. See \Cref{sec:checkpoint-sel} for details.


\subsection{Multi-Sourcing in Simultaneous MT}
\label{sec:simul-multisrc}

We use late averaging of the two selected checkpoints for multi-sourcing in simultaneous MT. The only aspects of multi-sourcing in simultaneous mode that differ from single-source or non-simultaneous mode are synchronization of the sources and how to count Average Lagging.

\paragraph{Synchronization}
In a realistic use-case, it is necessary to synchronize the original speech and simultaneous interpreting. However, we leave it for further work, as our goal is to inspect the limits of multi-sourcing. Therefore, we simulate a case where the sources are optimally synchronized, aligned and parallel to sentence level.

In multi-source mode, we sort all sentence prefixes by proportion of the character length to the sentence length. Each ``Read'' operation of the multi-source system then receives two prefixes in two languages. One of them is updated by one new token. Every such update is counted to local agreement size. We note that there are other strategies, e.g.\ count only English source updates to LA-$n$, but in this paper we have other goal than searching for the best strategy. 

\paragraph{AL in multi-source}
In multi-source setup, we only count Read operations of the English source to AL calculations that we report, and not of the German source because the sources are simultaneous. Counting only German
tokens differs negligibly, approximately by 0.1 tokens.

\subsection{Simultaneous Multi-Source with Artifical Noise}
\label{sec:simul-multisrc-noise}

We want to compare multi-sourcing model to single-sourcing with artificial ASR noise model as in \Cref{sec:asr-noise}.
We evaluate each system on the latency levels with local agreement sizes 2, 5, 10 and 15. Since each evaluation takes approximately 5 hours on 2000 sentences, we report only one run, and not average and deviation on multiple randomly noised inputs.

The results on ESIC dev set are in \Cref{fig:multisimul}. We can observe the same trends as in the offline case. The single source that is noised less achieves higher BLEU. Multi-sourcing outperforms both single sources when both noise levels are similar and when the English one is lower, e.g.\ in the case with 10\% WER in German and 20\% WER in English. We explain it again by the fact that the Czech reference is translated from English, and not German. 

Furthermore, on both ESIC and Newstest11 (\Cref{fig:multisimul}) 
we observe that multi-sourcing performs worse in the low latency modes, i.e.\ in AL$<$5 that roughly corresponds to LA$<$5. 
We assume that the proportional synchronization of the two sources is often inaccurate and may confuse late averaging. In higher latency modes, the synchronization noise at the end of input may be lowered by local agreement. Having validated the multi-source NMT is robust to ASR errors in both full sentence and simultaneous settings, we have paved the way for harder settings where multilingual interpretations of the original source available with different amounts of delay can be used for translation.


\section{Conclusion}

We have investigated the robustness of multi-source NMT to transcription errors in order to motivate its use in settings where ASRs for the original speech and its simultaneously interpreted equivalent are available. To this end, we first analyzed the 10-hour ESIC corpus and documented that the ASR errors in the two sources are indeed independent, indicating their complementary nature. We then simulated transcription noise for English and German when translating into Czech in single and multi-source NMT settings and observed that using multiple noisy sources is significantly better than individual noisy sources. We then repeated experiments in a simultaneous translation setting and showed that multi-source translation continues to be robust to noise. This robustness of multi-source NMT to noise motivates future research into simultaneous multi-source speech translation, where one source is available with a delay. We will also consider training models with simulated ASR errors to further increase their robustness, especially in multi-source settings.


\section{Limitations}
Although we have shown the robustness of multi-source NMT to transcription errors in a full-sentence and simultaneous settings, our work has the following limitations:

\begin{itemize}
    \item Our work does not address the case where the additional source, typically interpreted, is available after a delay. A delayed source may reduce the gains seen by multi-sourcing.
    \item We have only focused on the Local Agreement (LA-$n$) approach for simultaneous translation and exploration of other simultaneous approaches such as wait-$k$ remains.
    \item Human evaluation of translations is pending.
    \item Evaluation on other language pairs is pending.
\end{itemize}

\section*{Acknowledgements}

The research was partially supported by the
grants 19-26934X (NEUREM3) of the Czech Science Foundation, ``Grant Schemes at CU'' (reg.\ no.\ CZ.02.2.69/0.0/0.0/19\_073/0016935), 398120 of the Grant Agency of Charles University, and SVV
project number 260 698. Part of the work was done during an internship at NICT.

\bibliography{anthology,custom}
\bibliographystyle{acl_natbib}

\appendix

\section{\chrf{} Scores with Noisy Inputs}
\label{sec:chrf-appendix}

There is an evidence that \chrf{} correlates with human judgements better than BLEU. In \Cref{tab:wer-noise-chrf}, we can see that for multi-sourcing with noisy inputs on ESIC dev, \chrf{} are indeed higher than single-sourcing, and this correlates with the BLEU score gains in \Cref{tab:wer-noise}. On the other hand, for Newstest11, \chrf{} scores do not indicate any improvements. While the corresponding BLEU scores in \Cref{tab:wer-noise} indicated improvements of multi-sourcing with noisy inputs, the magnitude of these gains were minor, much smaller than those observed for ESIC. This gives us sufficient reason to believe that multi-sourcing should be useful in a setting like ESIC, where the reference is created from only one source, which is more realistic than the ``balanced'' use-case of Newstest11, where the reference originates from 5 languages.

\begin{table*}[ht]

\input{defoutertable.tex}
{\begin{center}%
\setlength\tabcolsep{1.5pt}%
\footnotesize
\scalebox{1.00}{
\begin{tabular}{rr|r|rrrrrrrrrrrr}
\input{esicdevchrftable.tex}
\input{newschrftable.tex}
\end{tabular}
}
\end{center}}

\caption{
\chrf{} (avg\plusm{}stddev) with transcription noise on ESIC dev set whose
reference translations was English and on Newstest11 (news11) with balanced reference source language. The area with the green background is where the \colorbox{colorone}{English} single-source outperforms \colorbox{colortwo}{German} single-source. \munderline{Black underlined} numbers indicate the area where multi-sourcing achieves higher score than both single-sourcing options. \textcolor{colorlow}{Red-colored} numbers are where at least one single-source scores higher.
}
    \label{tab:wer-noise-chrf}
\end{table*}

\section{Checkpoint Selection for Simultaneous Multi-Source}
\label{sec:checkpoint-sel}

The checkpoints that we selected for simultaneous multi-source decoding (recall \Cref{sec:creating}) was the multi-way checkpoint for English and bilingual one for German. 
\Cref{tab:finetuning} summarizes the results of fine-tuning for stability. BLEU decreased marginally (by 0.2 on English and 0.9 on German), while normalized erasure (NE) dropped by 40\% in English and 52\% on German. 

Based on some outputs, we explain higher NE in German-to-Czech by discrepancy in word orders. Many erasures were caused by an incorrect presumption of the final verb. Regardless, our fine-tuned models exhibit significantly reduced NE and can be reliably used for simultaneous translation using the LA-$n$ approach.

\begin{table}[h]
    \centering
    \begin{tabular}{c|cc|ccc}
         & \multicolumn{2}{c|}{En} & \multicolumn{2}{c}{De} \\
checkpoint & BLEU & NE & BLEU & NE \\
         \hline
starting    & 33.2 & 1.77 & 25.9 & 3.15 \\
selected & 33.0 & 1.21 & 25.0 & 1.52 \\
\hline
diff     & -0.2 & -40\% & -0.9 & -52\% \\
    \end{tabular}
    \caption{The results of fine-tuning for stability.}
    \label{tab:finetuning}
\end{table}

\section{Effect of Reference Source Language}
\label{sec:ref-src}

To explain the effect of reference source language, we run a contrastive evaluation on the subset of Newstest11 that consists only from the documents that originate in English. We compare BLEU measures with a reference translated directly from Czech, and with three additional references translated only from German \cite{news11}. 

The results of simultaneous mode (recall \Cref{sec:simul-mtmain}) are in \Cref{fig:newsmultisimul}. We observe the same trends as in offline mode in \Cref{sec:multisrc-st}. The BLEU score is higher for the single source with the language from which the reference was translated. When this source is noised substantially more than the other, multi-sourcing outperforms both by a small margin. 

In case of German references, the nearest margin to single-sourcing is much smaller than with the English references. We assume it is because the structural difference of English source and German-Czech references is larger than German to English-Czech reference. It is documented also by BLEU scores with zero noise (33 and 20 on references from English vs 16 and 30 on references from German).

\begin{figure*}[t]
    \centering
\input{newsmultisimulgrid}
    \caption{
    Single-sourcing vs multi-sourcing with different level of artificial ASR
	noise of the sources (\% WER) in simultaneous mode on 
	Newstest11 subset (598 sentences) originally in English.
	In the upper grid, the Czech reference is translated from English, while
	in the lower, there is average and standard deviation of BLEU counted
	against the 3 additional references translated from German \cite{news11}.
	Grey highlighting indicates area where multi-sourcing (En+De, blue line)
	outperforms or is on-par with both single sources in BLEU.
    }
    \label{fig:newsmultisimul}
\end{figure*}



\end{document}

%% file: authors.tex
\author{Dominik Macháček$^{1,2}$ \and Peter Polák$^1$ \and Ondřej Bojar$^1$ \and Raj Dabre$^2$
  \\ \\
  Charles University, Faculty of Mathematics and Physics, \\ 
  Institute of Formal and Applied Linguistics$^{1}$ \\
   \\
  National Institute of Information and Communications Technology, Kyoto, Japan$^{2}$\\
  \texttt{\{machacek,polak,bojar\}@ufal.mff.cuni.cz}, \texttt{raj.dabre@nict.go.jp}}

%% file: outertable.tex
\input{defoutertable.tex}

{\begin{center}%
\setlength\tabcolsep{1.5pt}%
\footnotesize
\scalebox{1.00}{
\begin{tabular}{rr|r|rrrrrrrrrrrr}
\input{innertable.tex}
\input{newstableedited.tex}
\end{tabular}
}
\end{center}}

%% file: defoutertable.tex
\def\plusm{$\pm$}
\def\colorfield#1#2#3{\cellcolor{#2}{\textcolor{#3}{#1}}}
\def\wer#1{#1}
\def\singlesrc{single-src.}
\def\rsinglesrc{s-src.}
%
%

\definecolor{colorone}{HTML}{E8F0C2}
\definecolor{colortwo}{HTML}{CAC2F0}
%
%
\definecolor{colorlow}{HTML}{0B51EE}
\definecolor{colorlow}{HTML}{B11030}

\definecolor{colorhigh}{RGB}{0,0,0}
\def\emptyother{}
\def\enwercol{\multicolumn{9}{c}{\textbf{En WER}}}
\def\dewercol{\multirow{9}{*}{\rotatebox[origin=c]{90}{\textbf{De WER}}}}
\def\munderline#1{\dunderline{1.1pt}{#1}}
\def\wrapn#1#2{#1\textsuperscript{{$\pm$#2}}}
\def\outertablemacro#1#2{
{\begin{center}%
\setlength\tabcolsep{1.5pt}%
\footnotesize
\scalebox{1.00}{
\begin{tabular}{rr|r|rrrrrrrrrrrr}
\input{#1}
\input{#2}
\end{tabular}
}
\end{center}}
}

%% file: innertable.tex
\\ \multicolumn{2}{c|}{BLEU} & \multicolumn{1}{c|}{\textbf{ESIC dev}} & \enwercol{} 
  \\ 
 & & \singlesrc{} & \wer{0~\%} & \wer{5~\%} & \wer{10~\%} & \wer{15~\%} & \wer{20~\%} & \wer{25~\%} & \wer{30~\%} & \wer{35~\%} & \wer{40~\%} &   \\ 
\hline
& \rsinglesrc{} &  & \colorfield{\wrapn{33.3}{0.0}}{white}{black} & \colorfield{\wrapn{29.7}{0.3}}{white}{black} & \colorfield{\wrapn{26.3}{0.4}}{white}{black} & \colorfield{\wrapn{22.9}{0.4}}{white}{black} & \colorfield{\wrapn{20.4}{0.5}}{white}{black} & \colorfield{\wrapn{18.2}{0.8}}{white}{black} & \colorfield{\wrapn{15.8}{0.1}}{white}{black} & \colorfield{\wrapn{14.0}{0.2}}{white}{black} & \colorfield{\wrapn{12.1}{0.1}}{white}{black} &   \\ 
\hline
 \dewercol{} 
 & \wer{0~\%} & \colorfield{\wrapn{26.1}{0.0}}{white}{black} & \colorfield{\wrapn{31.9}{0.0}}{colorone}{colorlow} & \colorfield{\munderline{\wrapn{30.0}{0.2}}}{colorone}{colorhigh} & \colorfield{\textbf{\munderline{\wrapn{28.5}{0.3}}}}{colorone}{colorhigh} & \colorfield{\munderline{\wrapn{26.6}{0.1}}}{colortwo}{colorhigh} & \colorfield{\wrapn{25.2}{0.4}}{colortwo}{colorlow} & \colorfield{\wrapn{23.8}{0.3}}{colortwo}{colorlow} & \colorfield{\wrapn{21.9}{0.3}}{colortwo}{colorlow} & \colorfield{\wrapn{20.5}{0.2}}{colortwo}{colorlow} & \colorfield{\wrapn{19.3}{0.3}}{colortwo}{colorlow} &   \\
 & \wer{5~\%} & \colorfield{\wrapn{23.5}{0.0}}{white}{black} & \colorfield{\wrapn{30.9}{0.1}}{colorone}{colorlow} & \colorfield{\wrapn{29.1}{0.2}}{colorone}{colorlow} & \colorfield{\munderline{\wrapn{27.6}{0.3}}}{colorone}{colorhigh} & \colorfield{\textbf{\munderline{\wrapn{25.7}{0.1}}}}{colortwo}{colorhigh} & \colorfield{\munderline{\wrapn{24.2}{0.4}}}{colortwo}{colorhigh} & \colorfield{\wrapn{22.8}{0.4}}{colortwo}{colorlow} & \colorfield{\wrapn{21.1}{0.4}}{colortwo}{colorlow} & \colorfield{\wrapn{19.6}{0.2}}{colortwo}{colorlow} & \colorfield{\wrapn{18.6}{0.2}}{colortwo}{colorlow} &   \\
 & \wer{10~\%} & \colorfield{\wrapn{21.6}{0.2}}{white}{black} & \colorfield{\wrapn{30.0}{0.2}}{colorone}{colorlow} & \colorfield{\wrapn{28.0}{0.1}}{colorone}{colorlow} & \colorfield{\munderline{\wrapn{26.6}{0.4}}}{colorone}{colorhigh} & \colorfield{\munderline{\wrapn{24.6}{0.3}}}{colorone}{colorhigh} & \colorfield{\munderline{\wrapn{23.4}{0.2}}}{colortwo}{colorhigh} & \colorfield{\munderline{\wrapn{21.9}{0.4}}}{colortwo}{colorhigh} & \colorfield{\wrapn{20.2}{0.1}}{colortwo}{colorlow} & \colorfield{\wrapn{18.7}{0.2}}{colortwo}{colorlow} & \colorfield{\wrapn{17.5}{0.5}}{colortwo}{colorlow} &   \\
 & \wer{15~\%} & \colorfield{\wrapn{19.0}{0.3}}{white}{black} & \colorfield{\wrapn{28.9}{0.2}}{colorone}{colorlow} & \colorfield{\wrapn{27.1}{0.1}}{colorone}{colorlow} & \colorfield{\wrapn{25.7}{0.4}}{colorone}{colorlow} & \colorfield{\munderline{\wrapn{23.7}{0.2}}}{colorone}{colorhigh} & \colorfield{\munderline{\wrapn{22.4}{0.4}}}{colorone}{colorhigh} & \colorfield{\munderline{\wrapn{21.0}{0.4}}}{colortwo}{colorhigh} & \colorfield{\munderline{\wrapn{19.3}{0.2}}}{colortwo}{colorhigh} & \colorfield{\wrapn{17.8}{0.3}}{colortwo}{colorlow} & \colorfield{\wrapn{16.7}{0.4}}{colortwo}{colorlow} &   \\
 & \wer{20~\%} & \colorfield{\wrapn{17.1}{0.3}}{white}{black} & \colorfield{\wrapn{27.9}{0.4}}{colorone}{colorlow} & \colorfield{\wrapn{26.6}{0.2}}{colorone}{colorlow} & \colorfield{\wrapn{24.9}{0.4}}{colorone}{colorlow} & \colorfield{\munderline{\wrapn{22.9}{0.1}}}{colorone}{colorhigh} & \colorfield{\munderline{\wrapn{21.7}{0.5}}}{colorone}{colorhigh} & \colorfield{\munderline{\wrapn{20.0}{0.4}}}{colorone}{colorhigh} & \colorfield{\munderline{\wrapn{18.3}{0.2}}}{colortwo}{colorhigh} & \colorfield{\wrapn{17.0}{0.1}}{colortwo}{colorlow} & \colorfield{\wrapn{15.7}{0.1}}{colortwo}{colorlow} &   \\
 & \wer{25~\%} & \colorfield{\wrapn{15.6}{0.3}}{white}{black} & \colorfield{\wrapn{27.1}{0.3}}{colorone}{colorlow} & \colorfield{\wrapn{25.7}{0.2}}{colorone}{colorlow} & \colorfield{\wrapn{24.1}{0.3}}{colorone}{colorlow} & \colorfield{\wrapn{22.1}{0.2}}{colorone}{colorlow} & \colorfield{\munderline{\wrapn{20.7}{0.4}}}{colorone}{colorhigh} & \colorfield{\munderline{\wrapn{19.2}{0.5}}}{colorone}{colorhigh} & \colorfield{\munderline{\wrapn{17.4}{0.2}}}{colorone}{colorhigh} & \colorfield{\munderline{\wrapn{16.3}{0.2}}}{colortwo}{colorhigh} & \colorfield{\wrapn{14.9}{0.1}}{colortwo}{colorlow} &   \\
 & \wer{30~\%} & \colorfield{\wrapn{13.8}{0.2}}{white}{black} & \colorfield{\wrapn{25.9}{0.3}}{colorone}{colorlow} & \colorfield{\wrapn{24.5}{0.4}}{colorone}{colorlow} & \colorfield{\wrapn{22.8}{0.3}}{colorone}{colorlow} & \colorfield{\wrapn{20.9}{0.3}}{colorone}{colorlow} & \colorfield{\wrapn{19.6}{0.2}}{colorone}{colorlow} & \colorfield{\munderline{\wrapn{18.3}{0.2}}}{colorone}{colorhigh} & \colorfield{\munderline{\wrapn{16.3}{0.4}}}{colorone}{colorhigh} & \colorfield{\munderline{\wrapn{15.1}{0.1}}}{colorone}{colorhigh} & \colorfield{\munderline{\wrapn{13.9}{0.2}}}{colortwo}{colorhigh} &   \\
 & \wer{35~\%} & \colorfield{\wrapn{12.5}{0.2}}{white}{black} & \colorfield{\wrapn{24.6}{0.4}}{colorone}{colorlow} & \colorfield{\wrapn{22.5}{0.4}}{colorone}{colorlow} & \colorfield{\wrapn{20.9}{0.2}}{colorone}{colorlow} & \colorfield{\wrapn{19.2}{0.1}}{colorone}{colorlow} & \colorfield{\wrapn{18.1}{0.5}}{colorone}{colorlow} & \colorfield{\wrapn{16.7}{0.3}}{colorone}{colorlow} & \colorfield{\wrapn{15.3}{0.3}}{colorone}{colorlow} & \colorfield{\munderline{\wrapn{14.1}{0.2}}}{colorone}{colorhigh} & \colorfield{\munderline{\wrapn{12.9}{0.1}}}{colortwo}{colorhigh} &   \\
 & \wer{40~\%} & \colorfield{\wrapn{10.8}{0.1}}{white}{black} & \colorfield{\wrapn{23.4}{0.4}}{colorone}{colorlow} & \colorfield{\wrapn{21.4}{0.1}}{colorone}{colorlow} & \colorfield{\wrapn{20.1}{0.3}}{colorone}{colorlow} & \colorfield{\wrapn{18.3}{0.5}}{colorone}{colorlow} & \colorfield{\wrapn{17.3}{0.2}}{colorone}{colorlow} & \colorfield{\wrapn{16.0}{0.1}}{colorone}{colorlow} & \colorfield{\wrapn{14.4}{0.1}}{colorone}{colorlow} & \colorfield{\wrapn{13.2}{0.2}}{colorone}{colorlow} & \colorfield{\munderline{\wrapn{12.1}{0.1}}}{colorone}{colorhigh} &   \\

%% file: newstableedited.tex
%
\\ \multicolumn{2}{c|}{BLEU} & \multicolumn{1}{c|}{\textbf{news11}} & \enwercol{} 
  \\ 
 & & \singlesrc{} & \wer{0~\%} & \wer{5~\%} & \wer{10~\%} & \wer{15~\%} & \wer{20~\%} & \wer{25~\%} & \wer{30~\%} & \wer{35~\%} & \wer{40~\%} &   \\ 
\hline
& \rsinglesrc{} &  & \colorfield{\wrapn{23.4}{0.0}}{white}{black} & \colorfield{\wrapn{21.1}{0.2}}{white}{black} & \colorfield{\wrapn{19.2}{0.1}}{white}{black} & \colorfield{\wrapn{17.1}{0.0}}{white}{black} & \colorfield{\wrapn{15.3}{0.1}}{white}{black} & \colorfield{\wrapn{13.6}{0.2}}{white}{black} & \colorfield{\wrapn{12.2}{0.2}}{white}{black} & \colorfield{\wrapn{10.6}{0.3}}{white}{black} & \colorfield{\wrapn{9.6}{0.0}}{white}{black} &   \\ 
\hline
 \dewercol{} 
 & \wer{0~\%} & \colorfield{\wrapn{22.9}{0.0}}{white}{black} & \colorfield{\munderline{\wrapn{24.0}{0.0}}}{colorone}{colorhigh} & \colorfield{\wrapn{22.7}{0.0}}{colortwo}{colorlow} & \colorfield{\wrapn{21.4}{0.2}}{colortwo}{colorlow} & \colorfield{\wrapn{20.2}{0.1}}{colortwo}{colorlow} & \colorfield{\wrapn{18.9}{0.2}}{colortwo}{colorlow} & \colorfield{\wrapn{17.8}{0.1}}{colortwo}{colorlow} & \colorfield{\wrapn{16.9}{0.1}}{colortwo}{colorlow} & \colorfield{\wrapn{15.5}{0.1}}{colortwo}{colorlow} & \colorfield{\wrapn{14.6}{0.2}}{colortwo}{colorlow} &   \\
 & \wer{5~\%} & \colorfield{\wrapn{20.6}{0.1}}{white}{black} & \colorfield{\wrapn{23.2}{0.1}}{colorone}{colorlow} & \colorfield{\munderline{\wrapn{21.8}{0.1}}}{colorone}{colorhigh} & \colorfield{\munderline{\wrapn{20.7}{0.0}}}{colortwo}{colorhigh} & \colorfield{\wrapn{19.2}{0.0}}{colortwo}{colorlow} & \colorfield{\wrapn{18.2}{0.1}}{colortwo}{colorlow} & \colorfield{\wrapn{17.1}{0.1}}{colortwo}{colorlow} & \colorfield{\wrapn{16.1}{0.1}}{colortwo}{colorlow} & \colorfield{\wrapn{14.8}{0.0}}{colortwo}{colorlow} & \colorfield{\wrapn{13.9}{0.1}}{colortwo}{colorlow} &   \\
 & \wer{10~\%} & \colorfield{\wrapn{18.8}{0.1}}{white}{black} & \colorfield{\wrapn{22.5}{0.1}}{colorone}{colorlow} & \colorfield{\munderline{\wrapn{21.3}{0.2}}}{colorone}{colorhigh} & \colorfield{\munderline{\wrapn{20.1}{0.1}}}{colorone}{colorhigh} & \colorfield{\wrapn{18.6}{0.2}}{colortwo}{colorlow} & \colorfield{\wrapn{17.7}{0.1}}{colortwo}{colorlow} & \colorfield{\wrapn{16.4}{0.1}}{colortwo}{colorlow} & \colorfield{\wrapn{15.5}{0.1}}{colortwo}{colorlow} & \colorfield{\wrapn{14.1}{0.1}}{colortwo}{colorlow} & \colorfield{\wrapn{13.2}{0.2}}{colortwo}{colorlow} &   \\
 & \wer{15~\%} & \colorfield{\wrapn{17.0}{0.3}}{white}{black} & \colorfield{\wrapn{21.6}{0.2}}{colorone}{colorlow} & \colorfield{\wrapn{20.3}{0.2}}{colorone}{colorlow} & \colorfield{\wrapn{19.1}{0.2}}{colorone}{colorlow} & \colorfield{\munderline{\wrapn{17.8}{0.1}}}{colorone}{colorhigh} & \colorfield{\wrapn{16.9}{0.1}}{colortwo}{colorlow} & \colorfield{\wrapn{15.6}{0.0}}{colortwo}{colorlow} & \colorfield{\wrapn{14.7}{0.1}}{colortwo}{colorlow} & \colorfield{\wrapn{13.4}{0.1}}{colortwo}{colorlow} & \colorfield{\wrapn{12.5}{0.1}}{colortwo}{colorlow} &   \\
 & \wer{20~\%} & \colorfield{\wrapn{15.4}{0.2}}{white}{black} & \colorfield{\wrapn{20.8}{0.0}}{colorone}{colorlow} & \colorfield{\wrapn{19.5}{0.1}}{colorone}{colorlow} & \colorfield{\wrapn{18.3}{0.1}}{colorone}{colorlow} & \colorfield{\wrapn{17.0}{0.2}}{colorone}{colorlow} & \colorfield{\munderline{\wrapn{16.0}{0.1}}}{colortwo}{colorhigh} & \colorfield{\wrapn{14.9}{0.1}}{colortwo}{colorlow} & \colorfield{\wrapn{14.0}{0.1}}{colortwo}{colorlow} & \colorfield{\wrapn{12.7}{0.2}}{colortwo}{colorlow} & \colorfield{\wrapn{12.0}{0.1}}{colortwo}{colorlow} &   \\
 & \wer{25~\%} & \colorfield{\wrapn{13.8}{0.1}}{white}{black} & \colorfield{\wrapn{19.9}{0.2}}{colorone}{colorlow} & \colorfield{\wrapn{18.7}{0.2}}{colorone}{colorlow} & \colorfield{\wrapn{17.7}{0.1}}{colorone}{colorlow} & \colorfield{\wrapn{16.3}{0.1}}{colorone}{colorlow} & \colorfield{\munderline{\wrapn{15.4}{0.0}}}{colorone}{colorhigh} & \colorfield{\munderline{\wrapn{14.0}{0.2}}}{colortwo}{colorhigh} & \colorfield{\wrapn{13.2}{0.1}}{colortwo}{colorlow} & \colorfield{\wrapn{11.9}{0.0}}{colortwo}{colorlow} & \colorfield{\wrapn{11.1}{0.1}}{colortwo}{colorlow} &   \\
 & \wer{30~\%} & \colorfield{\wrapn{12.3}{0.3}}{white}{black} & \colorfield{\wrapn{19.2}{0.3}}{colorone}{colorlow} & \colorfield{\wrapn{17.9}{0.2}}{colorone}{colorlow} & \colorfield{\wrapn{16.9}{0.3}}{colorone}{colorlow} & \colorfield{\wrapn{15.6}{0.1}}{colorone}{colorlow} & \colorfield{\wrapn{14.5}{0.2}}{colorone}{colorlow} & \colorfield{\wrapn{13.5}{0.3}}{colorone}{colorlow} & \colorfield{\munderline{\wrapn{12.7}{0.2}}}{colortwo}{colorhigh} & \colorfield{\wrapn{11.3}{0.1}}{colortwo}{colorlow} & \colorfield{\wrapn{10.6}{0.1}}{colortwo}{colorlow} &   \\
 & \wer{35~\%} & \colorfield{\wrapn{11.2}{0.1}}{white}{black} & \colorfield{\wrapn{18.4}{0.0}}{colorone}{colorlow} & \colorfield{\wrapn{17.1}{0.1}}{colorone}{colorlow} & \colorfield{\wrapn{16.1}{0.1}}{colorone}{colorlow} & \colorfield{\wrapn{15.0}{0.2}}{colorone}{colorlow} & \colorfield{\wrapn{13.8}{0.1}}{colorone}{colorlow} & \colorfield{\wrapn{12.7}{0.2}}{colorone}{colorlow} & \colorfield{\wrapn{11.7}{0.1}}{colorone}{colorlow} & \colorfield{\wrapn{10.6}{0.2}}{colortwo}{colorlow} & \colorfield{\wrapn{9.9}{0.2}}{colortwo}{colorlow} &   \\
 & \wer{40~\%} & \colorfield{\wrapn{9.9}{0.3}}{white}{black} & \colorfield{\wrapn{17.1}{0.0}}{colorone}{colorlow} & \colorfield{\wrapn{16.1}{0.2}}{colorone}{colorlow} & \colorfield{\wrapn{14.9}{0.2}}{colorone}{colorlow} & \colorfield{\wrapn{14.0}{0.1}}{colorone}{colorlow} & \colorfield{\wrapn{12.9}{0.1}}{colorone}{colorlow} & \colorfield{\wrapn{11.7}{0.2}}{colorone}{colorlow} & \colorfield{\wrapn{10.7}{0.1}}{colorone}{colorlow} & \colorfield{\wrapn{9.9}{0.1}}{colorone}{colorlow} & \colorfield{\wrapn{9.1}{0.2}}{colortwo}{colorlow} &   

%% file: multisimulgrid.tex
\def\plotwidth{.16\linewidth}
\begin{tabular}{c@{}c@{}c@{}c@{}cccc}
\multicolumn{5}{c}{\textbf{En WER}} \\
 & 0\% & 10\% & 20\% & 30\%  \\
0\% & 
\includegraphics[width=\plotwidth,valign=m]{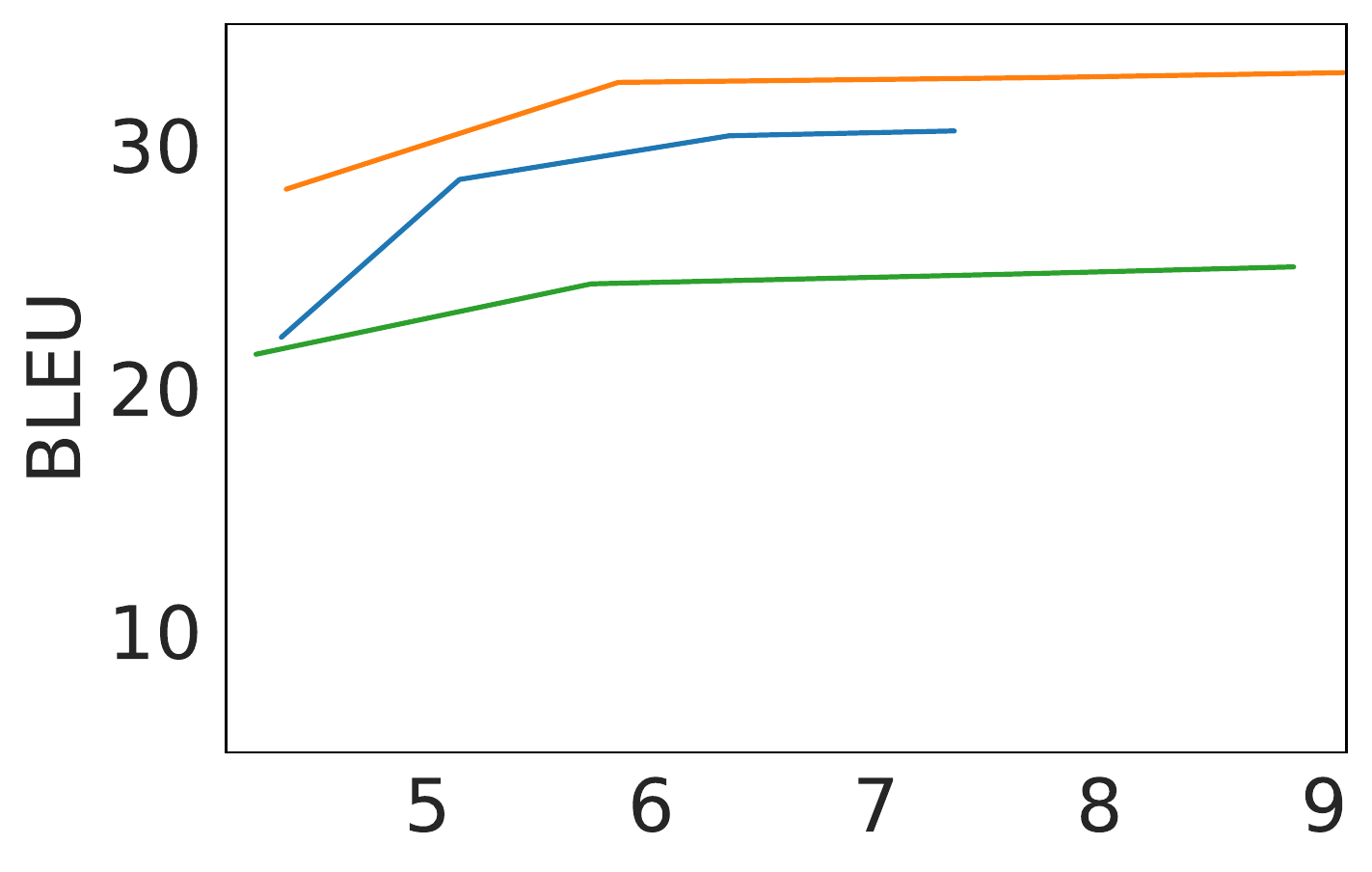} & 
\includegraphics[width=\plotwidth,valign=m]{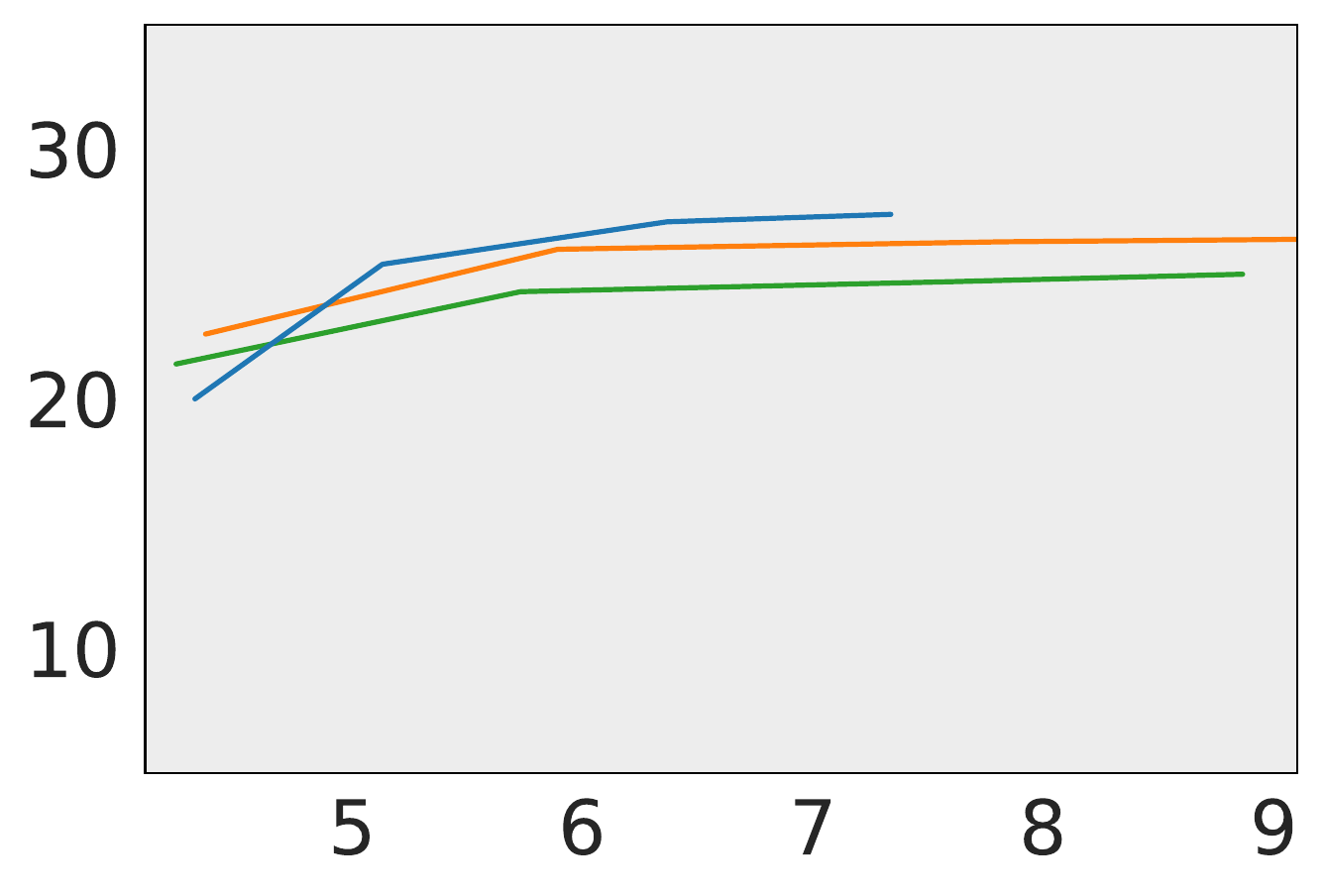} & 
\includegraphics[width=\plotwidth,valign=m]{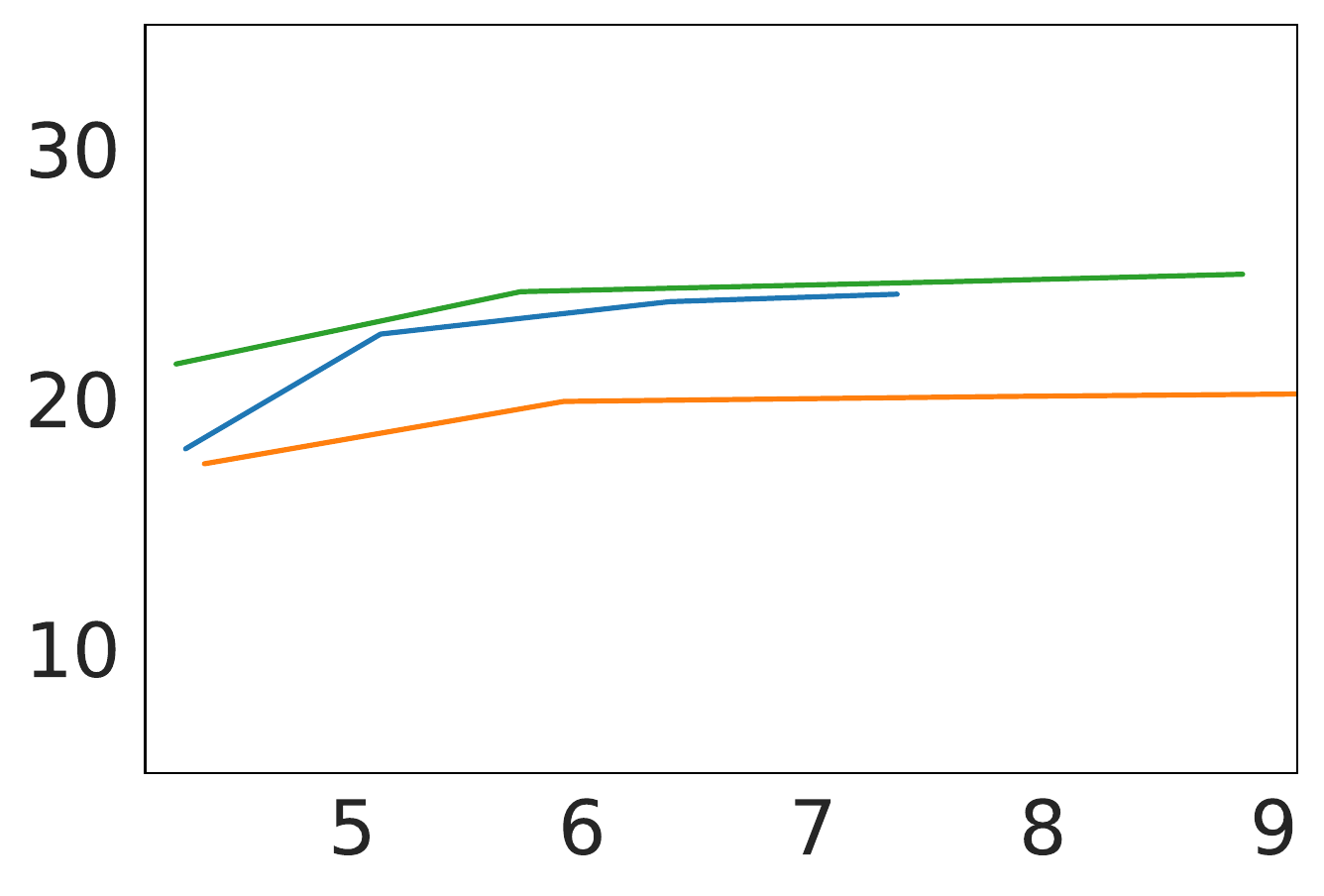} &
\includegraphics[width=\plotwidth,valign=m]{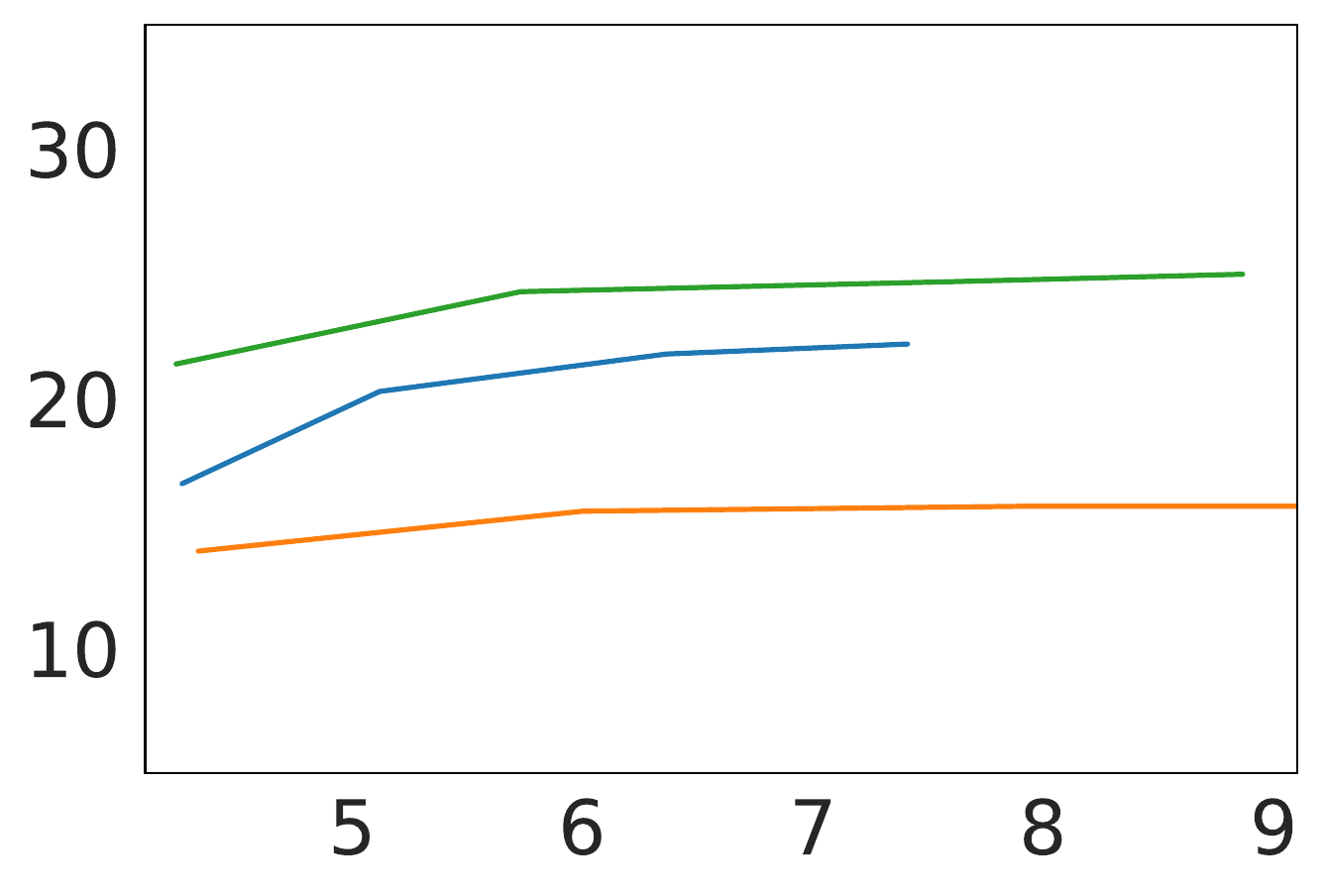} \\ 
                                                               
\multirow{4}{*}{\rotatebox[origin=c]{90}{\textbf{De WER}}}     
10\% &                                                         
\includegraphics[width=\plotwidth,valign=m]{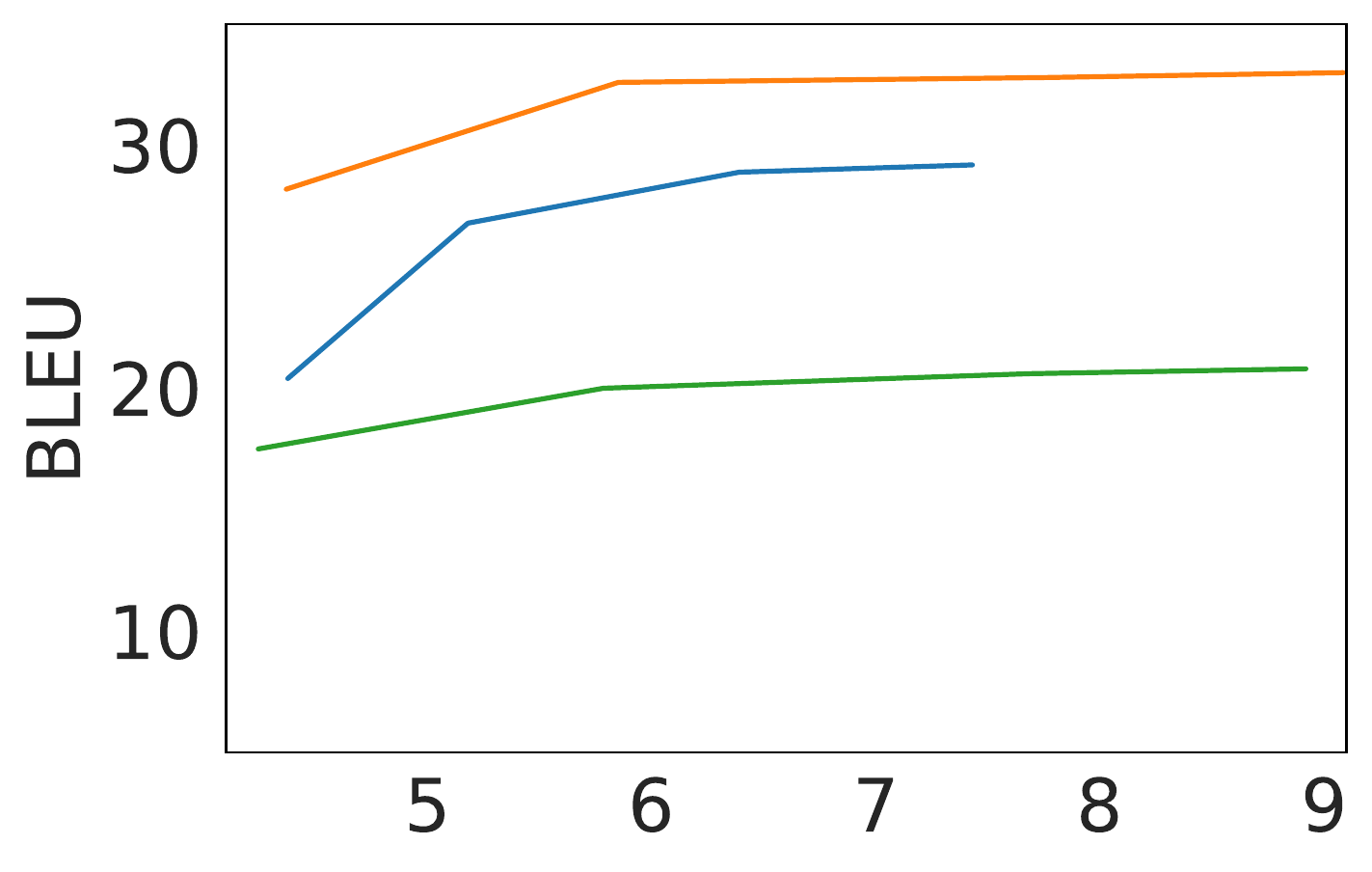} & 
\includegraphics[width=\plotwidth,valign=m]{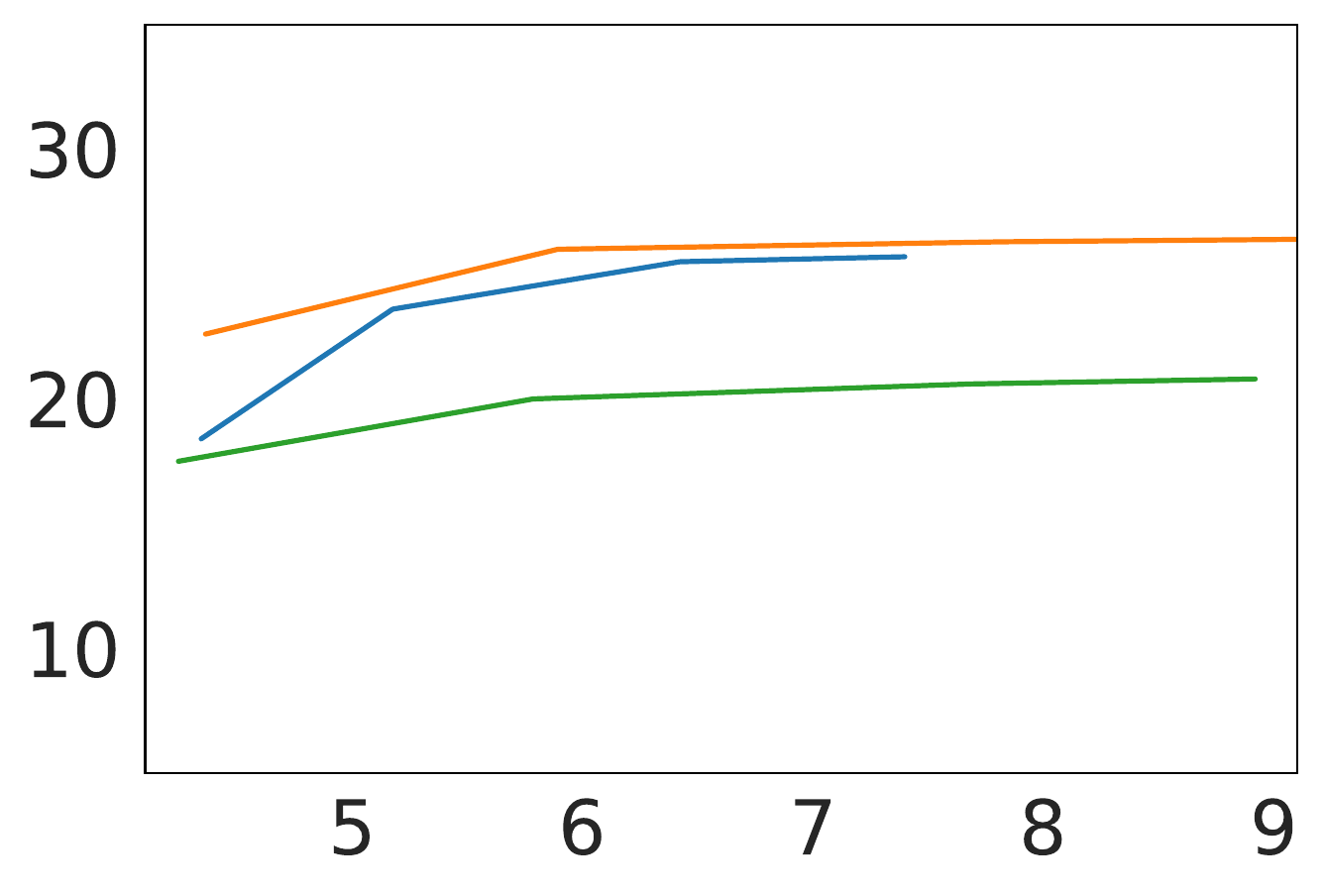} & 
\includegraphics[width=\plotwidth,valign=m]{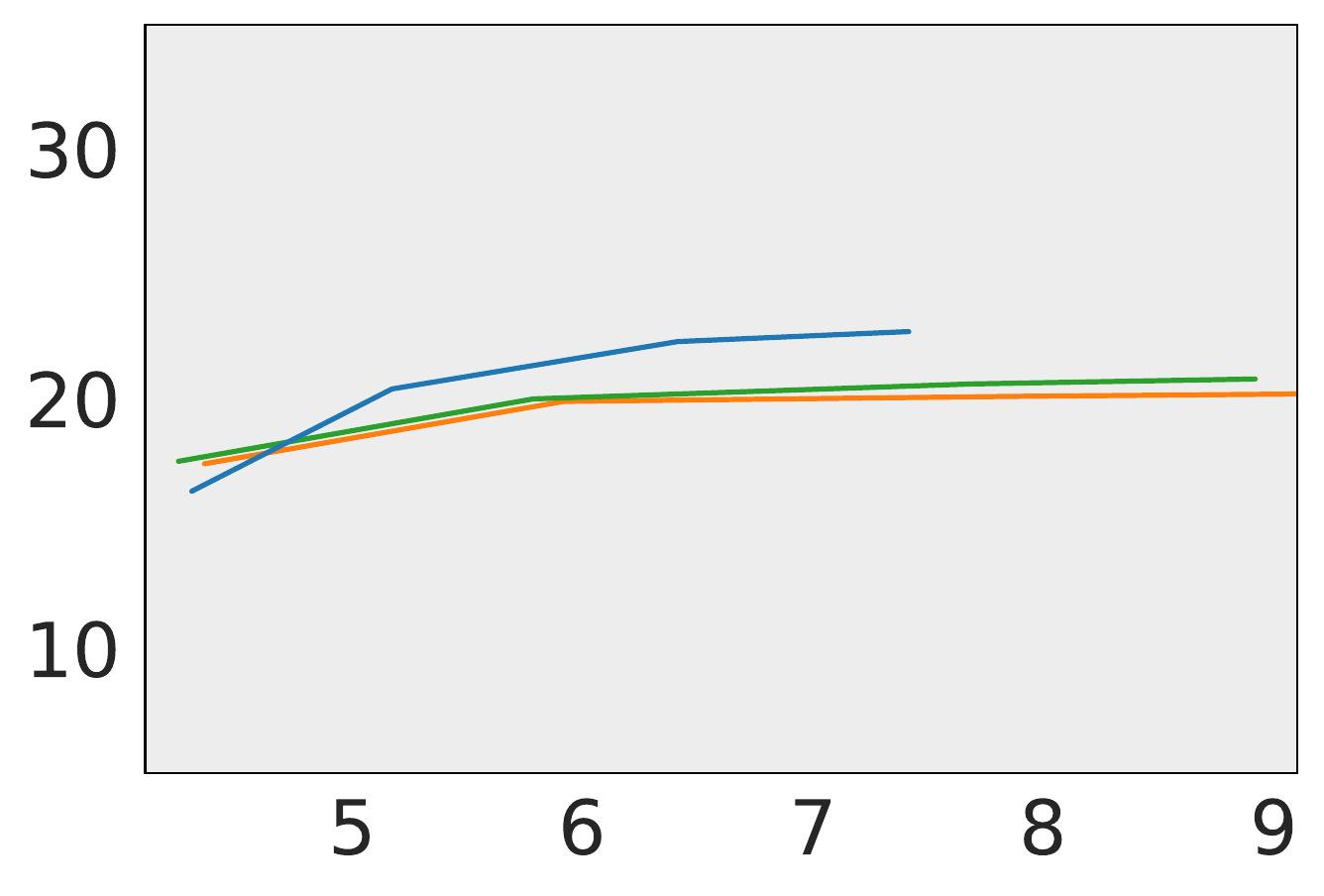} &
\includegraphics[width=\plotwidth,valign=m]{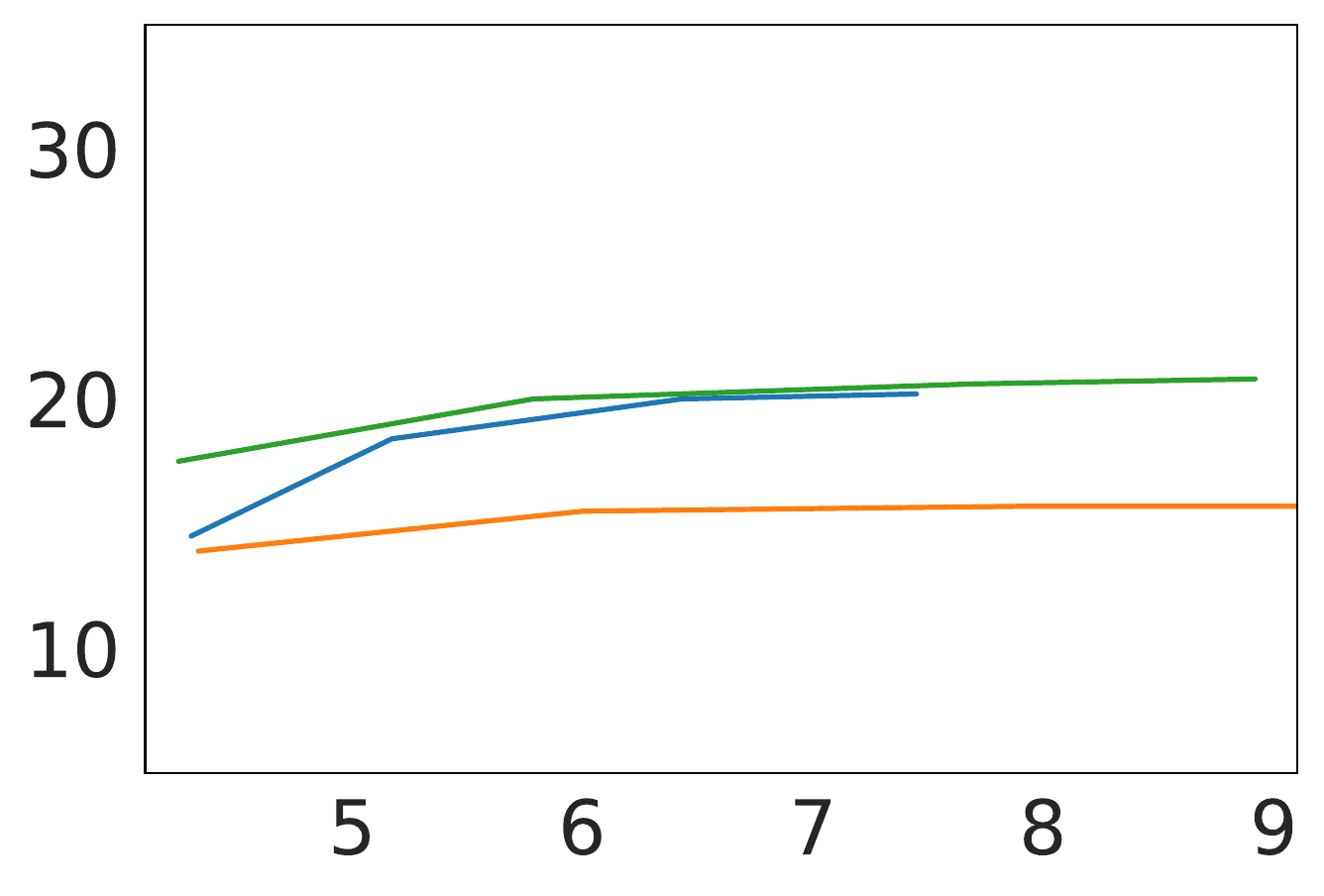} \\ 
                                                               
20\% &                                                         
\includegraphics[width=\plotwidth,valign=m]{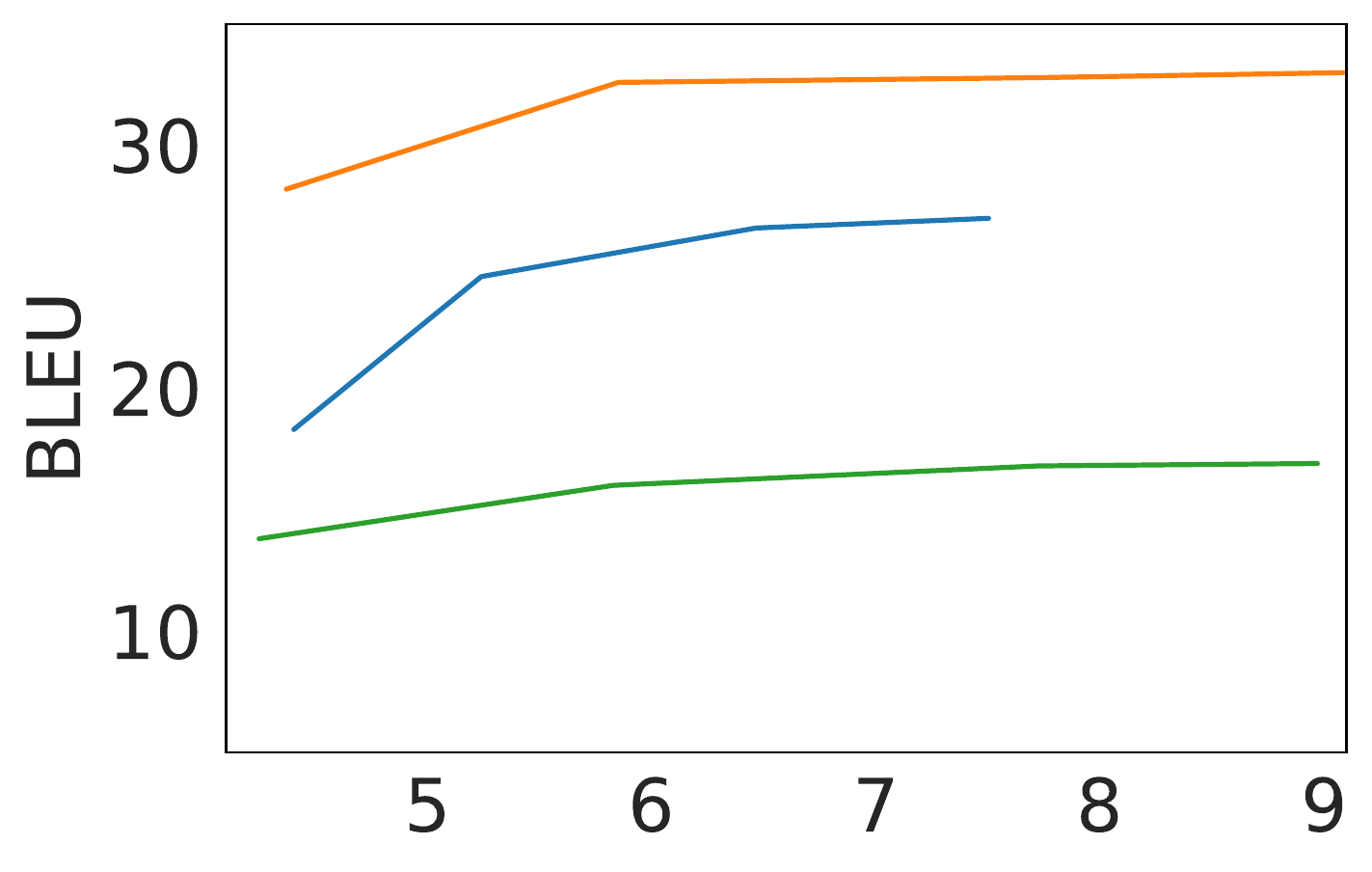} & 
\includegraphics[width=\plotwidth,valign=m]{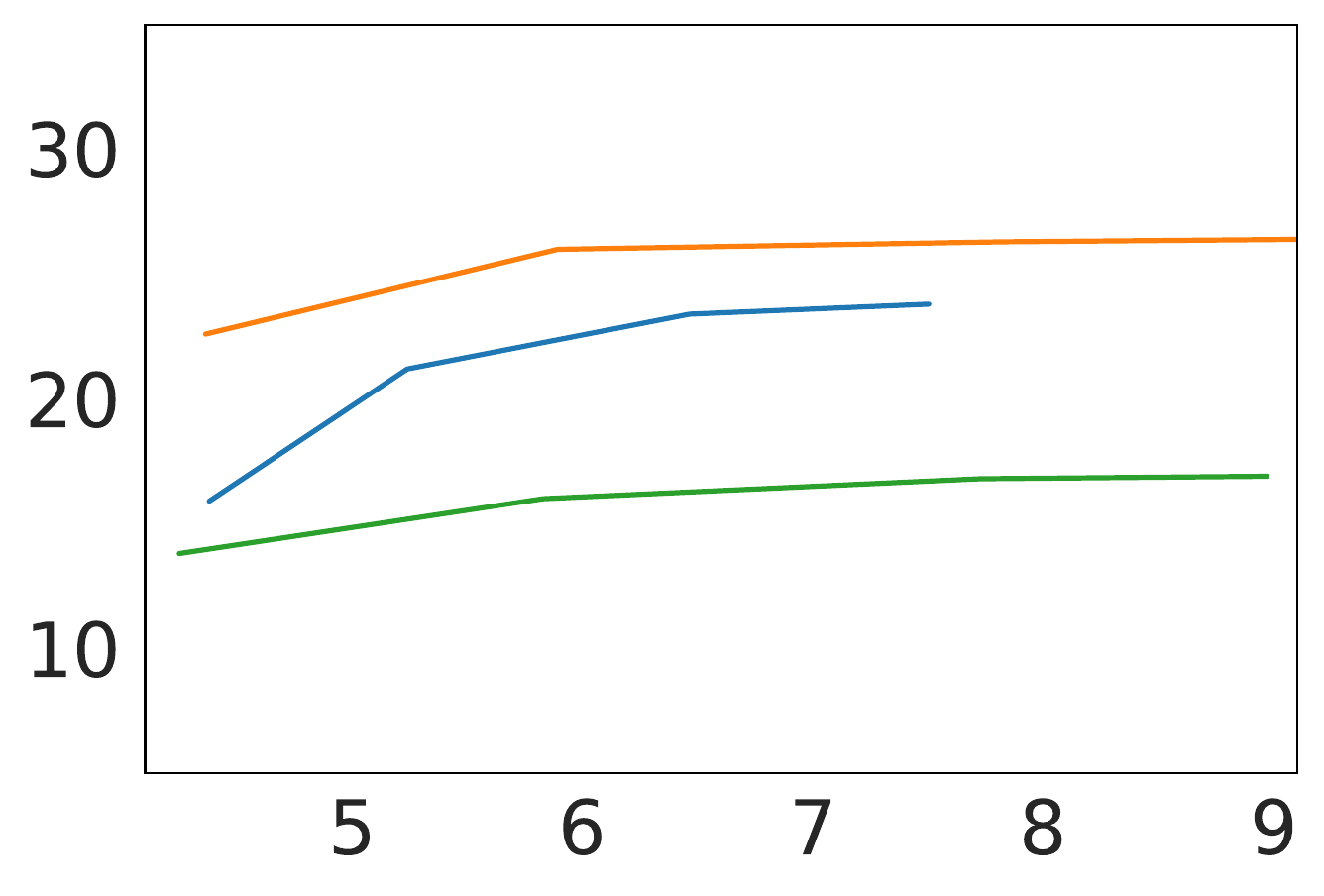} & 
\includegraphics[width=\plotwidth,valign=m]{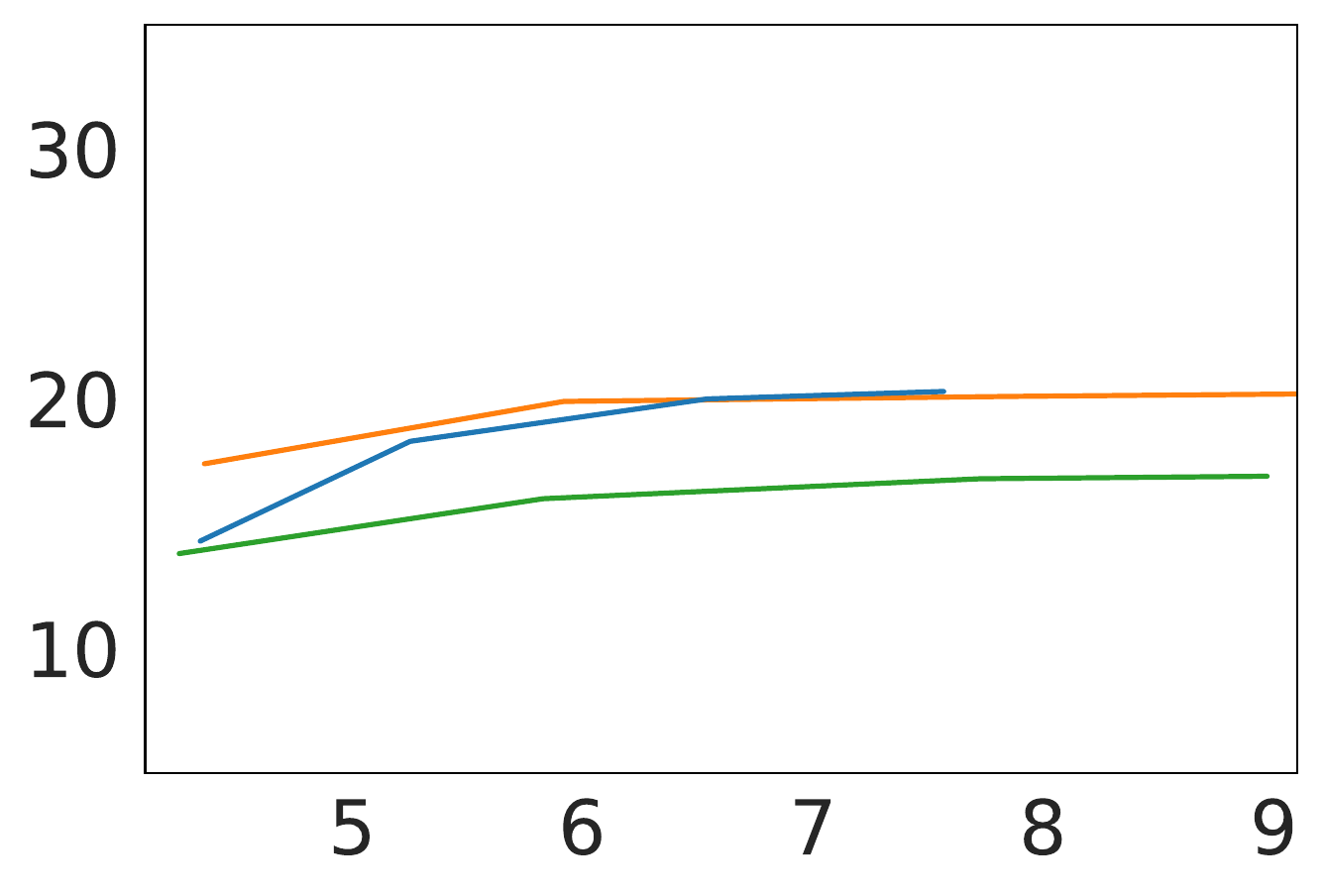} &
\includegraphics[width=\plotwidth,valign=m]{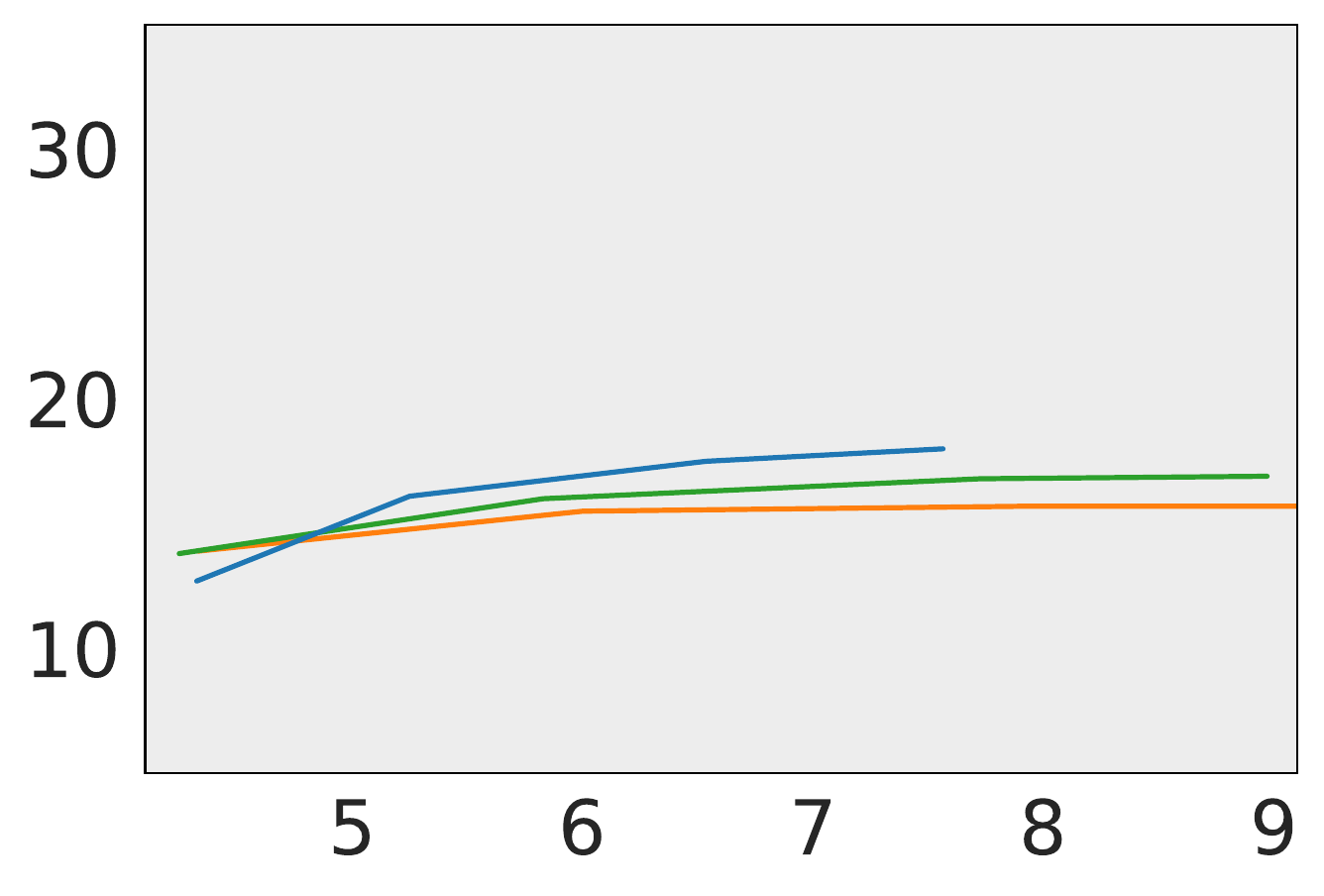} \\ 
                                                               
30\% &                                                         
\includegraphics[width=\plotwidth,valign=m]{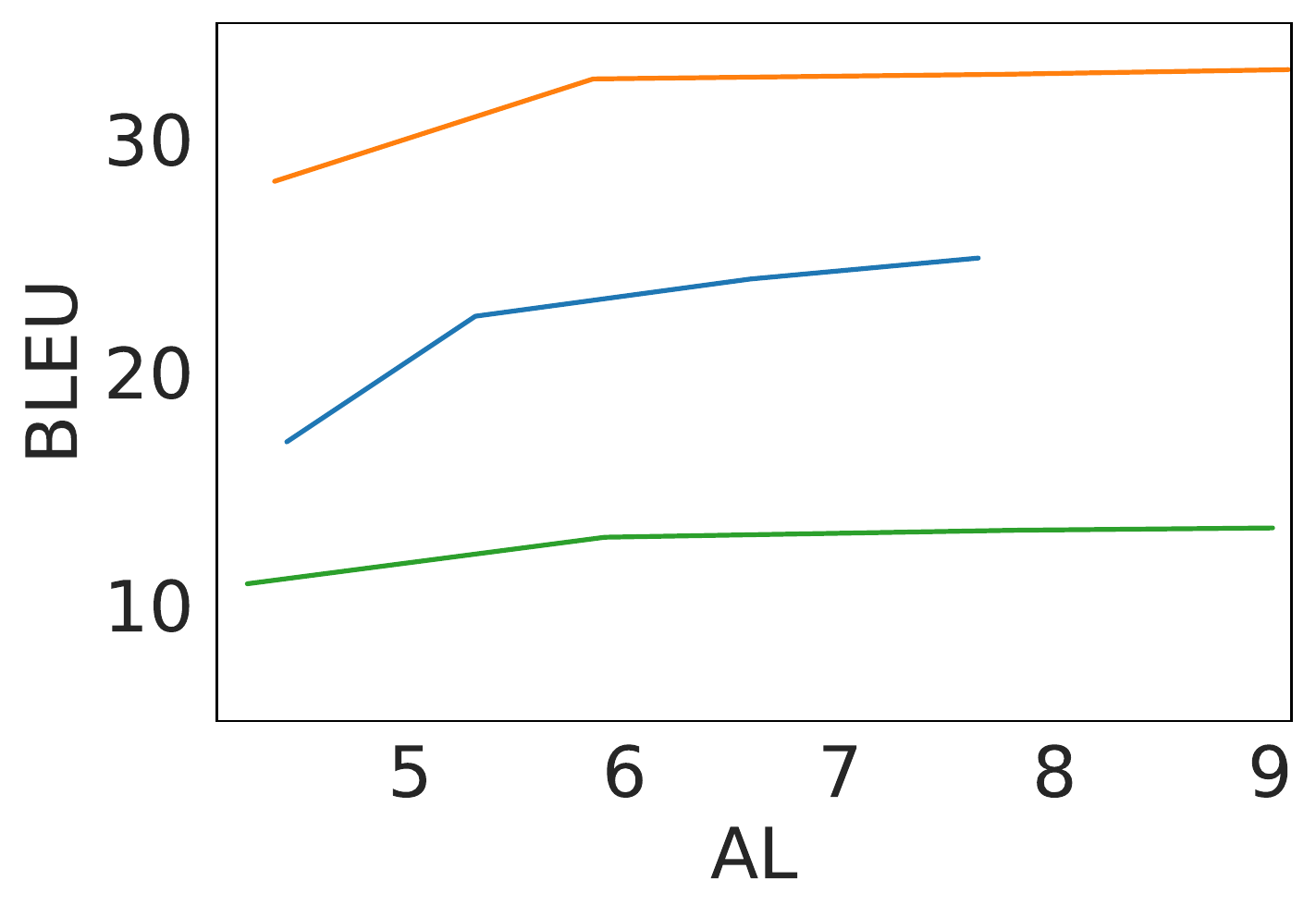} & 
\includegraphics[width=\plotwidth,valign=m]{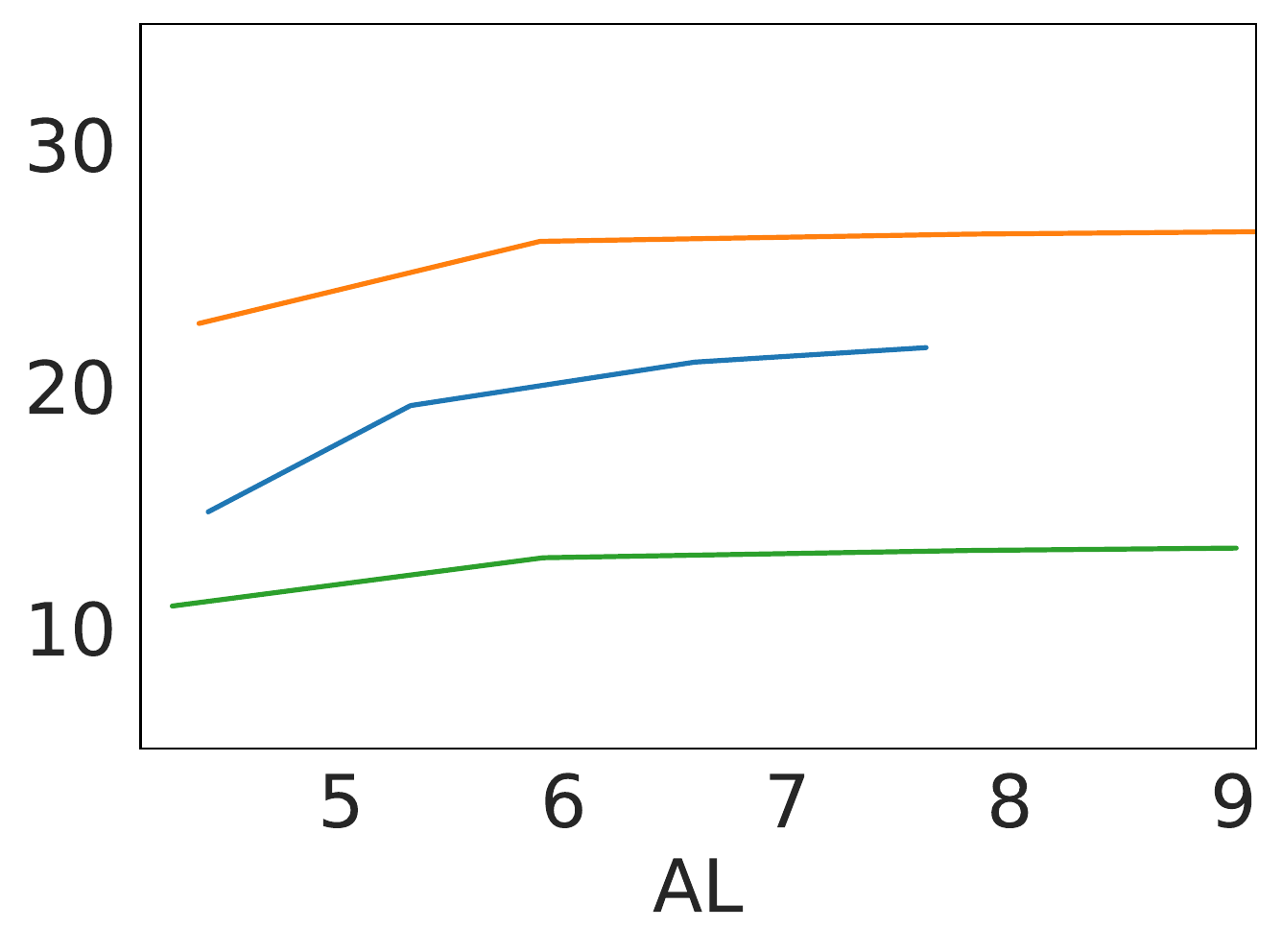} & 
\includegraphics[width=\plotwidth,valign=m]{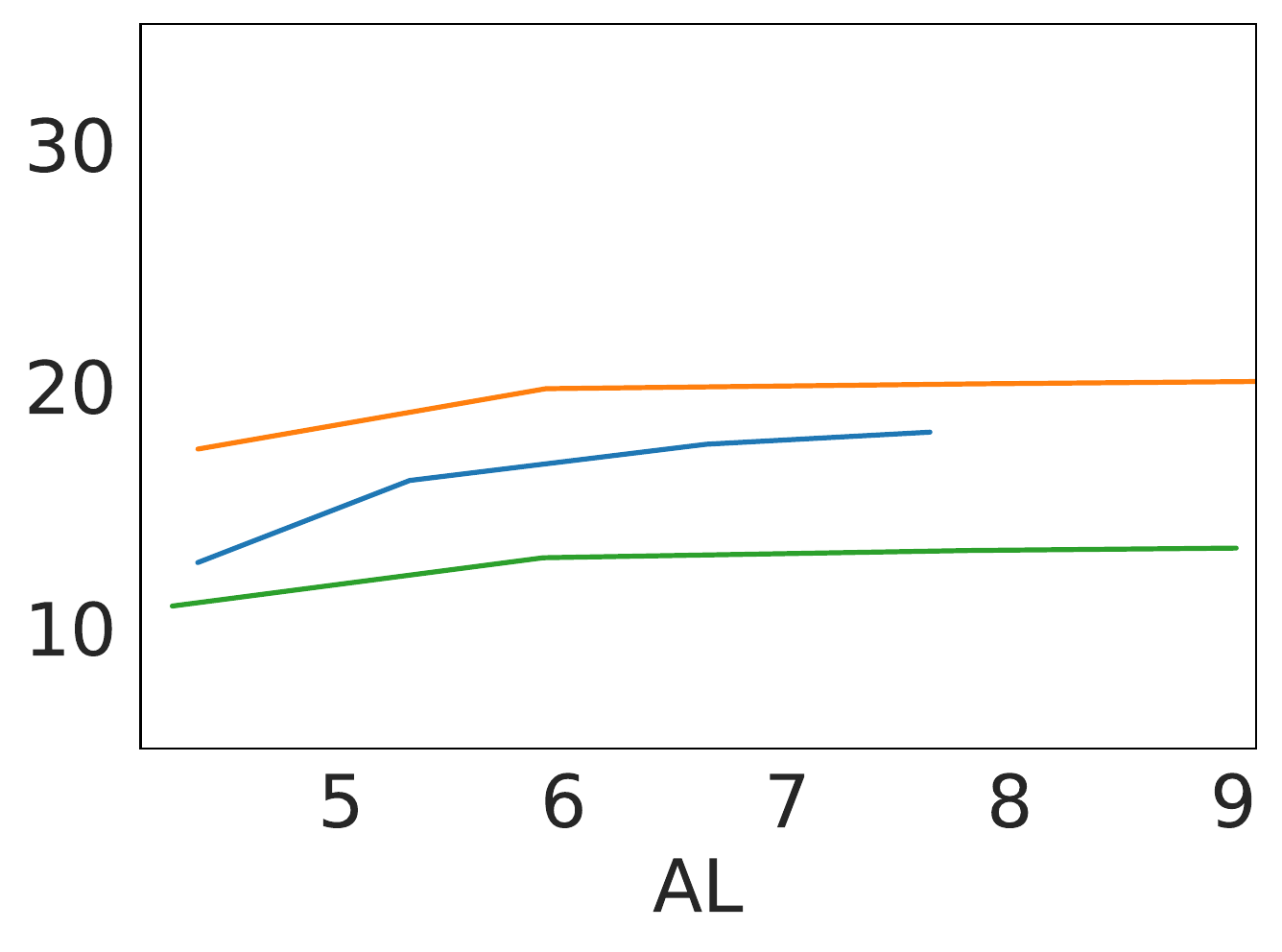} &
\includegraphics[width=\plotwidth,valign=m]{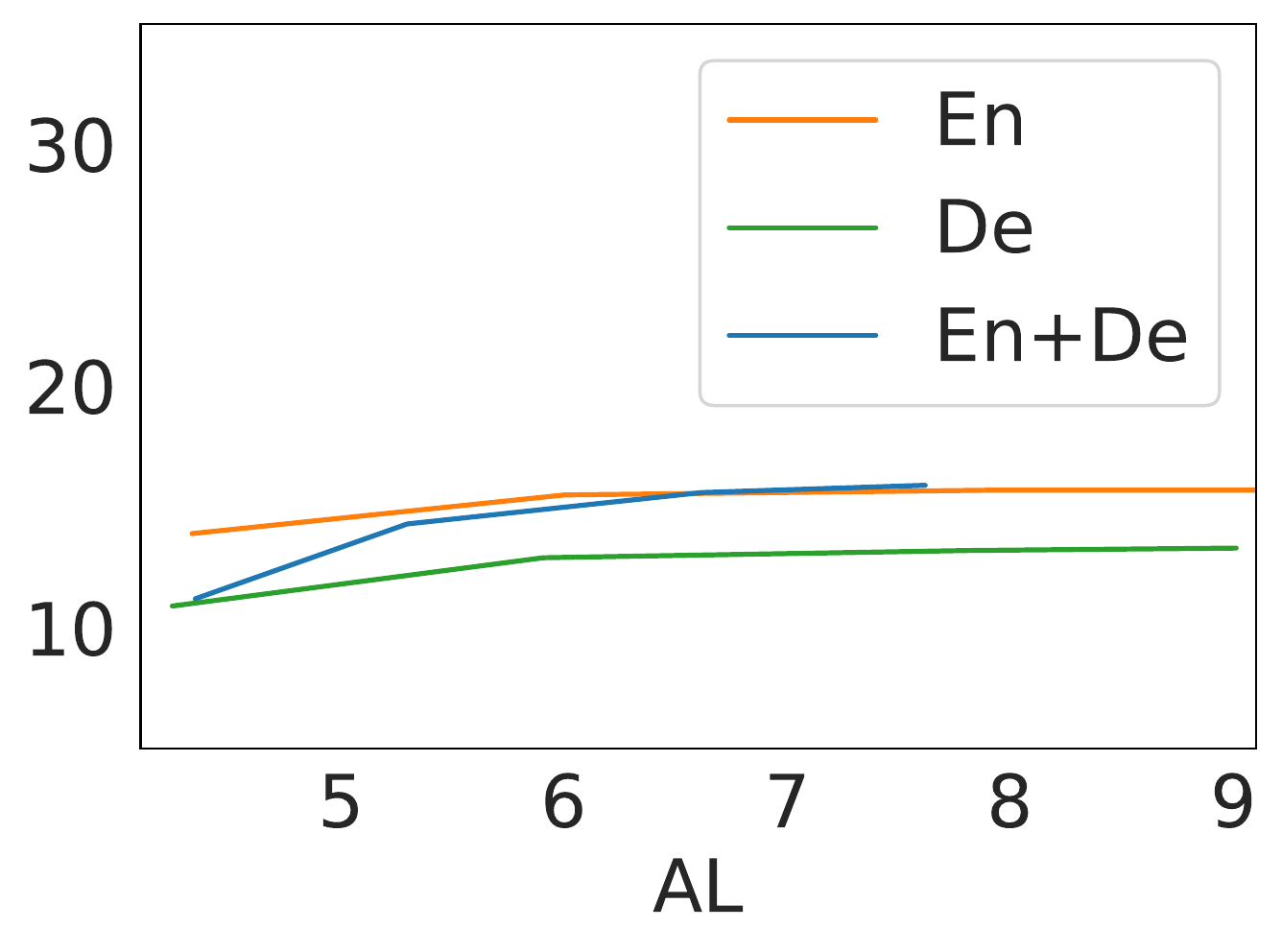} \\
\end{tabular}

%% file: esicdevchrftable.tex
\\ \multicolumn{2}{c|}{\chrf{}} & \multicolumn{1}{c|}{\textbf{ESIC dev}} & \enwercol{} 
  \\ 
 & & \singlesrc{} & \wer{0~\%} & \wer{5~\%} & \wer{10~\%} & \wer{15~\%} & \wer{20~\%} & \wer{25~\%} & \wer{30~\%} & \wer{35~\%} & \wer{40~\%} &   \\ 
\hline
& \rsinglesrc{} &  & \colorfield{\wrapn{60.2}{0.0}}{white}{black} & \colorfield{\wrapn{57.2}{0.2}}{white}{black} & \colorfield{\wrapn{54.4}{0.2}}{white}{black} & \colorfield{\wrapn{51.4}{0.3}}{white}{black} & \colorfield{\wrapn{49.2}{0.7}}{white}{black} & \colorfield{\wrapn{46.8}{0.8}}{white}{black} & \colorfield{\wrapn{44.3}{0.0}}{white}{black} & \colorfield{\wrapn{42.1}{0.3}}{white}{black} & \colorfield{\wrapn{40.1}{0.0}}{white}{black} &   \\ 
\hline
 \dewercol{} 
 & \wer{0~\%} & \colorfield{\wrapn{54.0}{0.0}}{white}{black} & \colorfield{\wrapn{58.6}{0.0}}{colorone}{colorlow} & \colorfield{\wrapn{56.9}{0.2}}{colorone}{colorlow} & \colorfield{\munderline{\wrapn{55.6}{0.1}}}{colorone}{colorhigh} & \colorfield{\wrapn{53.7}{0.2}}{colortwo}{colorlow} & \colorfield{\wrapn{52.3}{0.5}}{colortwo}{colorlow} & \colorfield{\wrapn{50.9}{0.4}}{colortwo}{colorlow} & \colorfield{\wrapn{49.2}{0.3}}{colortwo}{colorlow} & \colorfield{\wrapn{47.5}{0.2}}{colortwo}{colorlow} & \colorfield{\wrapn{46.1}{0.2}}{colortwo}{colorlow} &   \\
 & \wer{5~\%} & \colorfield{\wrapn{51.8}{0.1}}{white}{black} & \colorfield{\wrapn{57.7}{0.1}}{colorone}{colorlow} & \colorfield{\wrapn{56.2}{0.2}}{colorone}{colorlow} & \colorfield{\munderline{\wrapn{54.8}{0.1}}}{colorone}{colorhigh} & \colorfield{\munderline{\wrapn{52.9}{0.2}}}{colortwo}{colorhigh} & \colorfield{\wrapn{51.4}{0.6}}{colortwo}{colorlow} & \colorfield{\wrapn{50.0}{0.4}}{colortwo}{colorlow} & \colorfield{\wrapn{48.3}{0.3}}{colortwo}{colorlow} & \colorfield{\wrapn{46.7}{0.3}}{colortwo}{colorlow} & \colorfield{\wrapn{45.4}{0.2}}{colortwo}{colorlow} &   \\
 & \wer{10~\%} & \colorfield{\wrapn{49.9}{0.2}}{white}{black} & \colorfield{\wrapn{56.8}{0.2}}{colorone}{colorlow} & \colorfield{\wrapn{55.1}{0.1}}{colorone}{colorlow} & \colorfield{\wrapn{53.7}{0.3}}{colorone}{colorlow} & \colorfield{\munderline{\wrapn{51.8}{0.3}}}{colorone}{colorhigh} & \colorfield{\munderline{\wrapn{50.4}{0.3}}}{colortwo}{colorhigh} & \colorfield{\wrapn{49.0}{0.4}}{colortwo}{colorlow} & \colorfield{\wrapn{47.3}{0.1}}{colortwo}{colorlow} & \colorfield{\wrapn{45.6}{0.2}}{colortwo}{colorlow} & \colorfield{\wrapn{44.3}{0.2}}{colortwo}{colorlow} &   \\
 & \wer{15~\%} & \colorfield{\wrapn{47.6}{0.3}}{white}{black} & \colorfield{\wrapn{55.8}{0.0}}{colorone}{colorlow} & \colorfield{\wrapn{54.2}{0.1}}{colorone}{colorlow} & \colorfield{\wrapn{52.8}{0.3}}{colorone}{colorlow} & \colorfield{\wrapn{50.9}{0.2}}{colorone}{colorlow} & \colorfield{\munderline{\wrapn{49.6}{0.4}}}{colorone}{colorhigh} & \colorfield{\munderline{\wrapn{48.1}{0.5}}}{colortwo}{colorhigh} & \colorfield{\wrapn{46.4}{0.3}}{colortwo}{colorlow} & \colorfield{\wrapn{44.9}{0.3}}{colortwo}{colorlow} & \colorfield{\wrapn{43.6}{0.1}}{colortwo}{colorlow} &   \\
 & \wer{20~\%} & \colorfield{\wrapn{45.7}{0.3}}{white}{black} & \colorfield{\wrapn{54.9}{0.2}}{colorone}{colorlow} & \colorfield{\wrapn{53.5}{0.1}}{colorone}{colorlow} & \colorfield{\wrapn{51.9}{0.3}}{colorone}{colorlow} & \colorfield{\wrapn{50.2}{0.1}}{colorone}{colorlow} & \colorfield{\wrapn{48.7}{0.6}}{colorone}{colorlow} & \colorfield{\munderline{\wrapn{47.2}{0.4}}}{colorone}{colorhigh} & \colorfield{\wrapn{45.4}{0.2}}{colortwo}{colorlow} & \colorfield{\wrapn{43.9}{0.3}}{colortwo}{colorlow} & \colorfield{\wrapn{42.6}{0.3}}{colortwo}{colorlow} &   \\
 & \wer{25~\%} & \colorfield{\wrapn{44.0}{0.4}}{white}{black} & \colorfield{\wrapn{54.2}{0.4}}{colorone}{colorlow} & \colorfield{\wrapn{52.9}{0.1}}{colorone}{colorlow} & \colorfield{\wrapn{51.3}{0.2}}{colorone}{colorlow} & \colorfield{\wrapn{49.3}{0.2}}{colorone}{colorlow} & \colorfield{\wrapn{48.1}{0.4}}{colorone}{colorlow} & \colorfield{\wrapn{46.5}{0.4}}{colorone}{colorlow} & \colorfield{\munderline{\wrapn{44.7}{0.2}}}{colorone}{colorhigh} & \colorfield{\wrapn{43.3}{0.2}}{colortwo}{colorlow} & \colorfield{\wrapn{41.7}{0.0}}{colortwo}{colorlow} &   \\
 & \wer{30~\%} & \colorfield{\wrapn{42.1}{0.3}}{white}{black} & \colorfield{\wrapn{53.1}{0.3}}{colorone}{colorlow} & \colorfield{\wrapn{51.7}{0.3}}{colorone}{colorlow} & \colorfield{\wrapn{50.2}{0.2}}{colorone}{colorlow} & \colorfield{\wrapn{48.3}{0.3}}{colorone}{colorlow} & \colorfield{\wrapn{46.8}{0.3}}{colorone}{colorlow} & \colorfield{\wrapn{45.4}{0.5}}{colorone}{colorlow} & \colorfield{\wrapn{43.5}{0.3}}{colorone}{colorlow} & \colorfield{\munderline{\wrapn{42.2}{0.2}}}{colorone}{colorhigh} & \colorfield{\wrapn{40.6}{0.1}}{colortwo}{colorlow} &   \\
 & \wer{35~\%} & \colorfield{\wrapn{40.5}{0.2}}{white}{black} & \colorfield{\wrapn{52.0}{0.3}}{colorone}{colorlow} & \colorfield{\wrapn{50.0}{0.3}}{colorone}{colorlow} & \colorfield{\wrapn{48.7}{0.2}}{colorone}{colorlow} & \colorfield{\wrapn{46.9}{0.1}}{colorone}{colorlow} & \colorfield{\wrapn{45.7}{0.5}}{colorone}{colorlow} & \colorfield{\wrapn{44.1}{0.4}}{colorone}{colorlow} & \colorfield{\wrapn{42.4}{0.2}}{colorone}{colorlow} & \colorfield{\wrapn{41.0}{0.1}}{colorone}{colorlow} & \colorfield{\wrapn{39.7}{0.1}}{colortwo}{colorlow} &   \\
 & \wer{40~\%} & \colorfield{\wrapn{38.6}{0.2}}{white}{black} & \colorfield{\wrapn{51.1}{0.2}}{colorone}{colorlow} & \colorfield{\wrapn{49.2}{0.2}}{colorone}{colorlow} & \colorfield{\wrapn{47.8}{0.3}}{colorone}{colorlow} & \colorfield{\wrapn{46.0}{0.4}}{colorone}{colorlow} & \colorfield{\wrapn{44.8}{0.3}}{colorone}{colorlow} & \colorfield{\wrapn{43.2}{0.4}}{colorone}{colorlow} & \colorfield{\wrapn{41.5}{0.1}}{colorone}{colorlow} & \colorfield{\wrapn{39.8}{0.3}}{colorone}{colorlow} & \colorfield{\wrapn{38.6}{0.1}}{colorone}{colorlow} &   \\

%% file: newschrftable.tex
\\ \multicolumn{2}{c|}{\chrf{}} & \multicolumn{1}{c|}{\textbf{news11}} & \enwercol{} 
  \\ 
 & & \singlesrc{} & \wer{0~\%} & \wer{5~\%} & \wer{10~\%} & \wer{15~\%} & \wer{20~\%} & \wer{25~\%} & \wer{30~\%} & \wer{35~\%} & \wer{40~\%} &   \\ 
\hline
& \rsinglesrc{} &  & \colorfield{\wrapn{51.0}{0.0}}{white}{black} & \colorfield{\wrapn{48.8}{0.2}}{white}{black} & \colorfield{\wrapn{46.9}{0.1}}{white}{black} & \colorfield{\wrapn{44.9}{0.1}}{white}{black} & \colorfield{\wrapn{43.0}{0.1}}{white}{black} & \colorfield{\wrapn{41.0}{0.1}}{white}{black} & \colorfield{\wrapn{39.4}{0.1}}{white}{black} & \colorfield{\wrapn{37.3}{0.1}}{white}{black} & \colorfield{\wrapn{35.8}{0.0}}{white}{black} &   \\ 
\hline
 \dewercol{} 
 & \wer{0~\%} & \colorfield{\wrapn{50.3}{0.0}}{white}{black} & \colorfield{\wrapn{50.8}{0.0}}{colorone}{colorlow} & \colorfield{\wrapn{49.5}{0.0}}{colortwo}{colorlow} & \colorfield{\wrapn{48.0}{0.1}}{colortwo}{colorlow} & \colorfield{\wrapn{46.7}{0.1}}{colortwo}{colorlow} & \colorfield{\wrapn{45.3}{0.2}}{colortwo}{colorlow} & \colorfield{\wrapn{43.8}{0.3}}{colortwo}{colorlow} & \colorfield{\wrapn{42.6}{0.2}}{colortwo}{colorlow} & \colorfield{\wrapn{41.0}{0.1}}{colortwo}{colorlow} & \colorfield{\wrapn{39.7}{0.1}}{colortwo}{colorlow} &   \\
 & \wer{5~\%} & \colorfield{\wrapn{48.3}{0.1}}{white}{black} & \colorfield{\wrapn{50.0}{0.0}}{colorone}{colorlow} & \colorfield{\wrapn{48.6}{0.0}}{colorone}{colorlow} & \colorfield{\wrapn{47.3}{0.0}}{colortwo}{colorlow} & \colorfield{\wrapn{45.7}{0.1}}{colortwo}{colorlow} & \colorfield{\wrapn{44.5}{0.1}}{colortwo}{colorlow} & \colorfield{\wrapn{43.1}{0.1}}{colortwo}{colorlow} & \colorfield{\wrapn{41.8}{0.0}}{colortwo}{colorlow} & \colorfield{\wrapn{40.1}{0.1}}{colortwo}{colorlow} & \colorfield{\wrapn{38.8}{0.1}}{colortwo}{colorlow} &   \\
 & \wer{10~\%} & \colorfield{\wrapn{46.5}{0.2}}{white}{black} & \colorfield{\wrapn{49.2}{0.1}}{colorone}{colorlow} & \colorfield{\wrapn{47.9}{0.2}}{colorone}{colorlow} & \colorfield{\wrapn{46.4}{0.1}}{colorone}{colorlow} & \colorfield{\wrapn{44.8}{0.1}}{colortwo}{colorlow} & \colorfield{\wrapn{43.8}{0.0}}{colortwo}{colorlow} & \colorfield{\wrapn{42.3}{0.0}}{colortwo}{colorlow} & \colorfield{\wrapn{40.9}{0.1}}{colortwo}{colorlow} & \colorfield{\wrapn{39.2}{0.2}}{colortwo}{colorlow} & \colorfield{\wrapn{38.0}{0.1}}{colortwo}{colorlow} &   \\
 & \wer{15~\%} & \colorfield{\wrapn{44.7}{0.2}}{white}{black} & \colorfield{\wrapn{48.1}{0.1}}{colorone}{colorlow} & \colorfield{\wrapn{46.7}{0.1}}{colorone}{colorlow} & \colorfield{\wrapn{45.4}{0.1}}{colorone}{colorlow} & \colorfield{\wrapn{43.9}{0.1}}{colorone}{colorlow} & \colorfield{\wrapn{42.8}{0.0}}{colortwo}{colorlow} & \colorfield{\wrapn{41.2}{0.0}}{colortwo}{colorlow} & \colorfield{\wrapn{40.1}{0.1}}{colortwo}{colorlow} & \colorfield{\wrapn{38.4}{0.0}}{colortwo}{colorlow} & \colorfield{\wrapn{37.1}{0.0}}{colortwo}{colorlow} &   \\
 & \wer{20~\%} & \colorfield{\wrapn{42.9}{0.1}}{white}{black} & \colorfield{\wrapn{47.1}{0.0}}{colorone}{colorlow} & \colorfield{\wrapn{45.8}{0.1}}{colorone}{colorlow} & \colorfield{\wrapn{44.3}{0.0}}{colorone}{colorlow} & \colorfield{\wrapn{42.9}{0.1}}{colorone}{colorlow} & \colorfield{\wrapn{41.7}{0.1}}{colorone}{colorlow} & \colorfield{\wrapn{40.4}{0.1}}{colortwo}{colorlow} & \colorfield{\wrapn{39.1}{0.0}}{colortwo}{colorlow} & \colorfield{\wrapn{37.4}{0.1}}{colortwo}{colorlow} & \colorfield{\wrapn{36.3}{0.1}}{colortwo}{colorlow} &   \\
 & \wer{25~\%} & \colorfield{\wrapn{41.1}{0.1}}{white}{black} & \colorfield{\wrapn{46.1}{0.2}}{colorone}{colorlow} & \colorfield{\wrapn{44.8}{0.0}}{colorone}{colorlow} & \colorfield{\wrapn{43.6}{0.1}}{colorone}{colorlow} & \colorfield{\wrapn{42.0}{0.1}}{colorone}{colorlow} & \colorfield{\wrapn{40.8}{0.1}}{colorone}{colorlow} & \colorfield{\wrapn{39.3}{0.2}}{colortwo}{colorlow} & \colorfield{\wrapn{38.1}{0.1}}{colortwo}{colorlow} & \colorfield{\wrapn{36.4}{0.1}}{colortwo}{colorlow} & \colorfield{\wrapn{35.3}{0.1}}{colortwo}{colorlow} &   \\
 & \wer{30~\%} & \colorfield{\wrapn{39.4}{0.2}}{white}{black} & \colorfield{\wrapn{45.3}{0.3}}{colorone}{colorlow} & \colorfield{\wrapn{43.9}{0.2}}{colorone}{colorlow} & \colorfield{\wrapn{42.6}{0.2}}{colorone}{colorlow} & \colorfield{\wrapn{41.1}{0.1}}{colorone}{colorlow} & \colorfield{\wrapn{39.9}{0.2}}{colorone}{colorlow} & \colorfield{\wrapn{38.5}{0.2}}{colorone}{colorlow} & \colorfield{\wrapn{37.3}{0.1}}{colortwo}{colorlow} & \colorfield{\wrapn{35.7}{0.0}}{colortwo}{colorlow} & \colorfield{\wrapn{34.5}{0.2}}{colortwo}{colorlow} &   \\
 & \wer{35~\%} & \colorfield{\wrapn{38.0}{0.2}}{white}{black} & \colorfield{\wrapn{44.3}{0.2}}{colorone}{colorlow} & \colorfield{\wrapn{42.9}{0.3}}{colorone}{colorlow} & \colorfield{\wrapn{41.5}{0.1}}{colorone}{colorlow} & \colorfield{\wrapn{40.2}{0.2}}{colorone}{colorlow} & \colorfield{\wrapn{38.9}{0.2}}{colorone}{colorlow} & \colorfield{\wrapn{37.6}{0.1}}{colorone}{colorlow} & \colorfield{\wrapn{36.4}{0.1}}{colorone}{colorlow} & \colorfield{\wrapn{34.9}{0.0}}{colortwo}{colorlow} & \colorfield{\wrapn{33.7}{0.2}}{colortwo}{colorlow} &   \\
 & \wer{40~\%} & \colorfield{\wrapn{36.2}{0.2}}{white}{black} & \colorfield{\wrapn{43.2}{0.2}}{colorone}{colorlow} & \colorfield{\wrapn{41.9}{0.2}}{colorone}{colorlow} & \colorfield{\wrapn{40.5}{0.2}}{colorone}{colorlow} & \colorfield{\wrapn{39.1}{0.1}}{colorone}{colorlow} & \colorfield{\wrapn{37.9}{0.1}}{colorone}{colorlow} & \colorfield{\wrapn{36.4}{0.2}}{colorone}{colorlow} & \colorfield{\wrapn{35.2}{0.1}}{colorone}{colorlow} & \colorfield{\wrapn{33.8}{0.2}}{colorone}{colorlow} & \colorfield{\wrapn{32.8}{0.2}}{colortwo}{colorlow} & 

%% file: newsmultisimulgrid.tex
\def\plotwidth{.20\linewidth}
News11 En part, En ref
\begin{tabular}{c@{}c@{}c@{}c@{}cccc}
\multicolumn{5}{c}{\textbf{En WER}} \\
 & 0\% & 10\% & 20\% & 30\%  \\
0\% & 
\includegraphics[width=\plotwidth,valign=m]{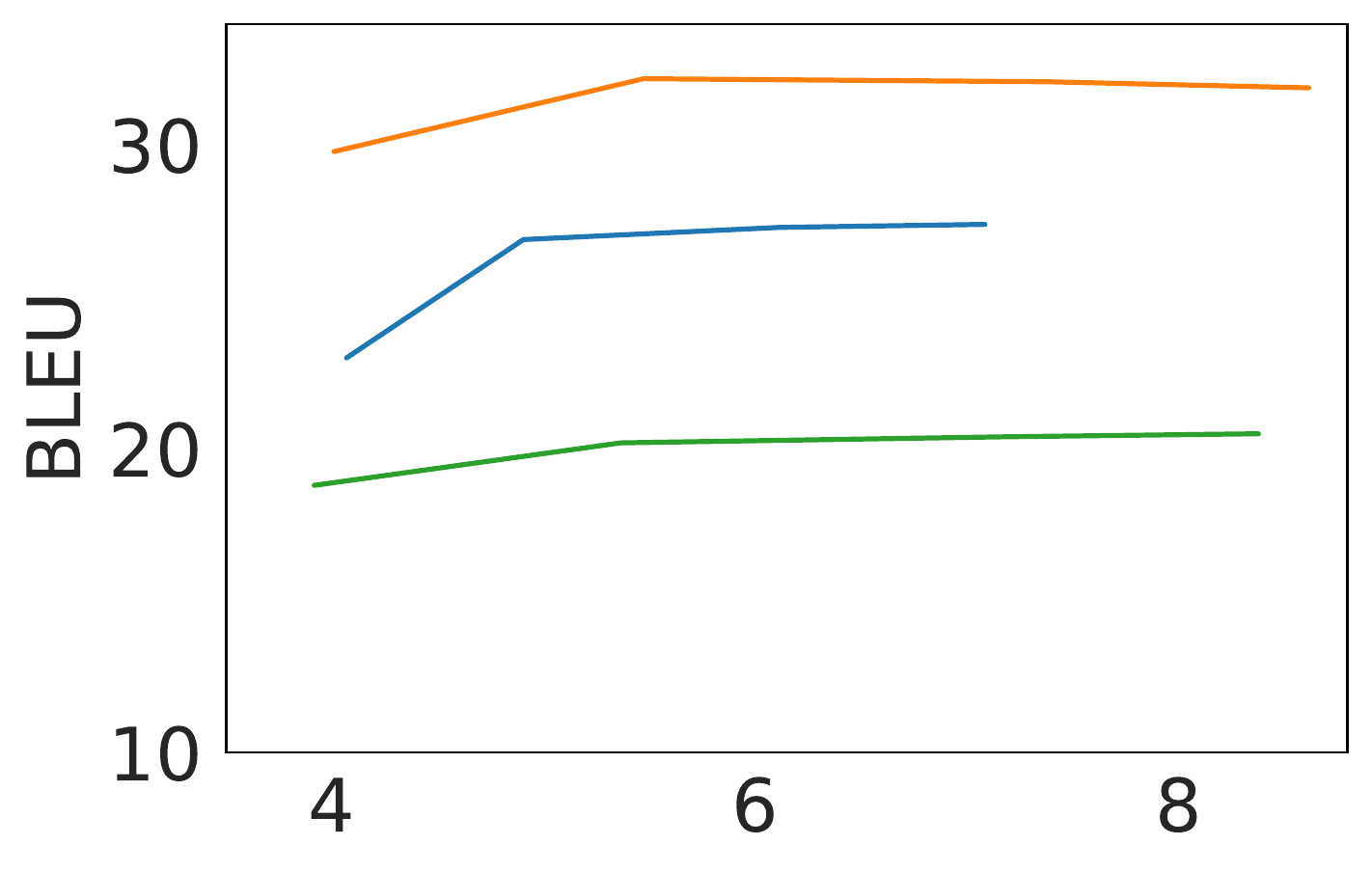} & 
\includegraphics[width=\plotwidth,valign=m]{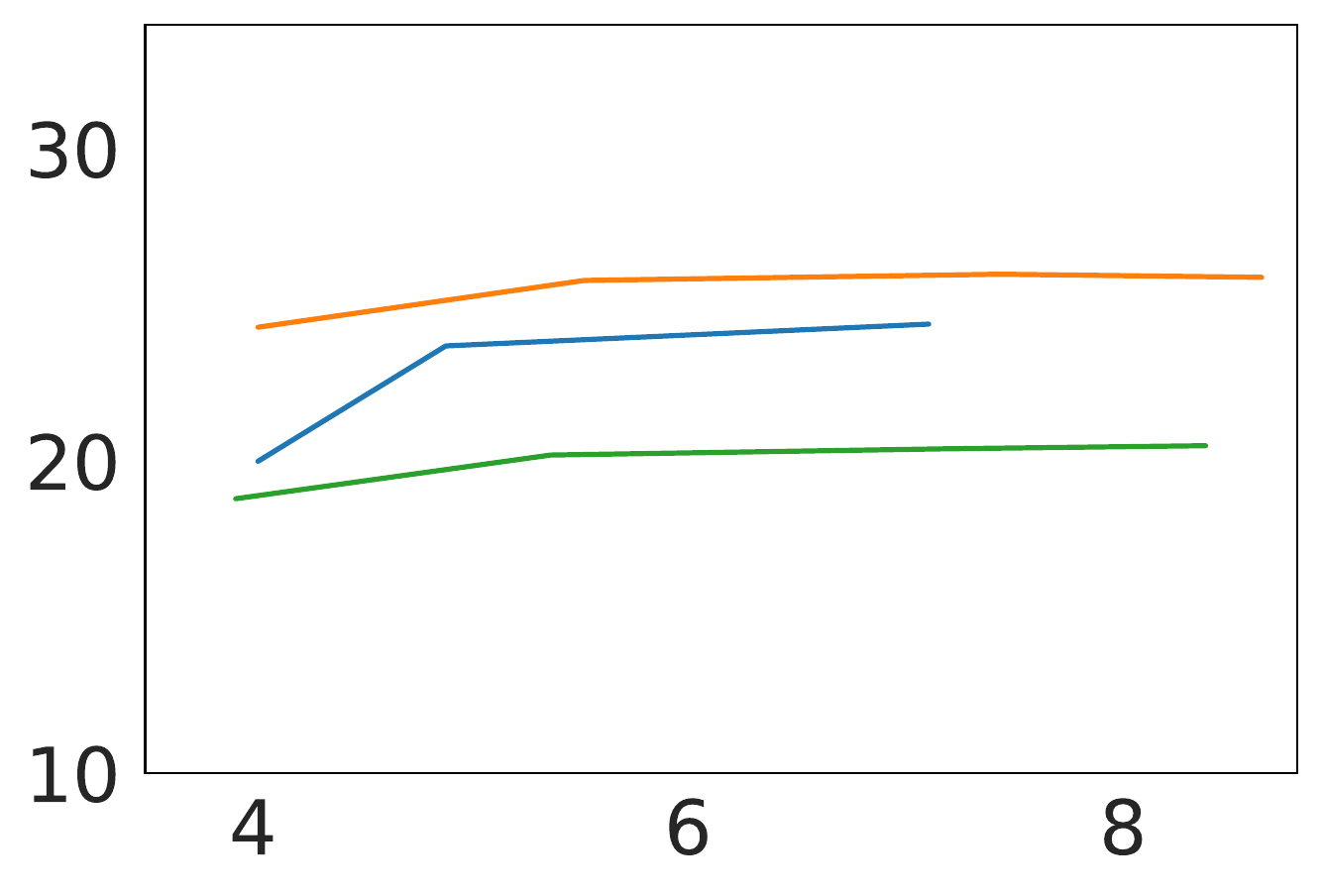} & 
\includegraphics[width=\plotwidth,valign=m]{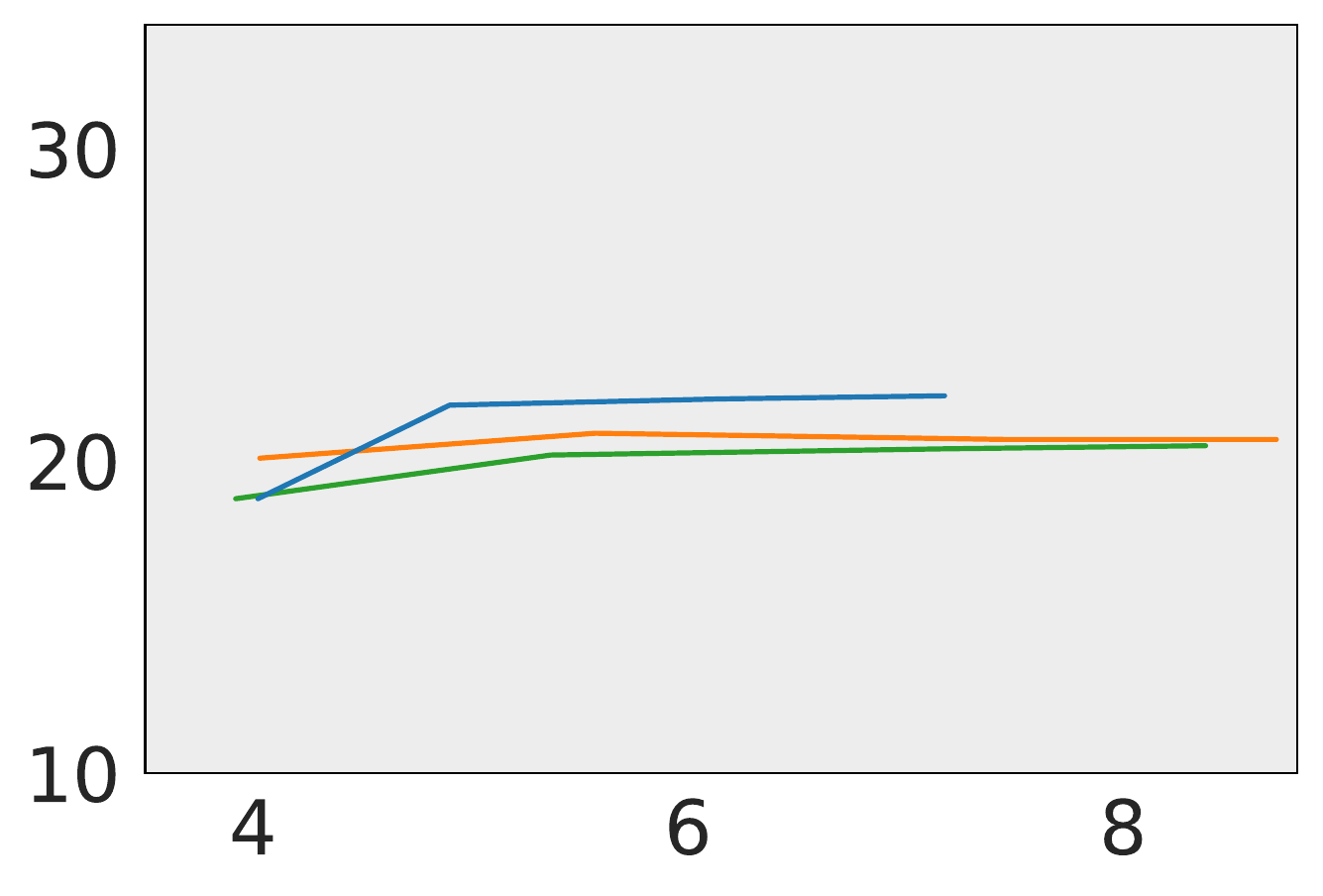} &
\includegraphics[width=\plotwidth,valign=m]{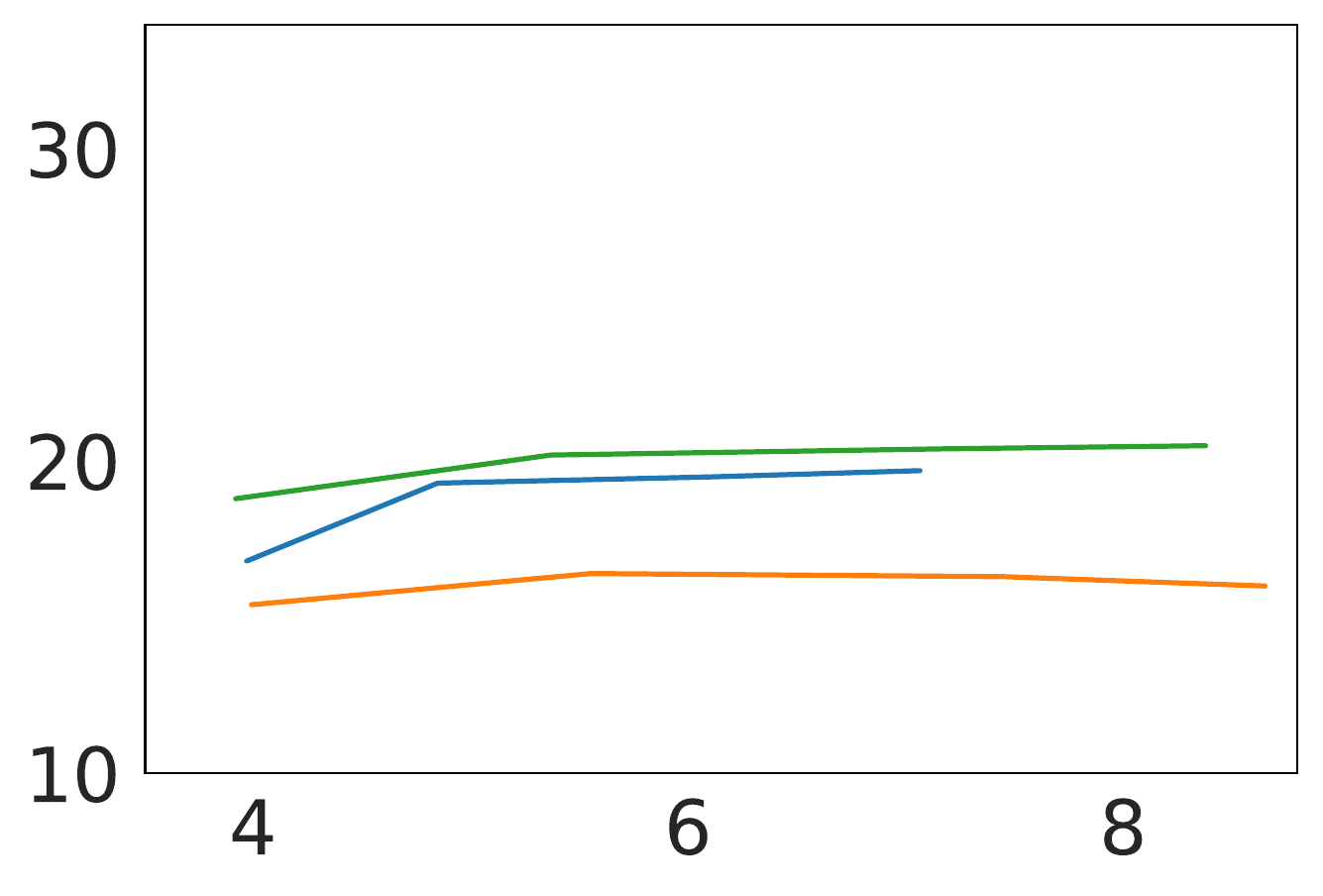} \\ 
                                                                                
\multirow{4}{*}{\rotatebox[origin=c]{90}{\textbf{De WER}}}                      
10\% &                                                                          
\includegraphics[width=\plotwidth,valign=m]{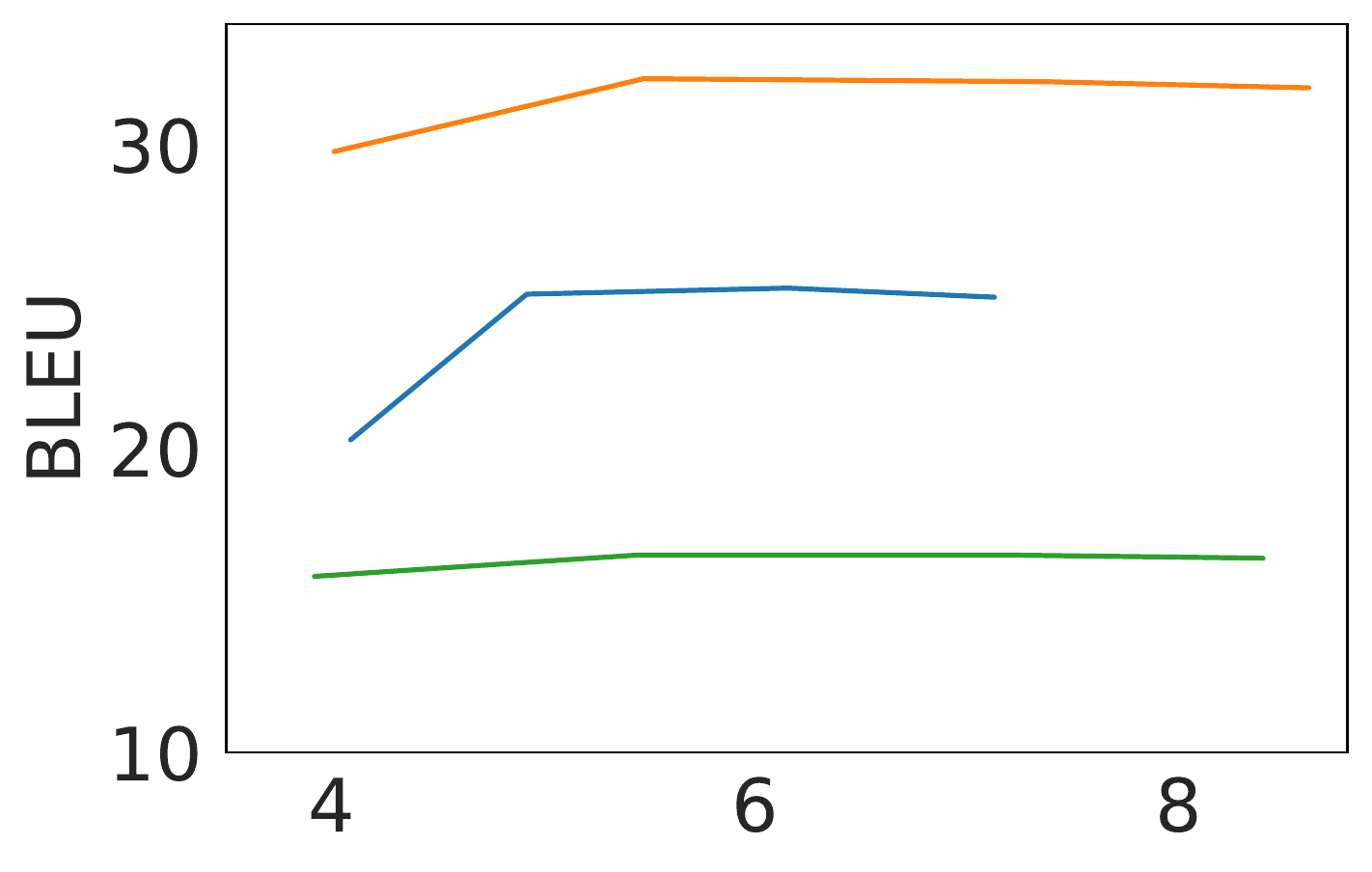} & 
\includegraphics[width=\plotwidth,valign=m]{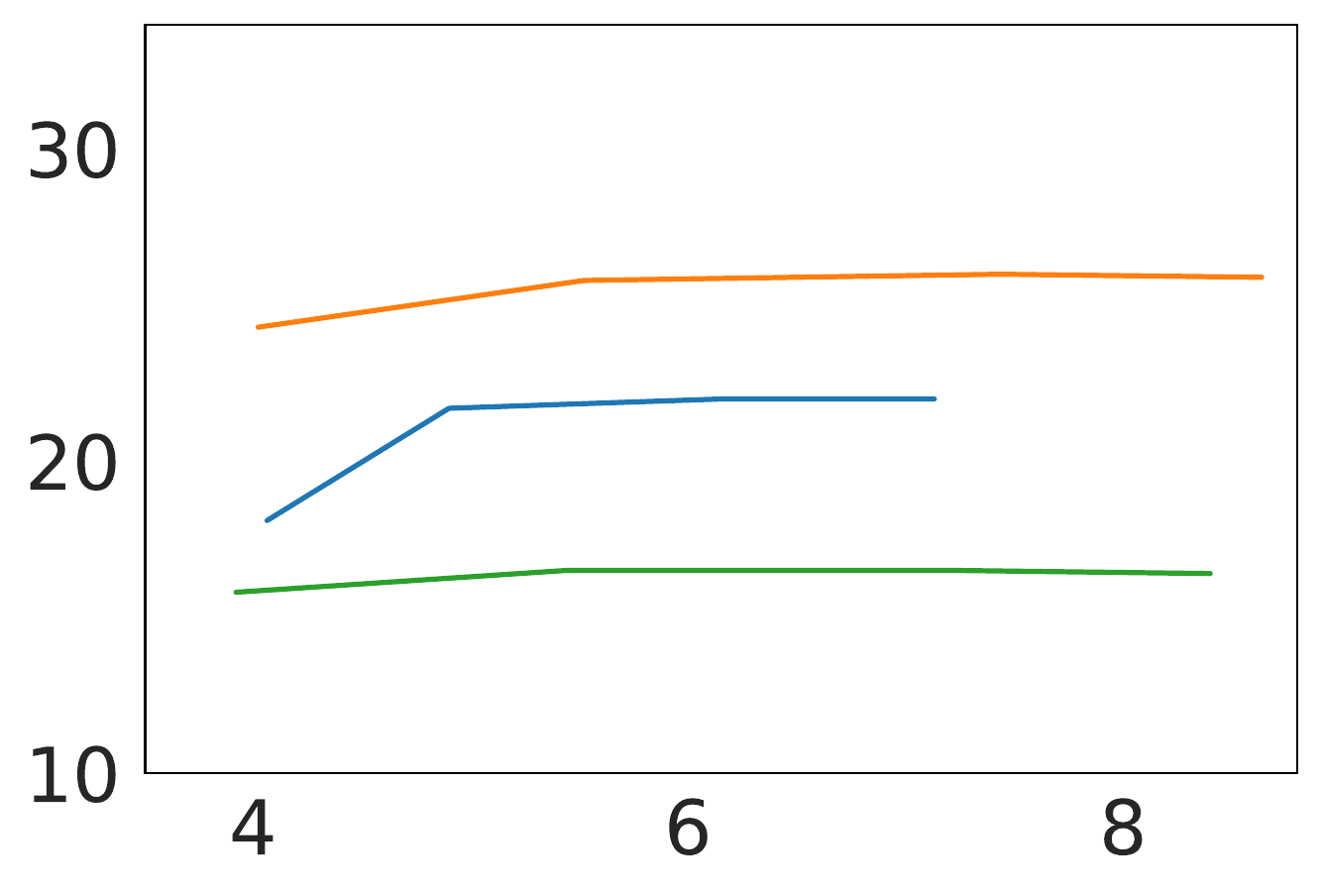} & 
\includegraphics[width=\plotwidth,valign=m]{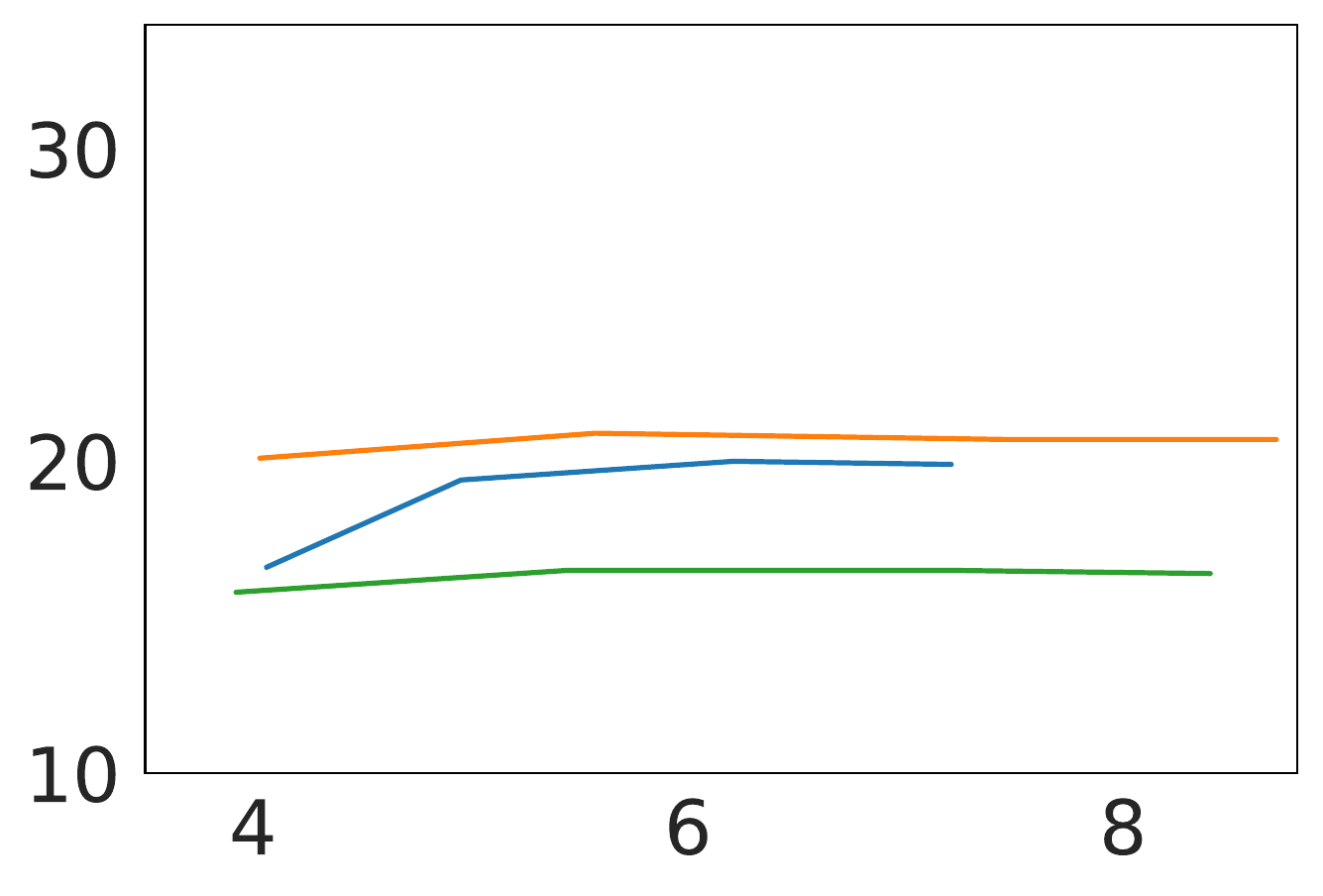} &
\includegraphics[width=\plotwidth,valign=m]{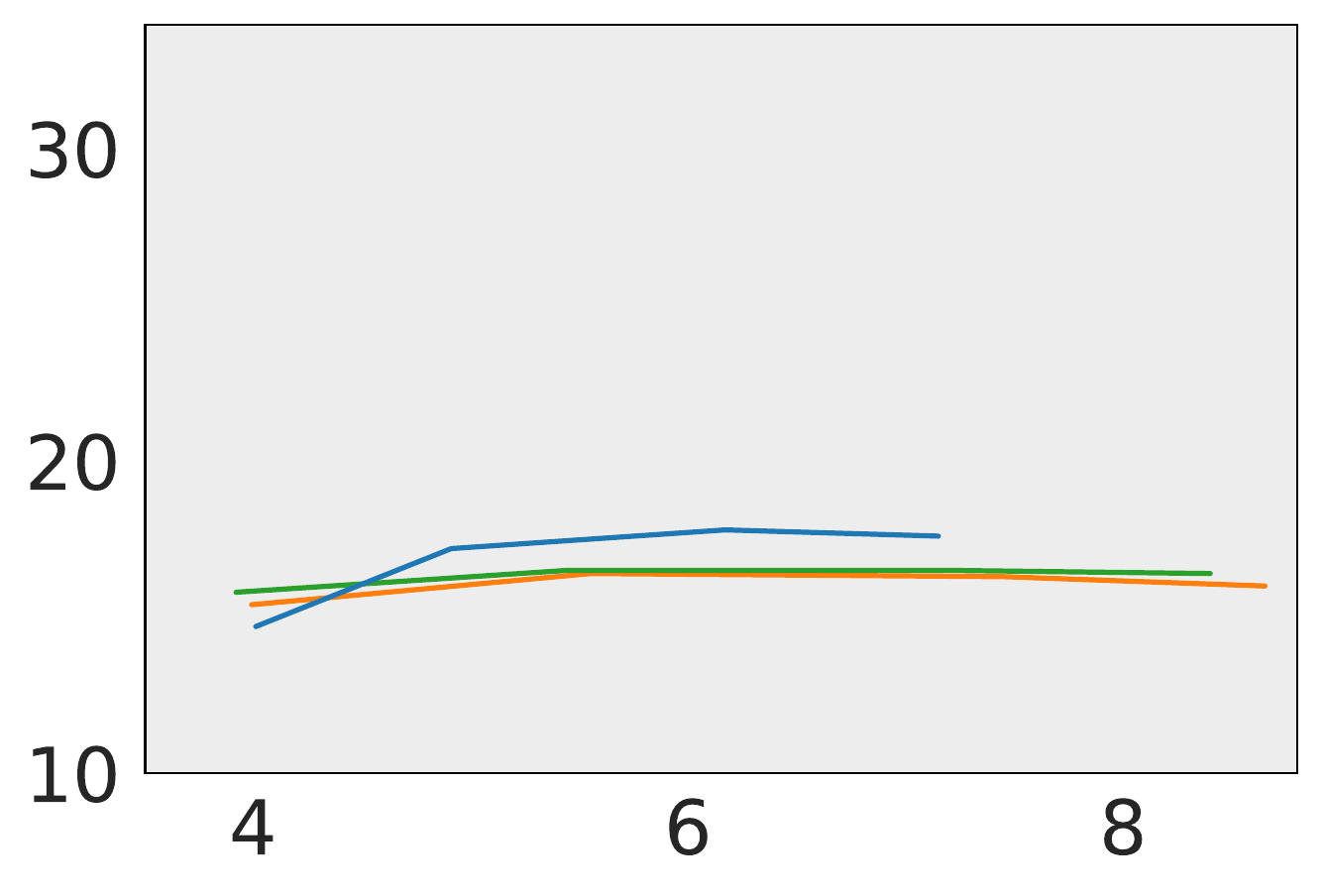} \\ 
                                                                                
20\% &                                                                          
\includegraphics[width=\plotwidth,valign=m]{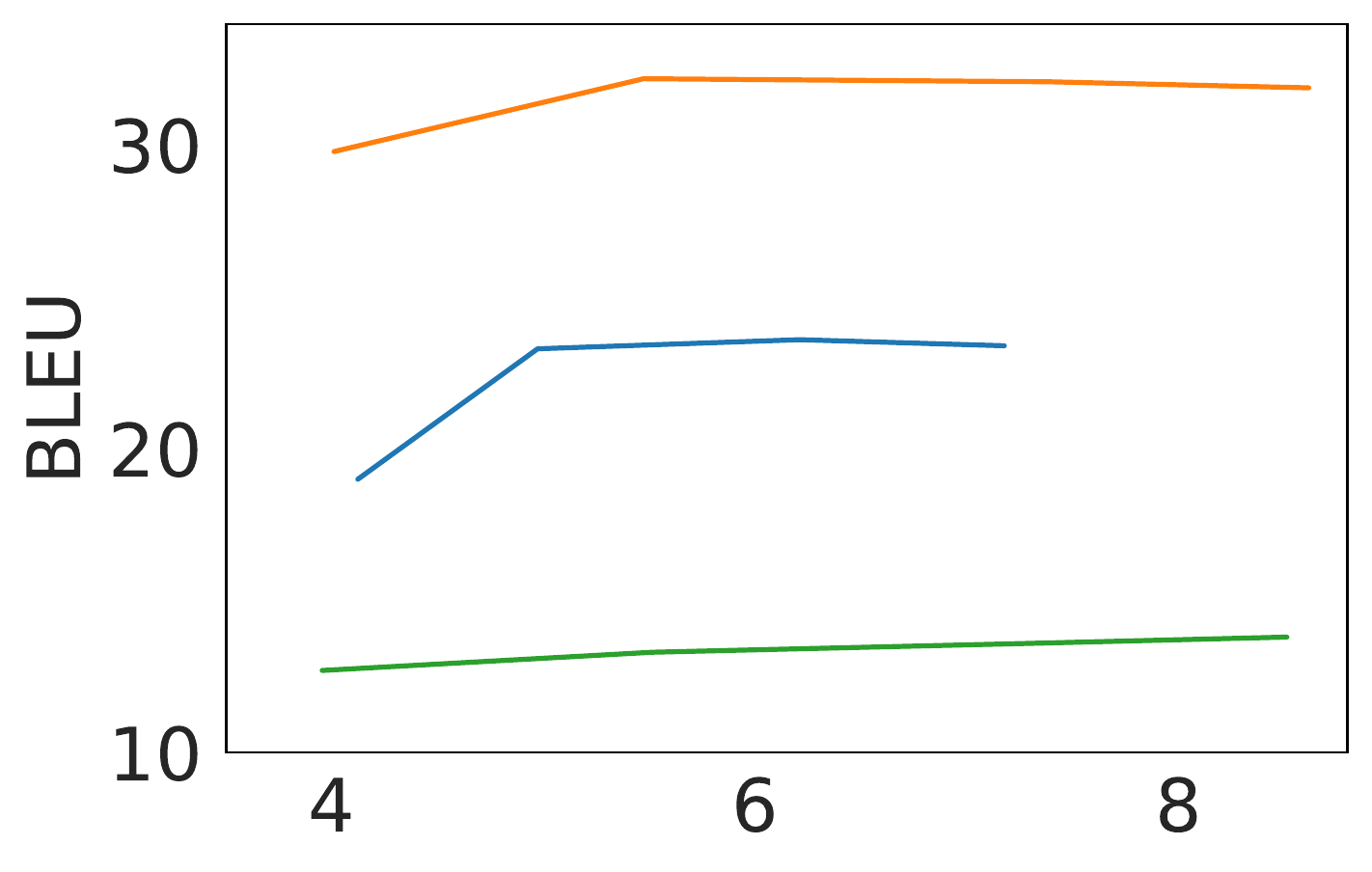} & 
\includegraphics[width=\plotwidth,valign=m]{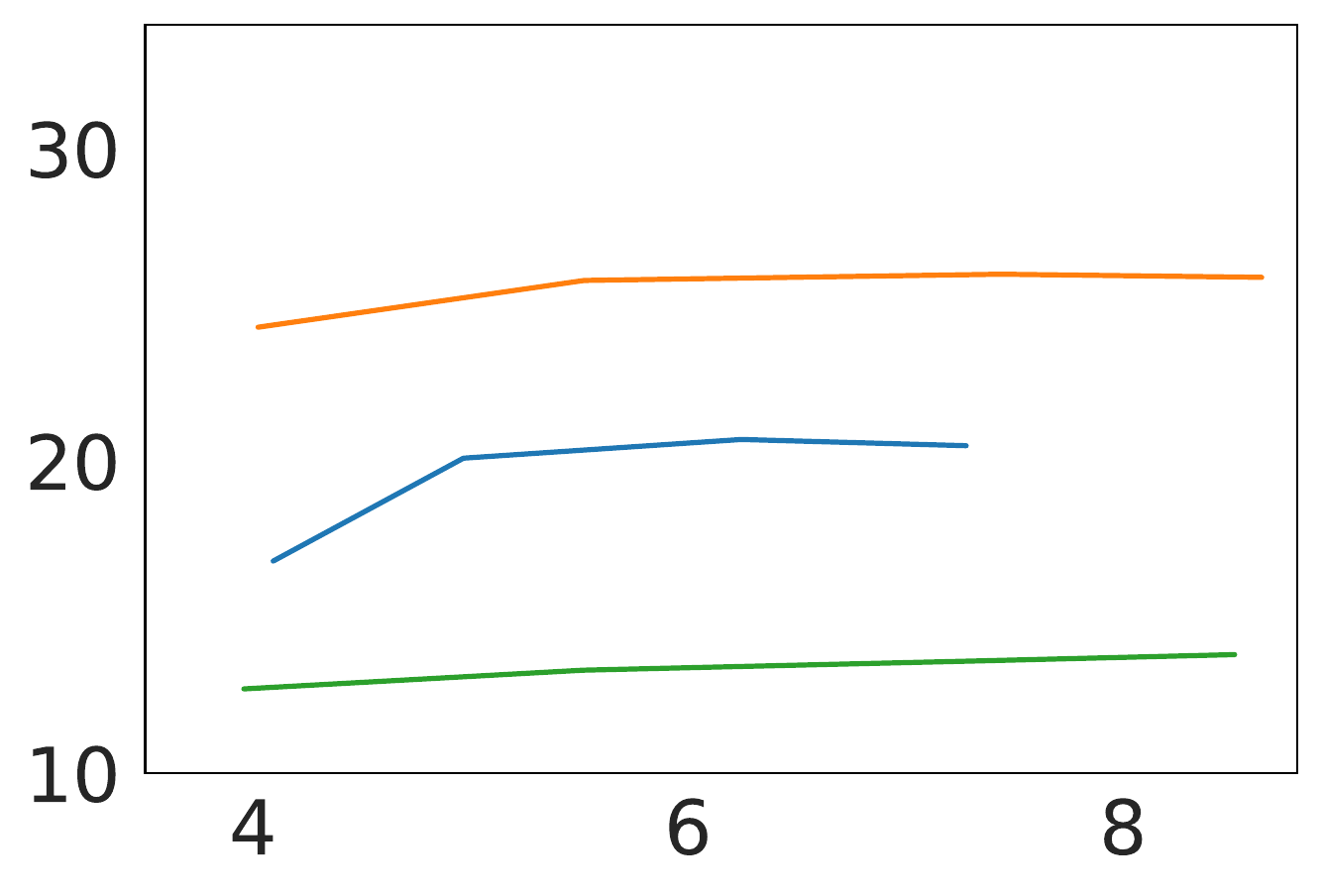} & 
\includegraphics[width=\plotwidth,valign=m]{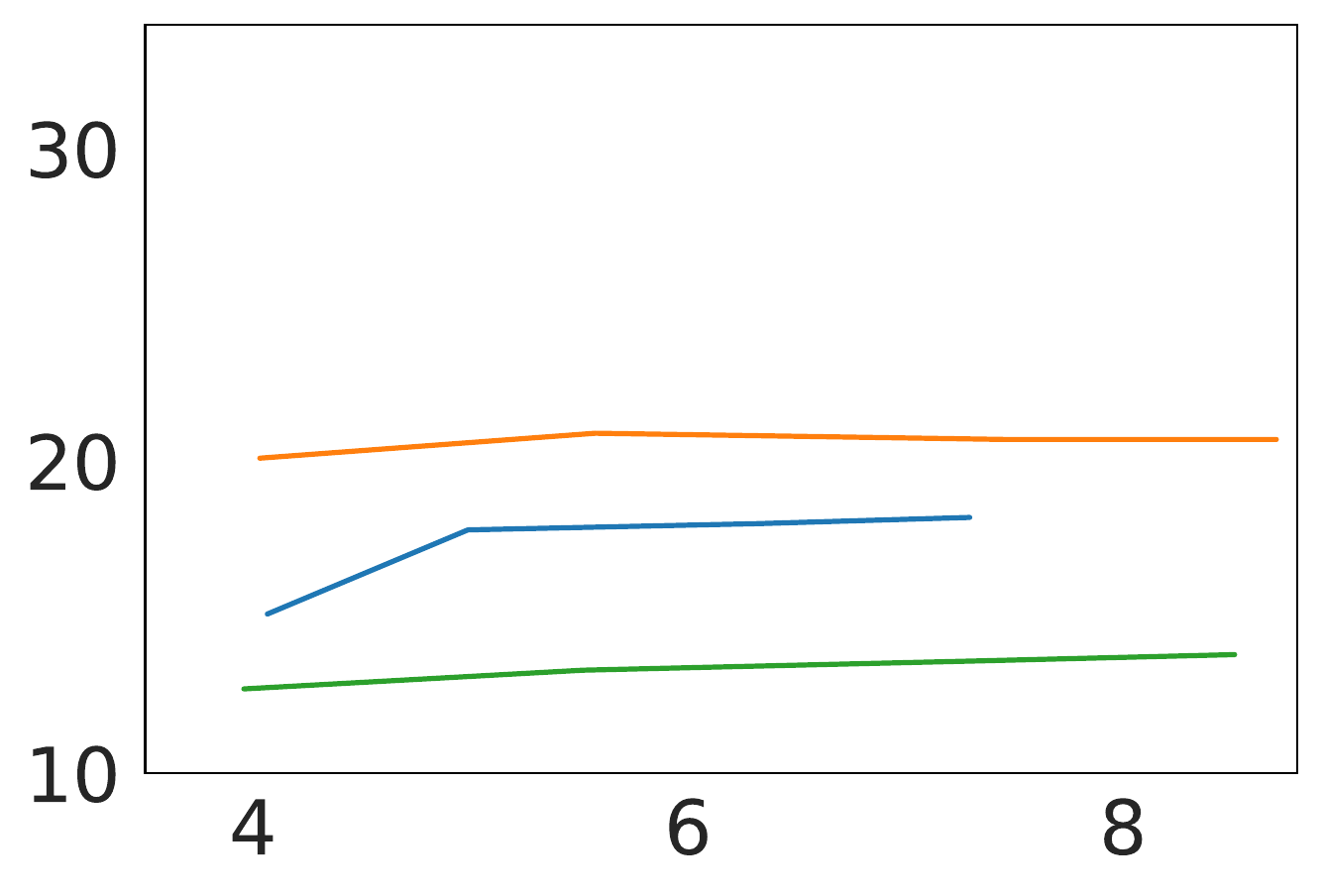} &
\includegraphics[width=\plotwidth,valign=m]{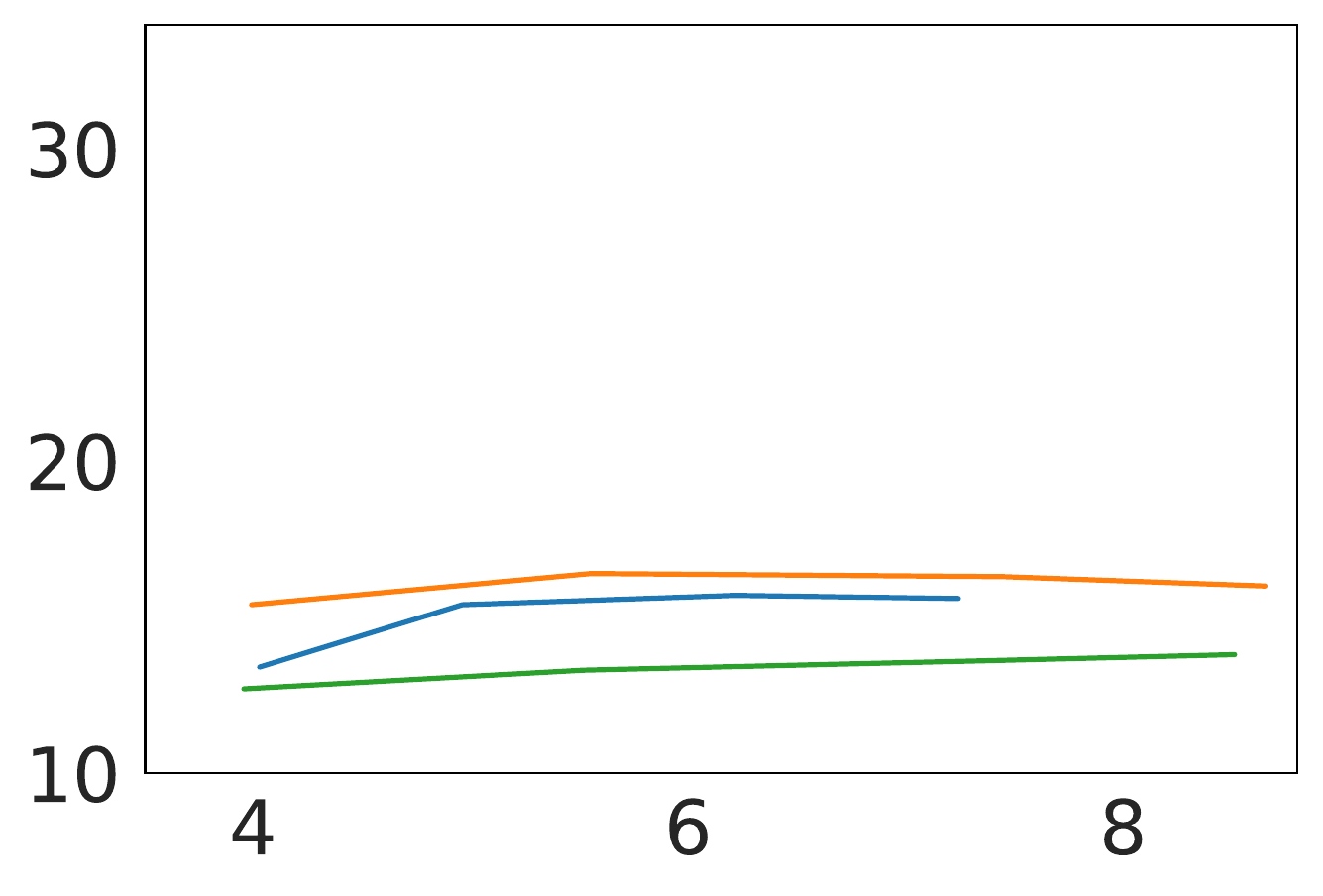} \\ 
                                                                                
30\% &                                                                          
\includegraphics[width=\plotwidth,valign=m]{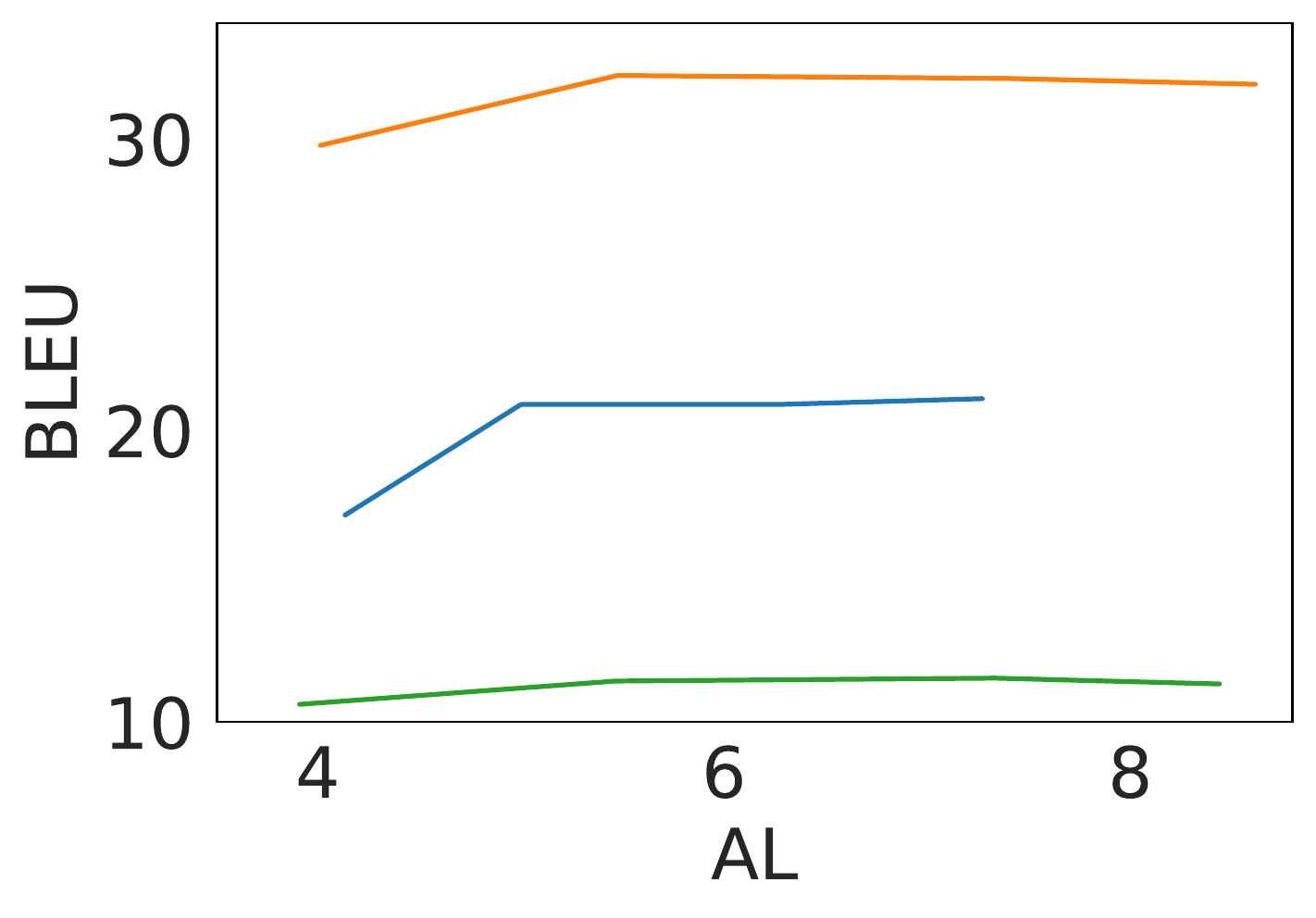} & 
\includegraphics[width=\plotwidth,valign=m]{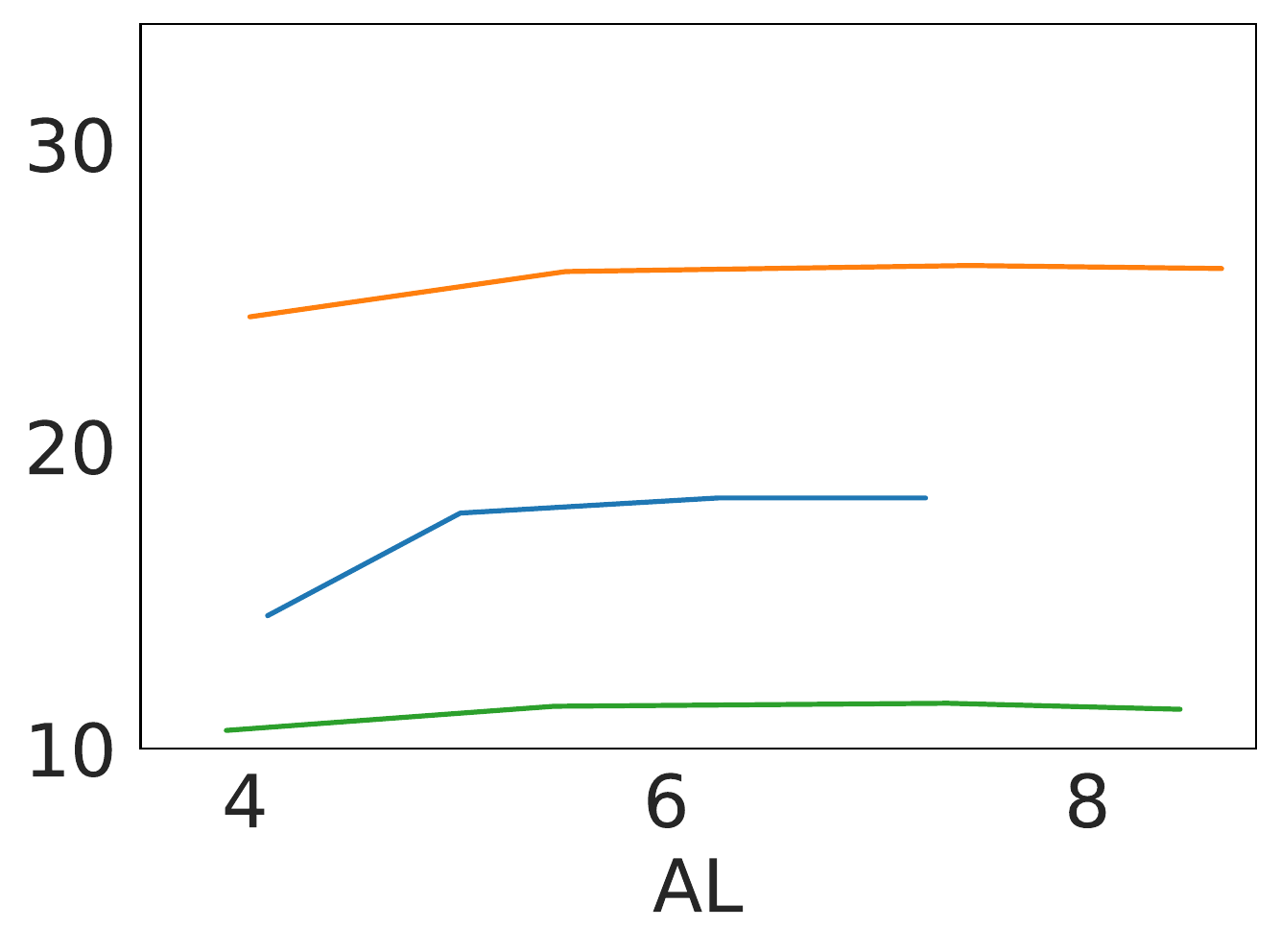} & 
\includegraphics[width=\plotwidth,valign=m]{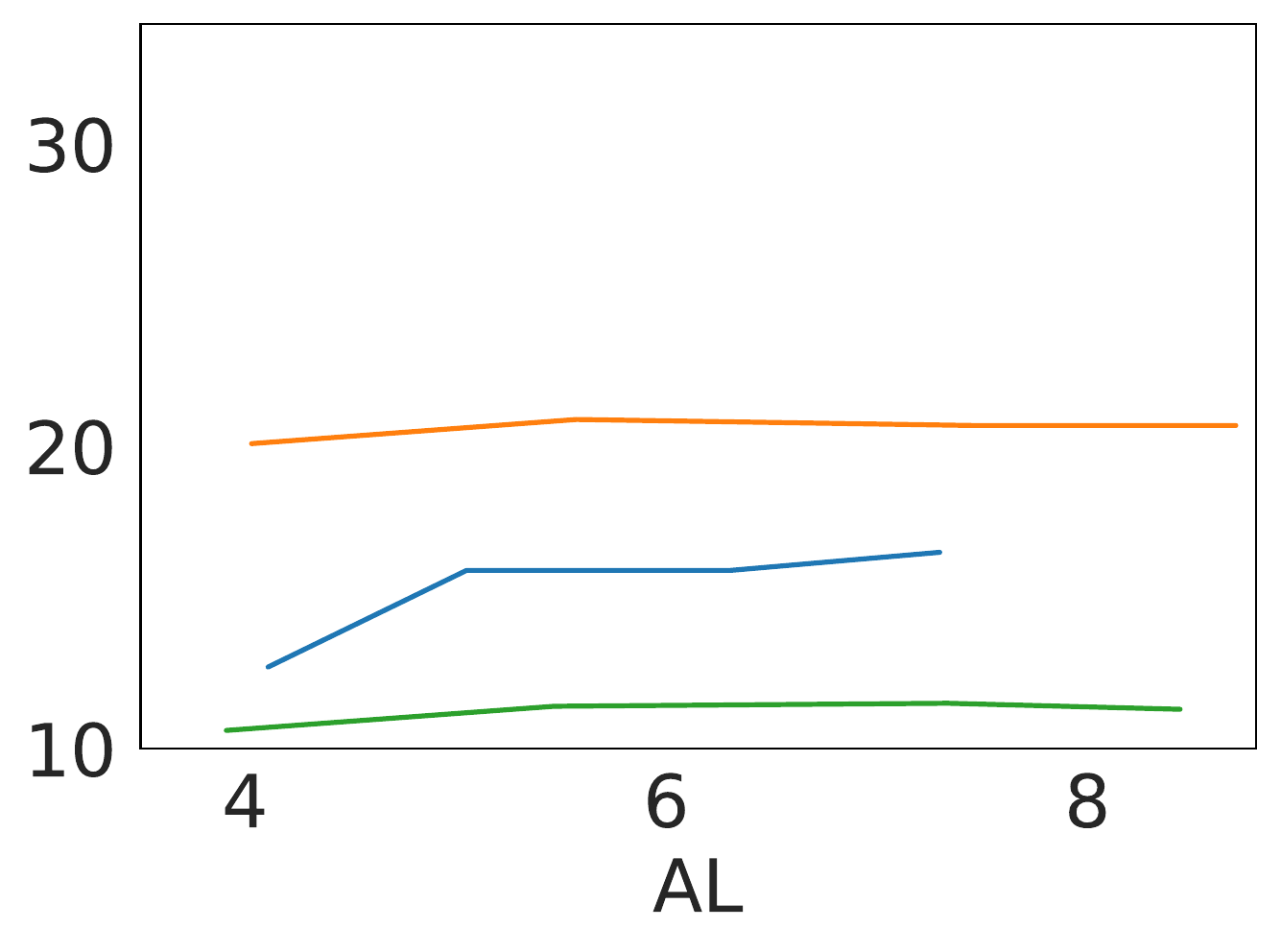} &
\includegraphics[width=\plotwidth,valign=m]{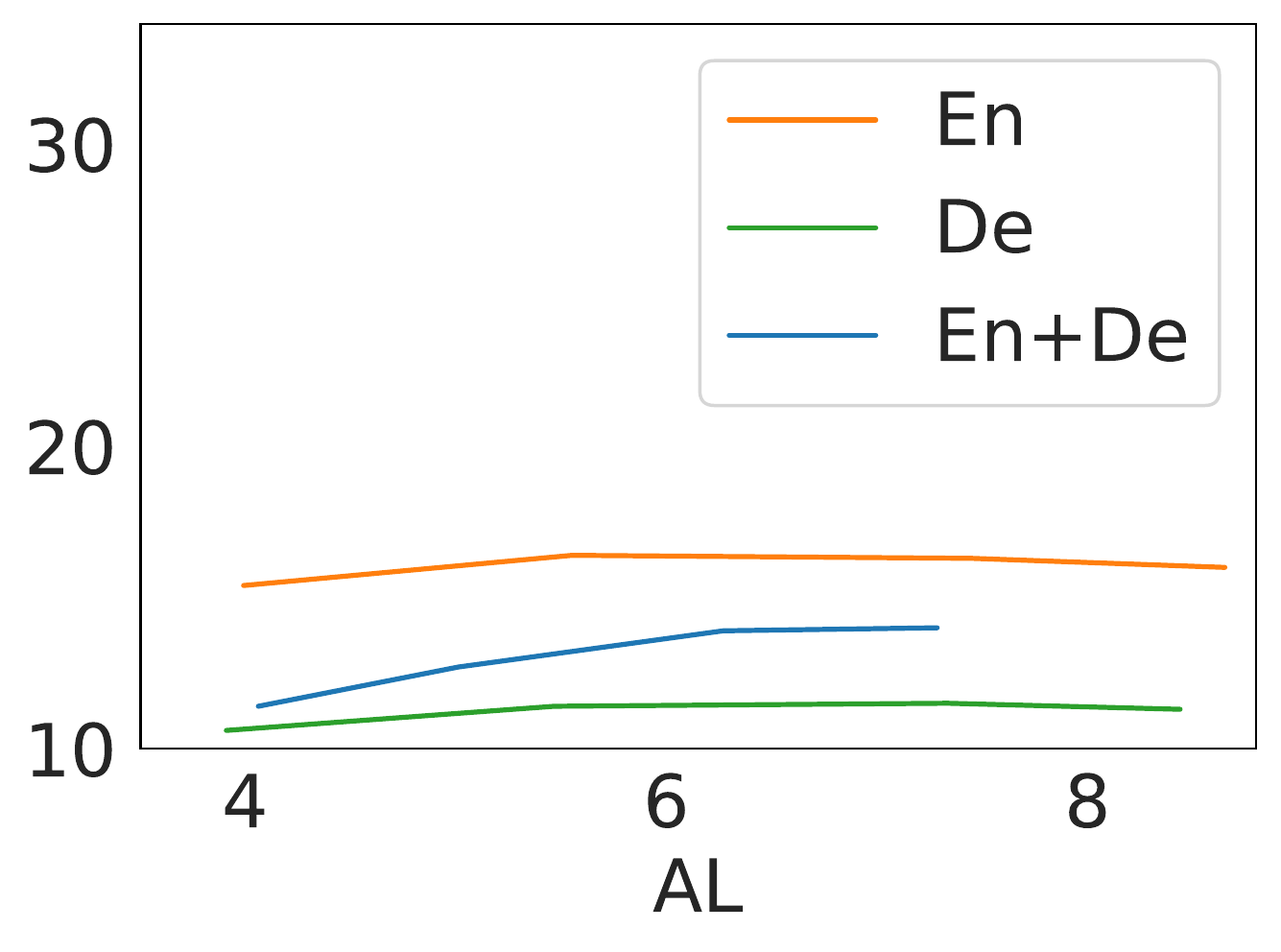} \\
\end{tabular}
News11~En~part,~De~refs
\begin{tabular}{c@{}c@{}c@{}c@{}cccc}
\multicolumn{5}{c}{\textbf{En WER}} \\
 & 0 \% & 10 \% & 20 \% & 30 \%  \\
0\% & 
\includegraphics[width=\plotwidth,valign=m]{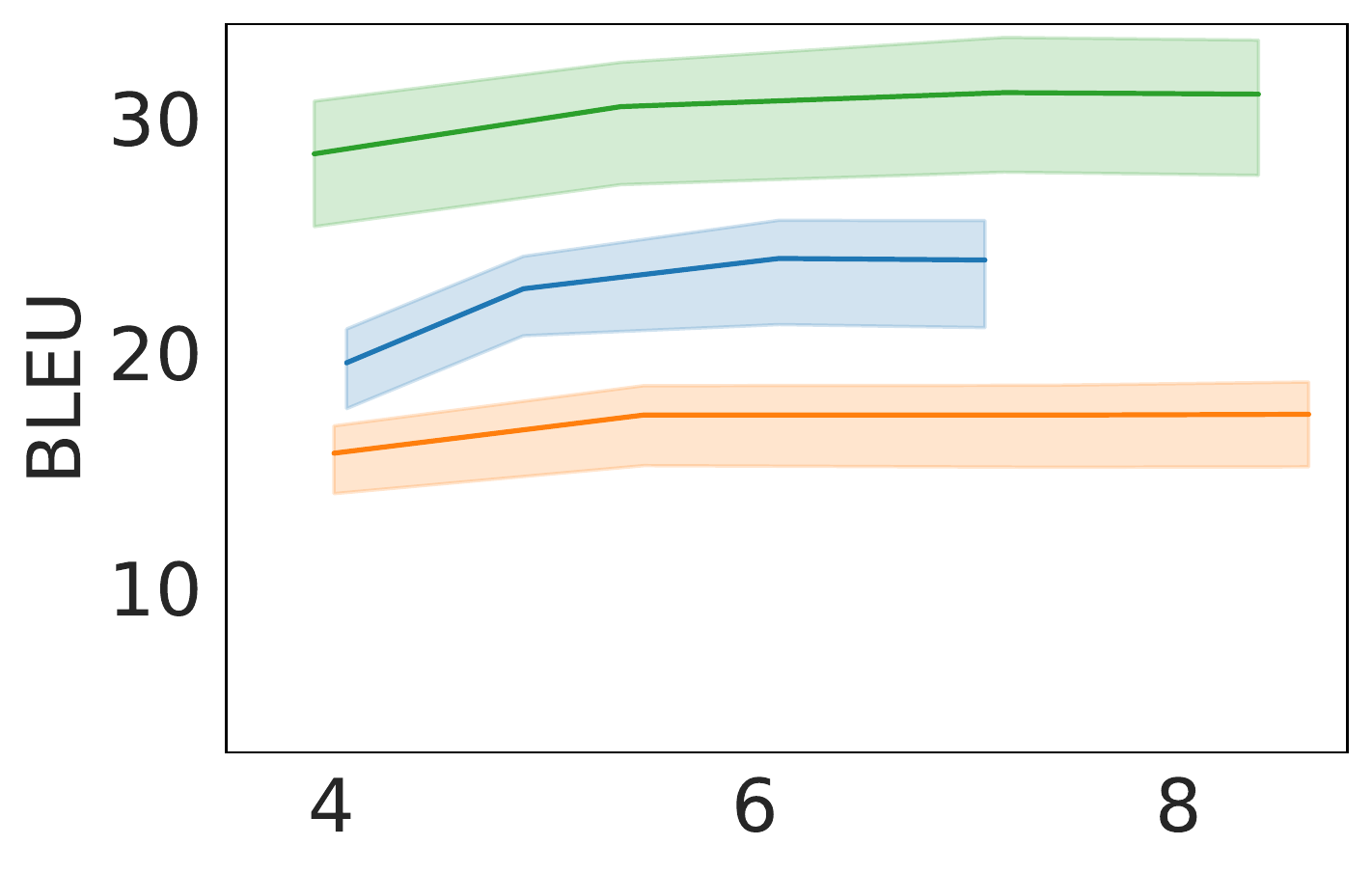} & 
\includegraphics[width=\plotwidth,valign=m]{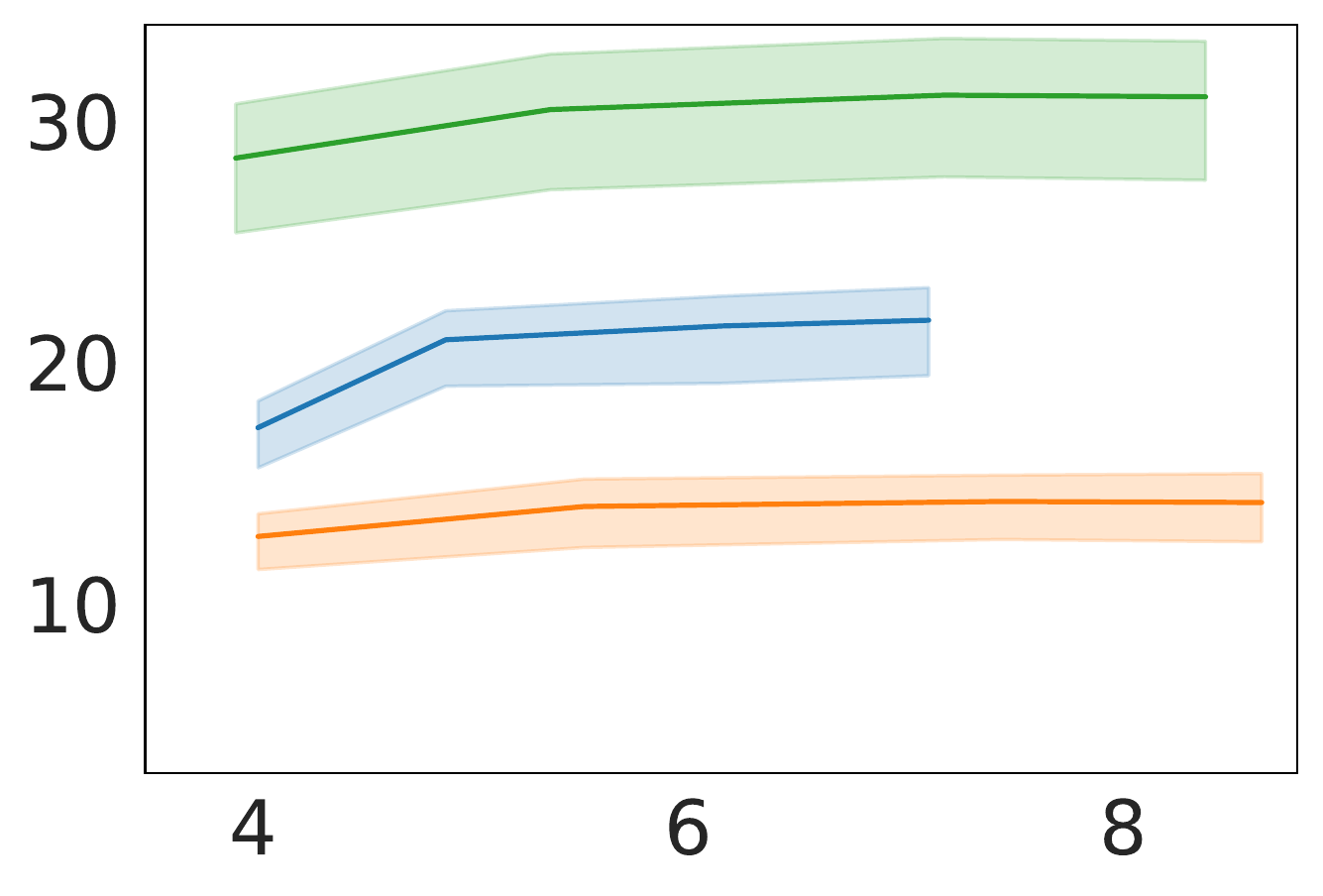} & 
\includegraphics[width=\plotwidth,valign=m]{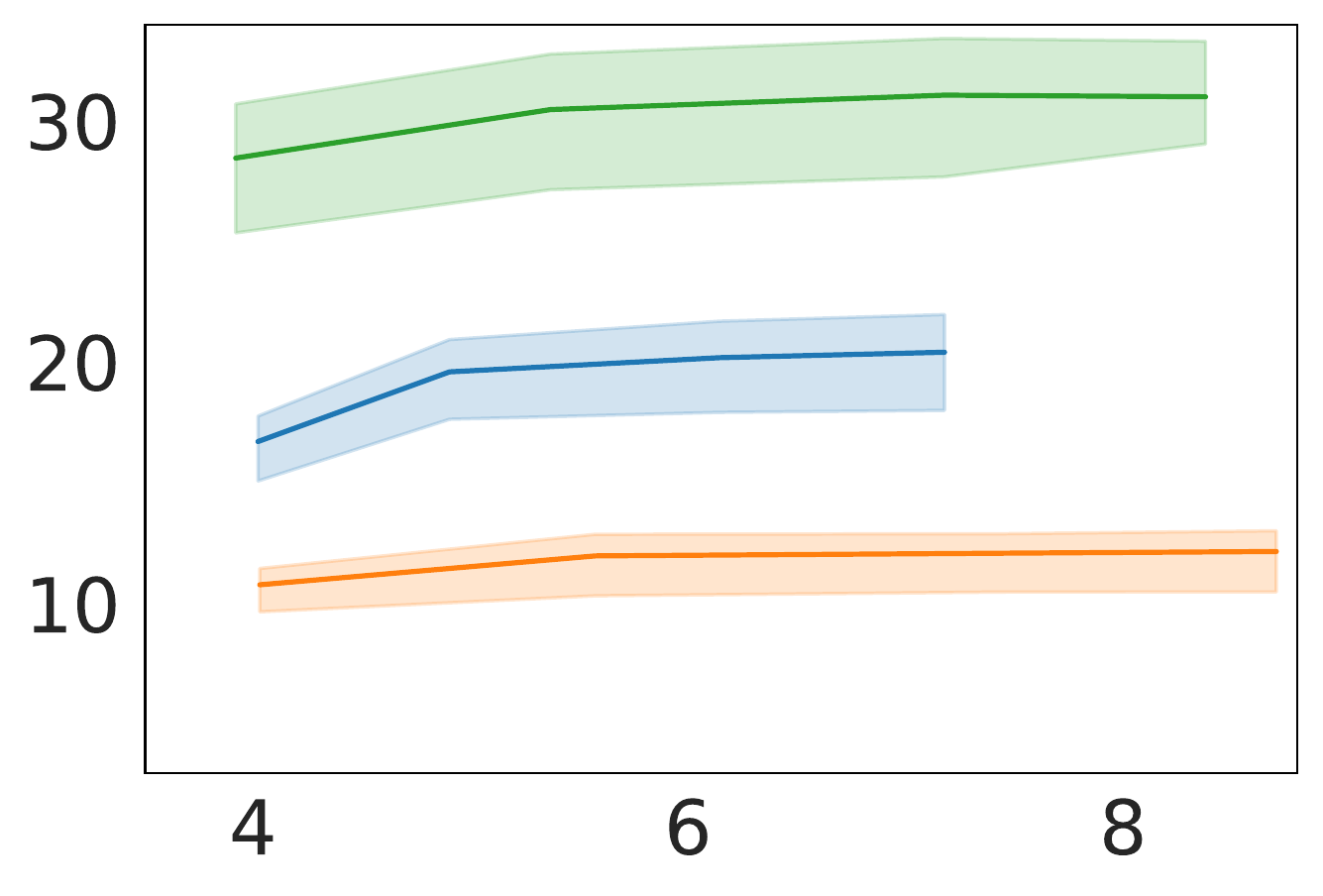} &
\includegraphics[width=\plotwidth,valign=m]{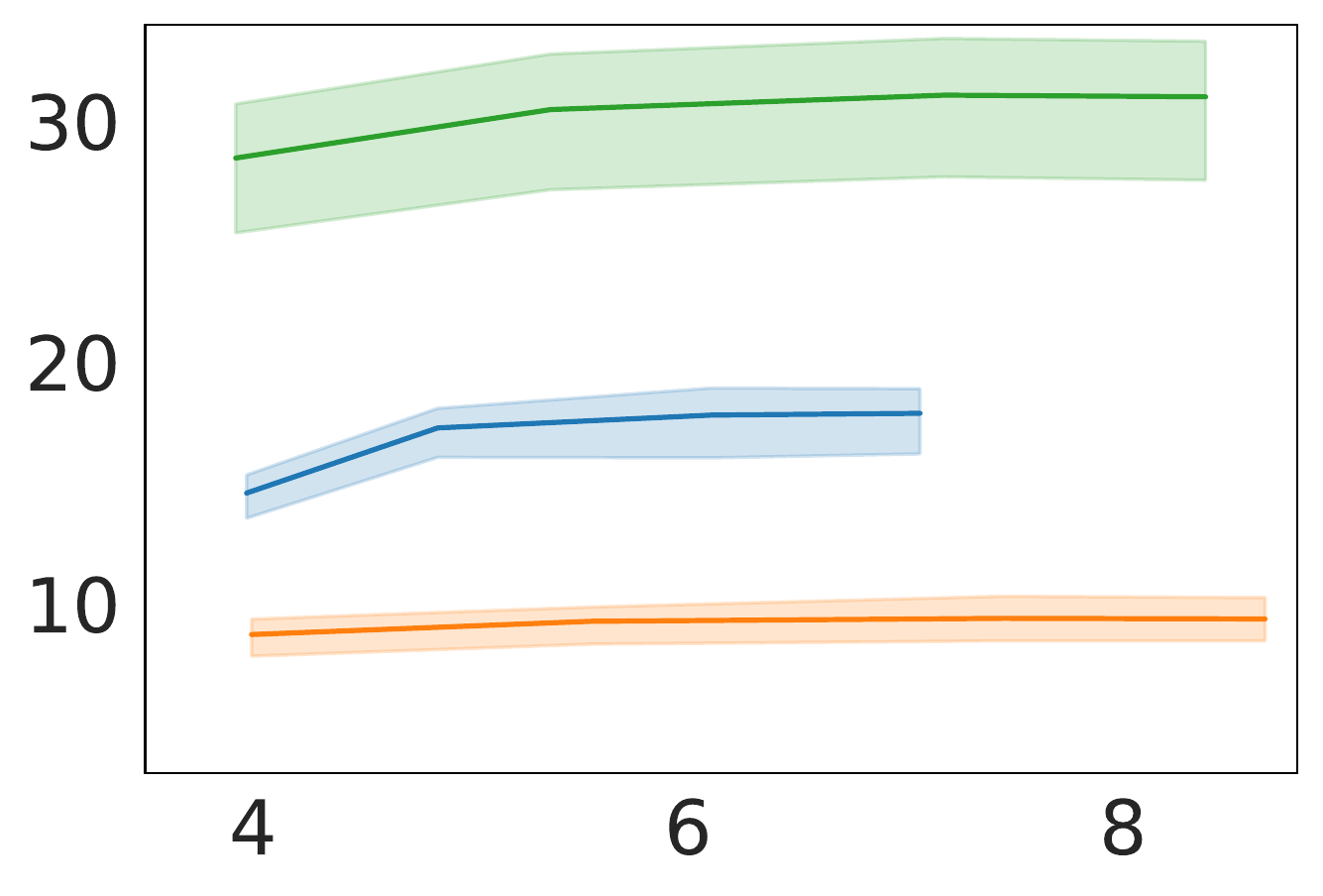} \\ 
\multirow{4}{*}{\rotatebox[origin=c]{90}{\textbf{De WER}}}%
10\% &                                                                         
\includegraphics[width=\plotwidth,valign=m]{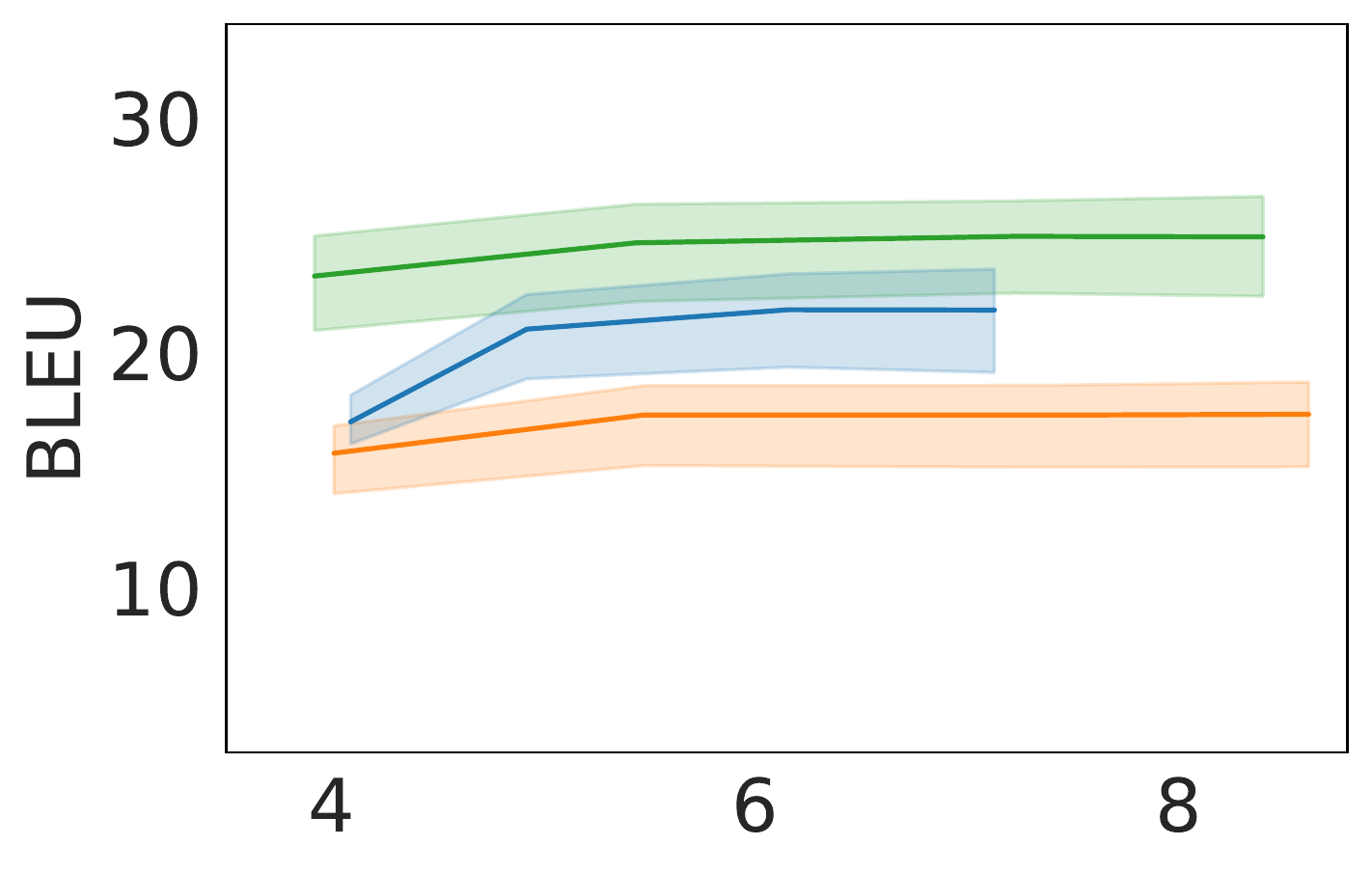} & 
\includegraphics[width=\plotwidth,valign=m]{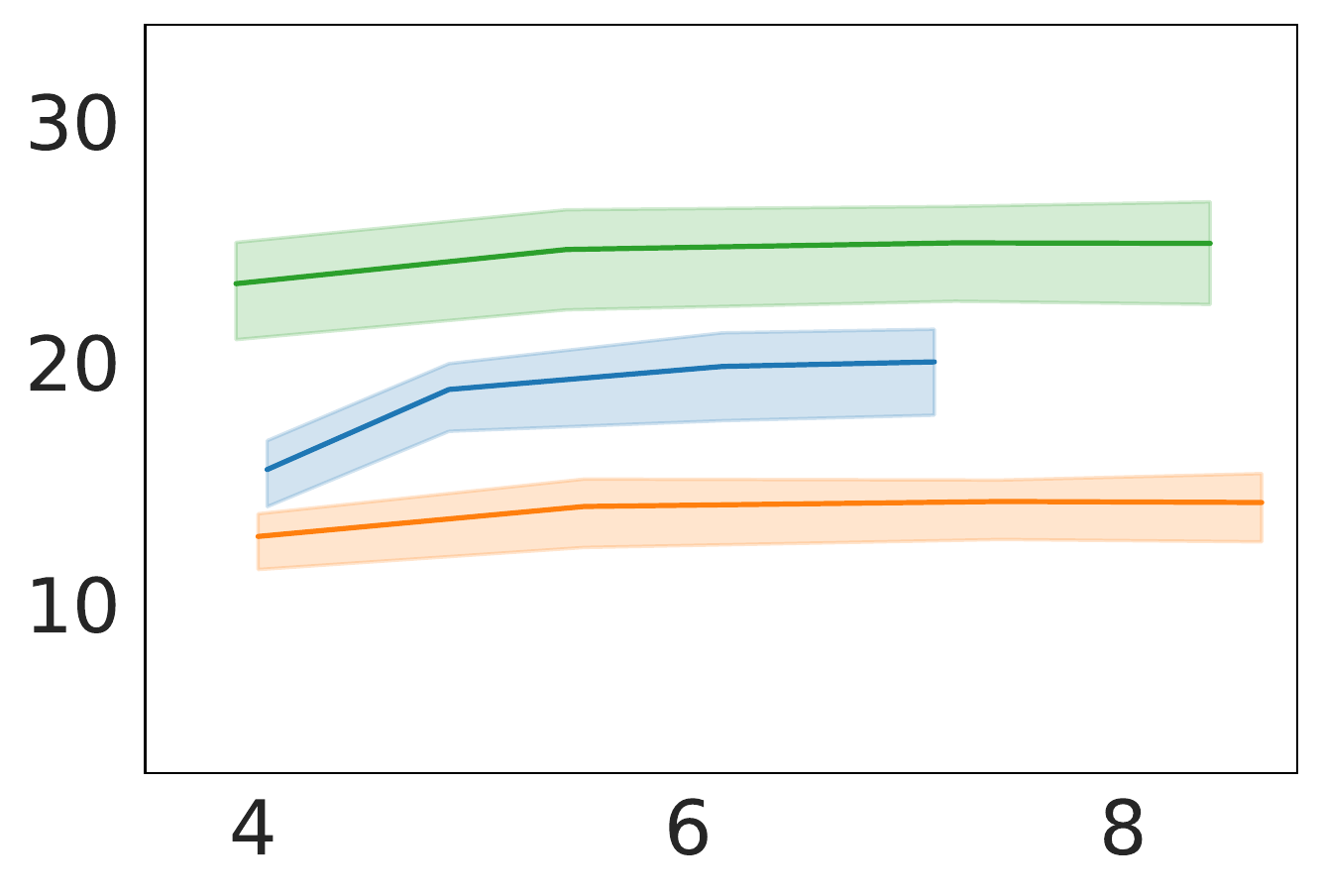} & 
\includegraphics[width=\plotwidth,valign=m]{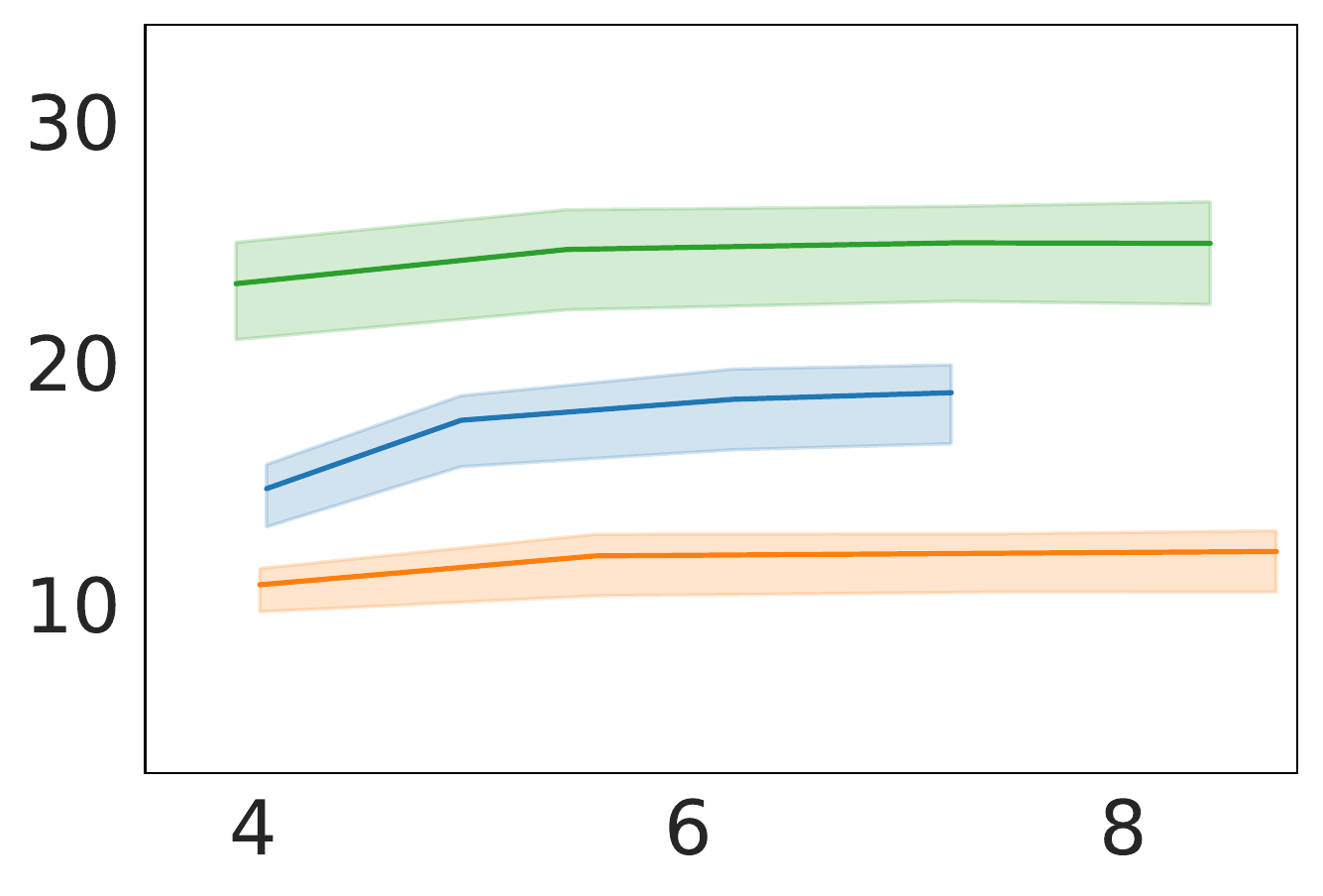} &
\includegraphics[width=\plotwidth,valign=m]{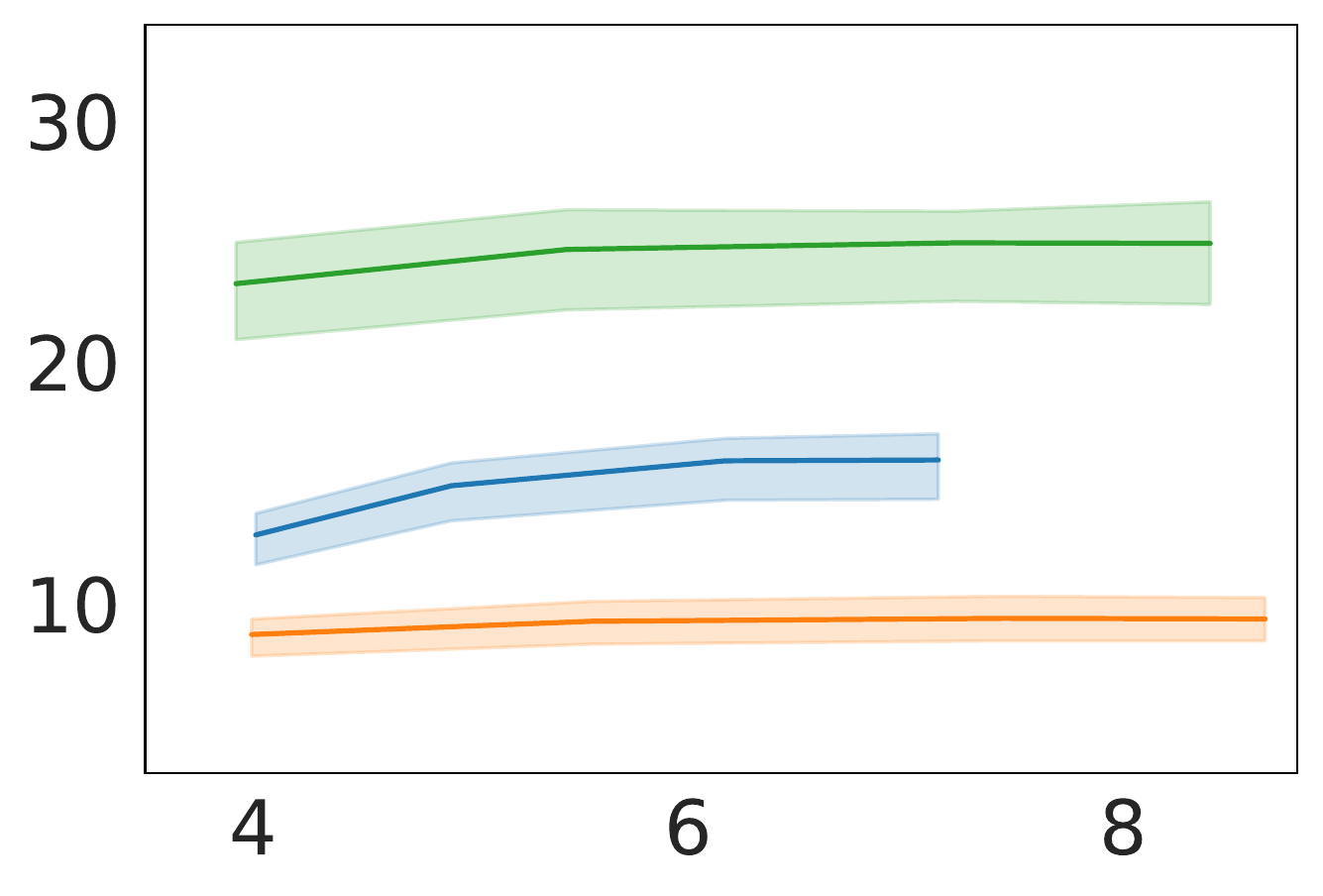} \\ 
                                                                               
20\% &                                                                         
\includegraphics[width=\plotwidth,valign=m]{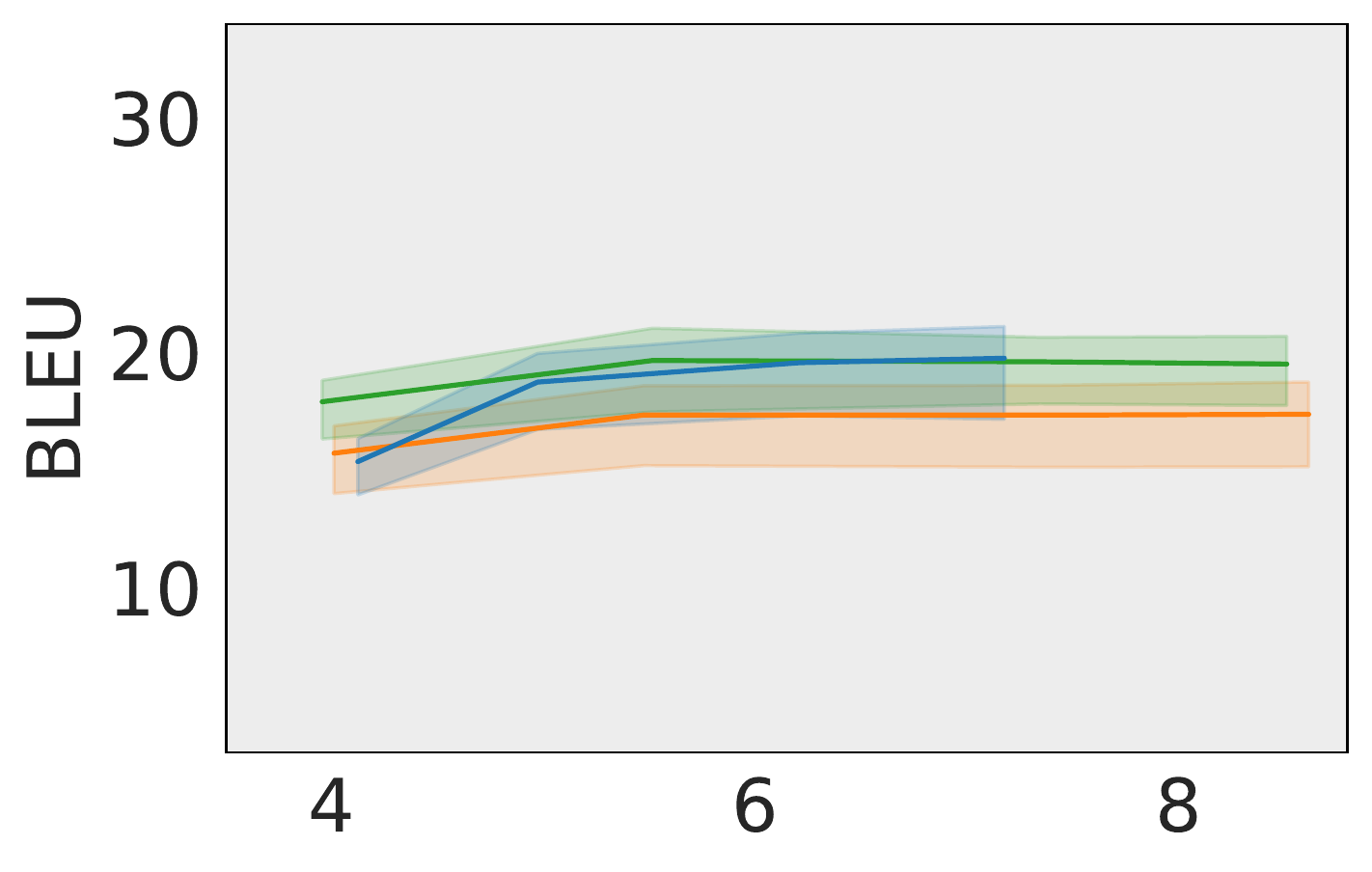} & 
\includegraphics[width=\plotwidth,valign=m]{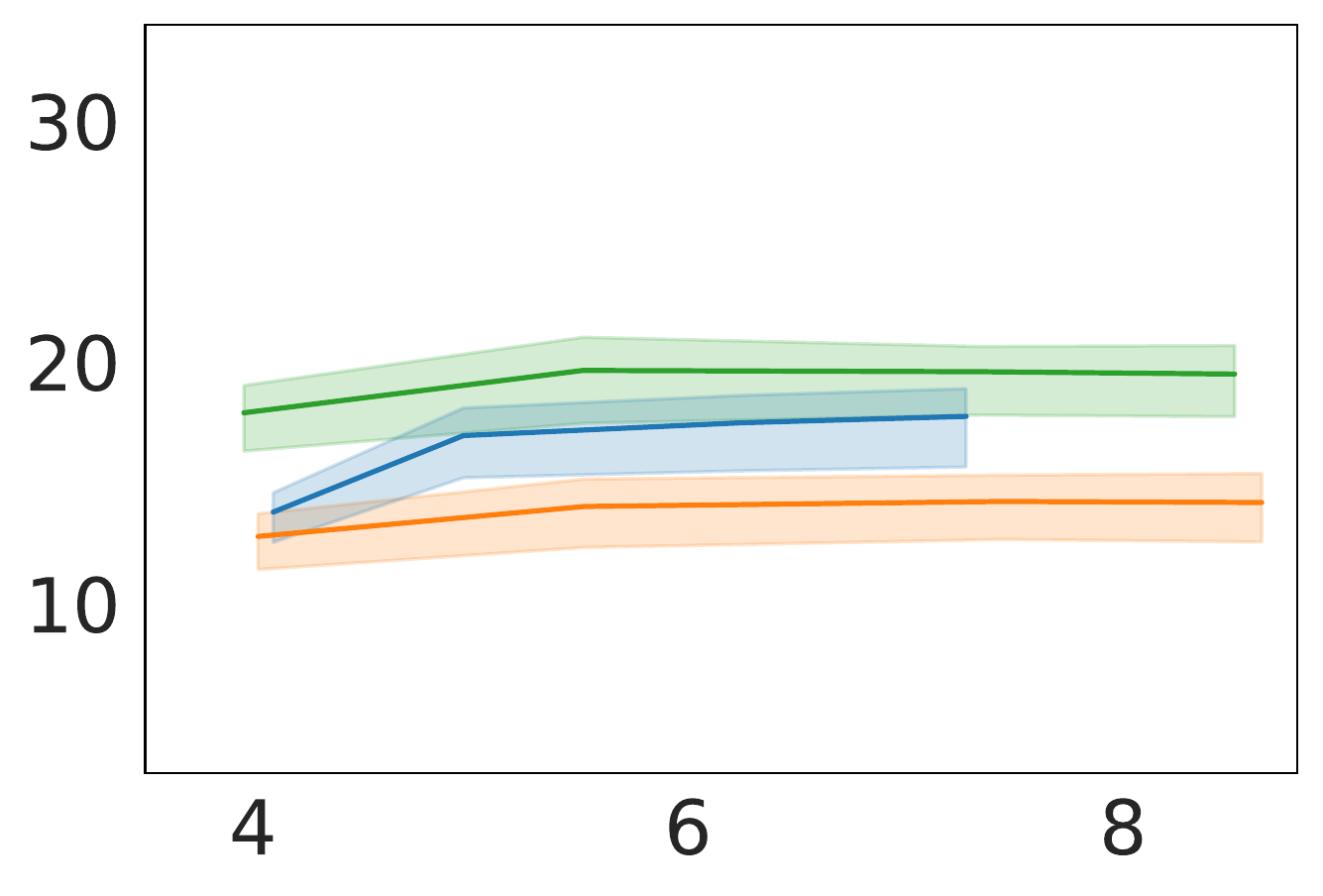} & 
\includegraphics[width=\plotwidth,valign=m]{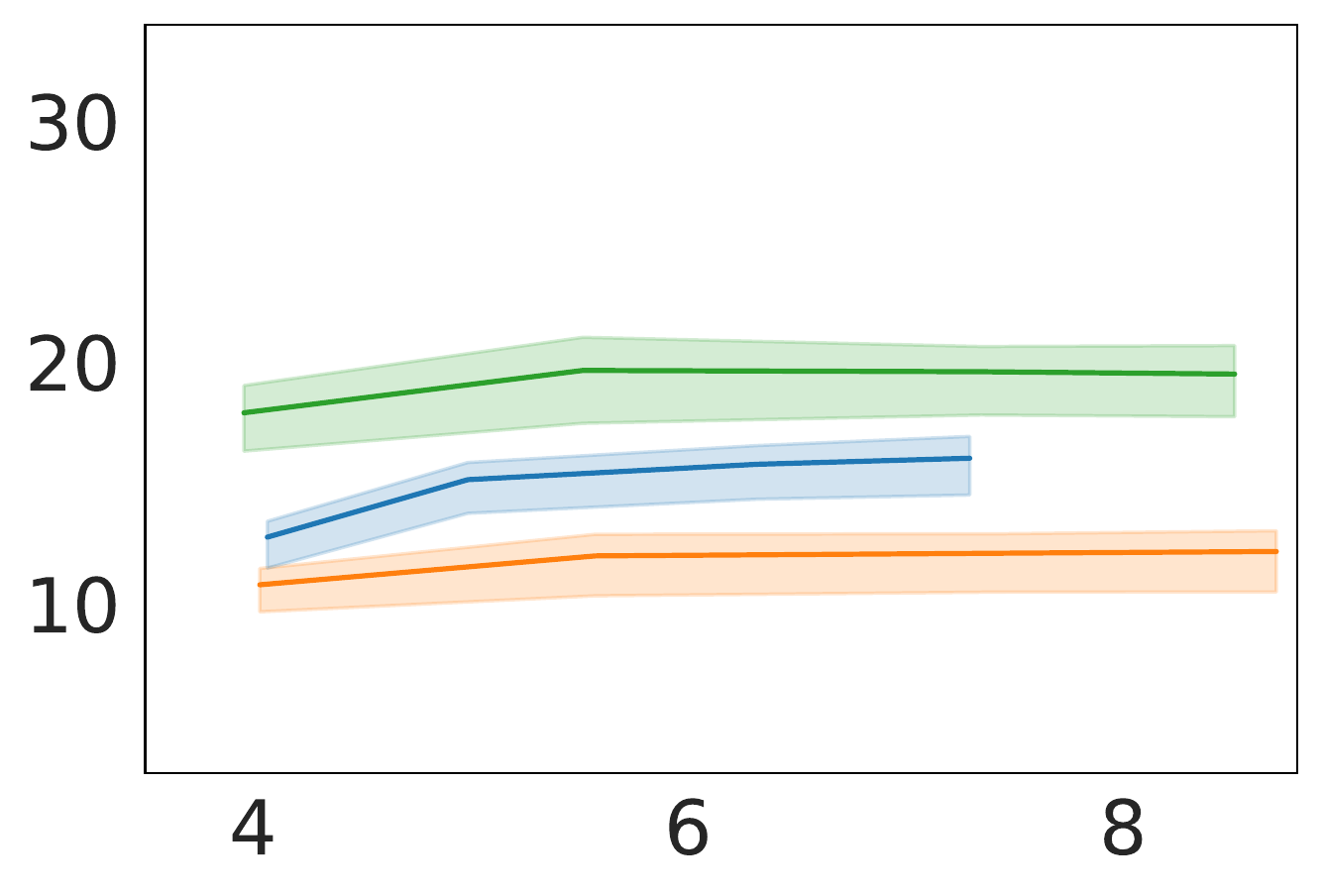} &
\includegraphics[width=\plotwidth,valign=m]{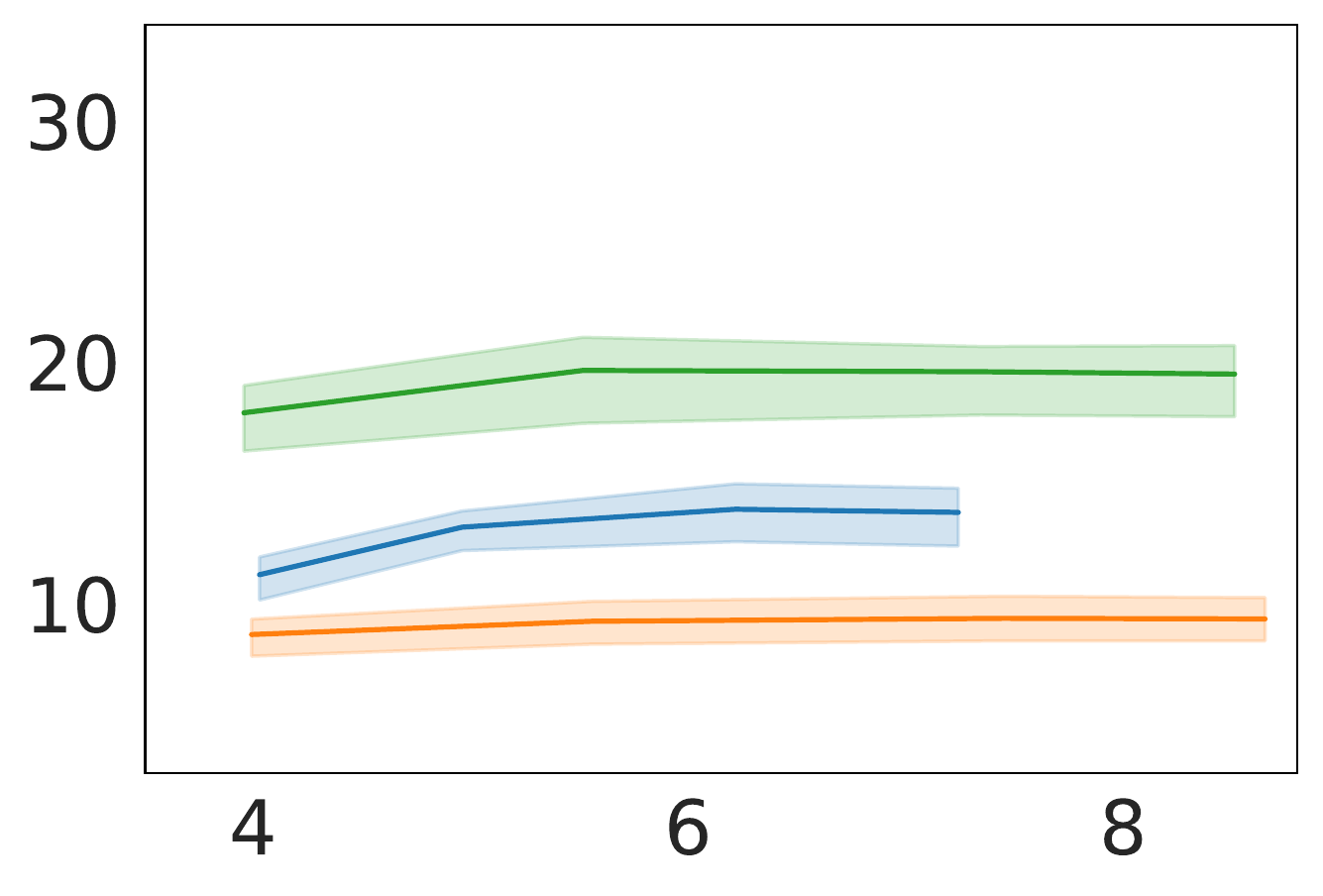} \\ 
                                                                               
30\% &                                                                         
\includegraphics[width=\plotwidth,valign=m]{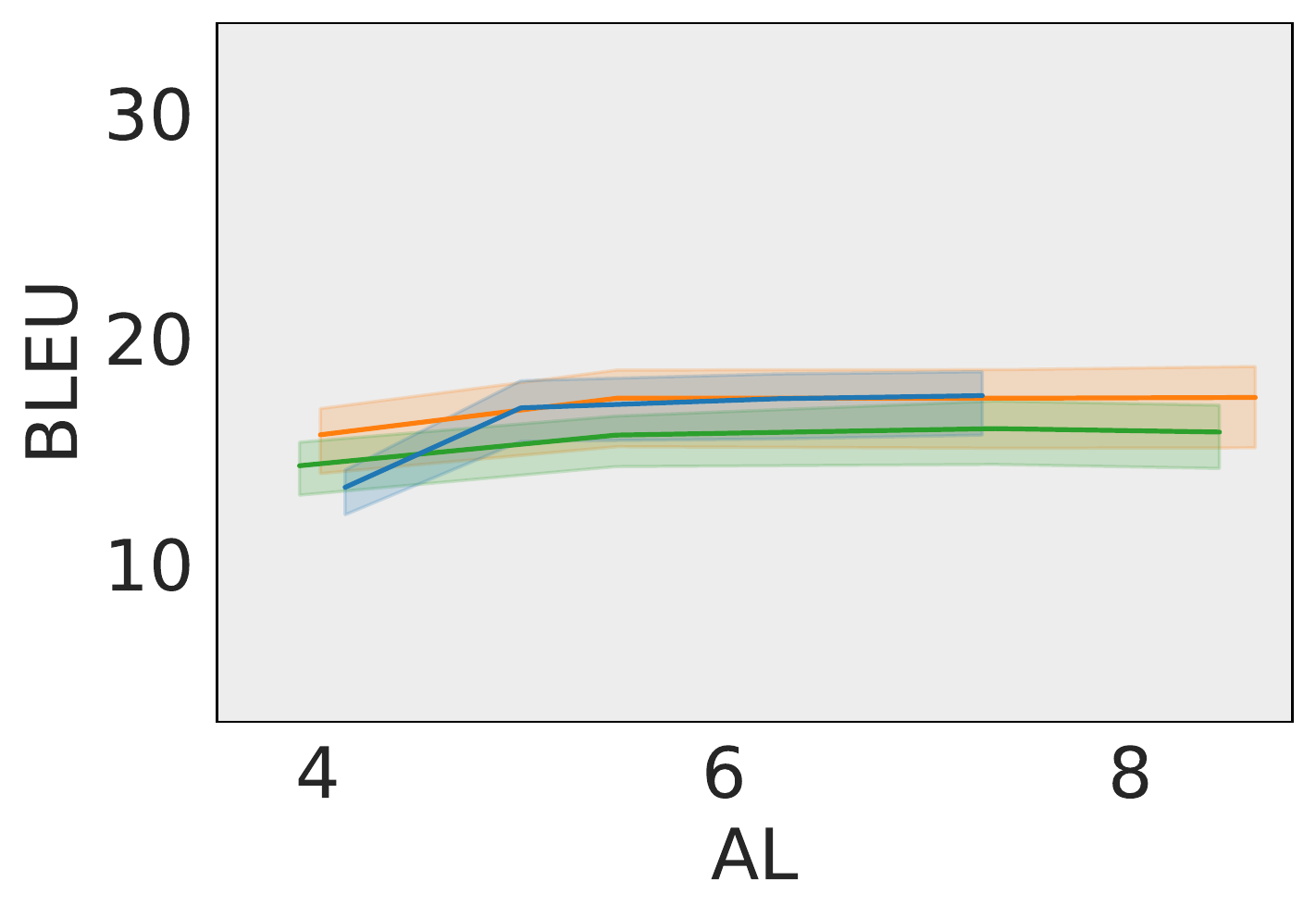} & 
\includegraphics[width=\plotwidth,valign=m]{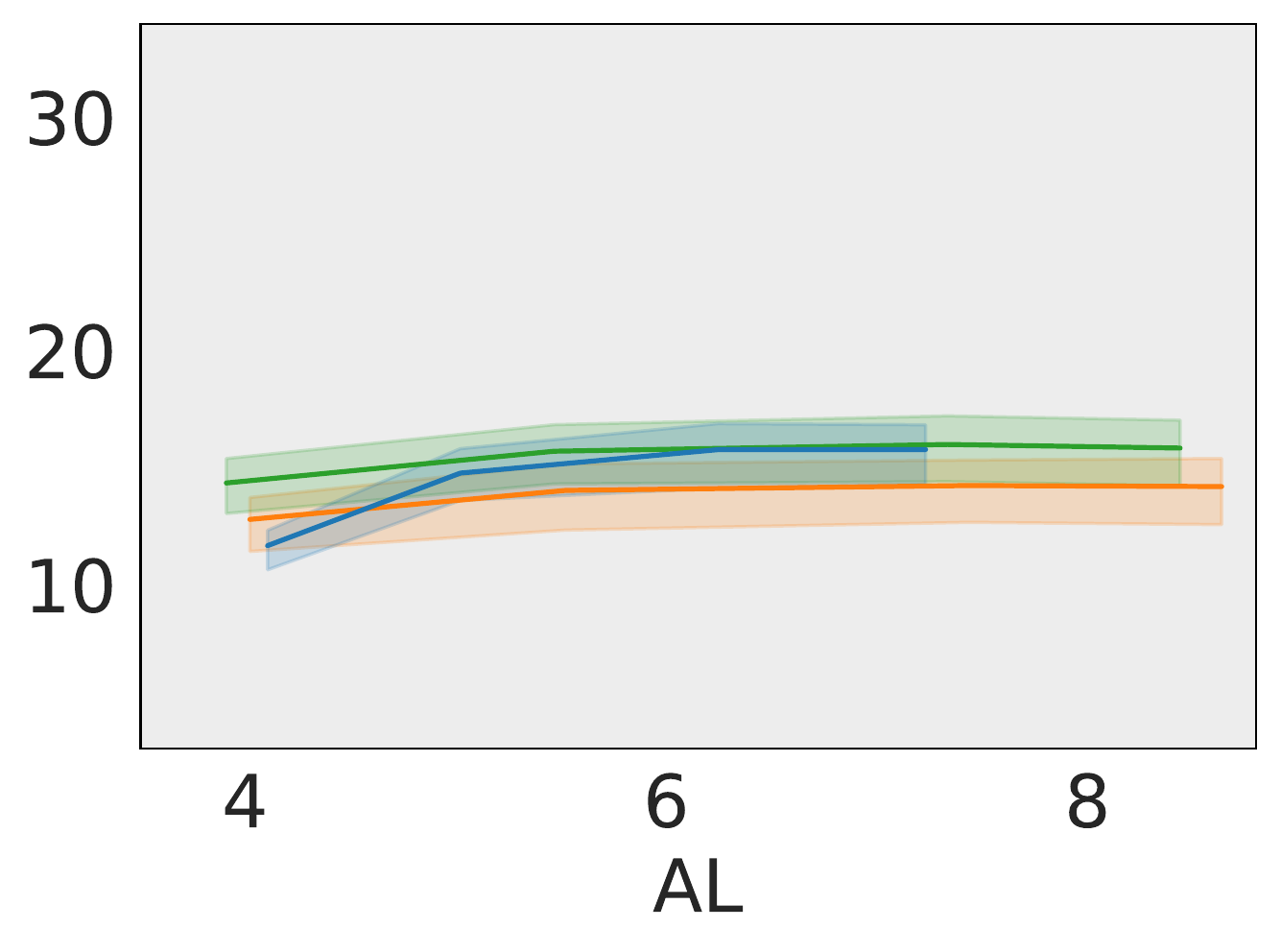} & 
\includegraphics[width=\plotwidth,valign=m]{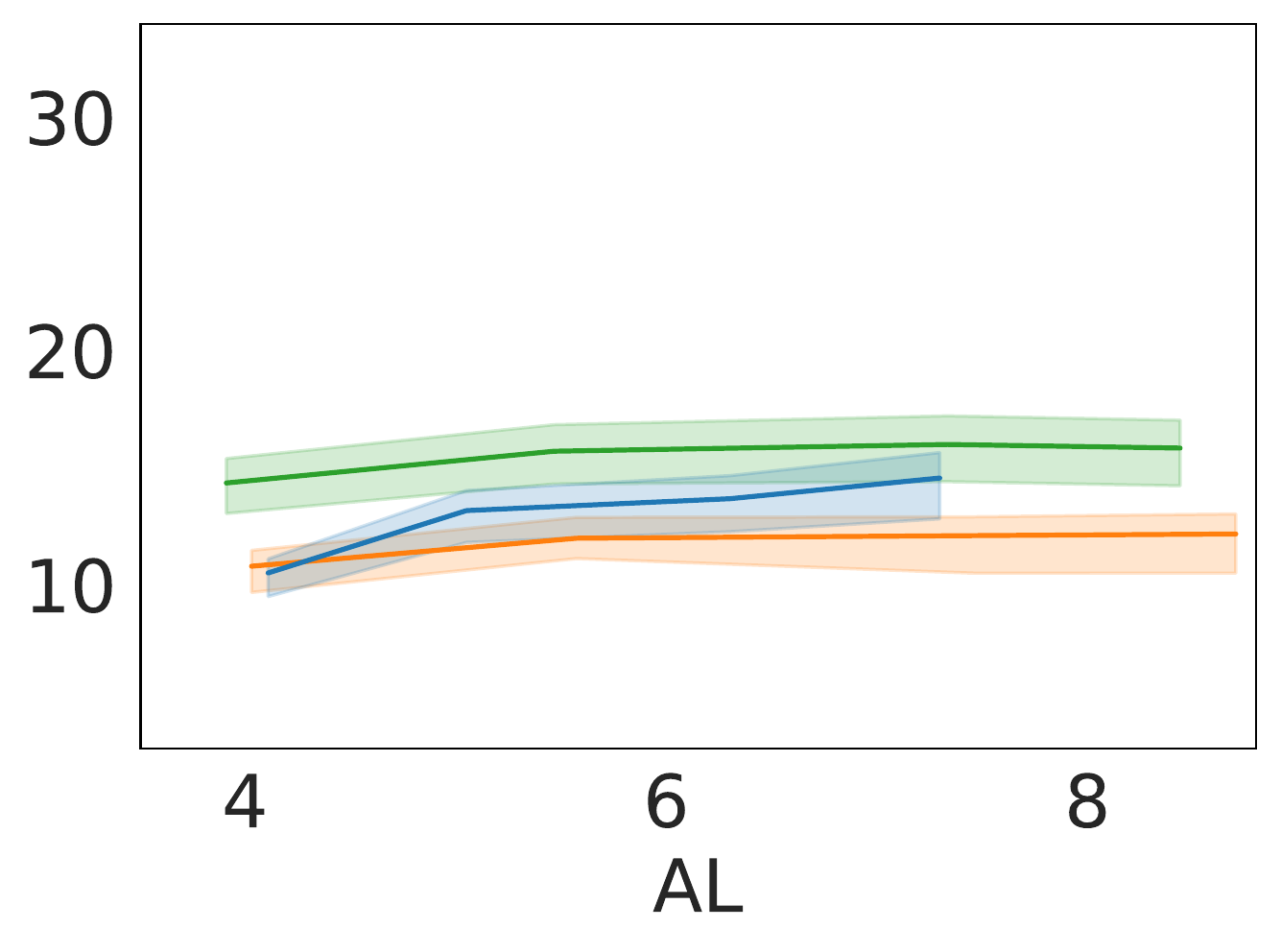} &
\includegraphics[width=\plotwidth,valign=m]{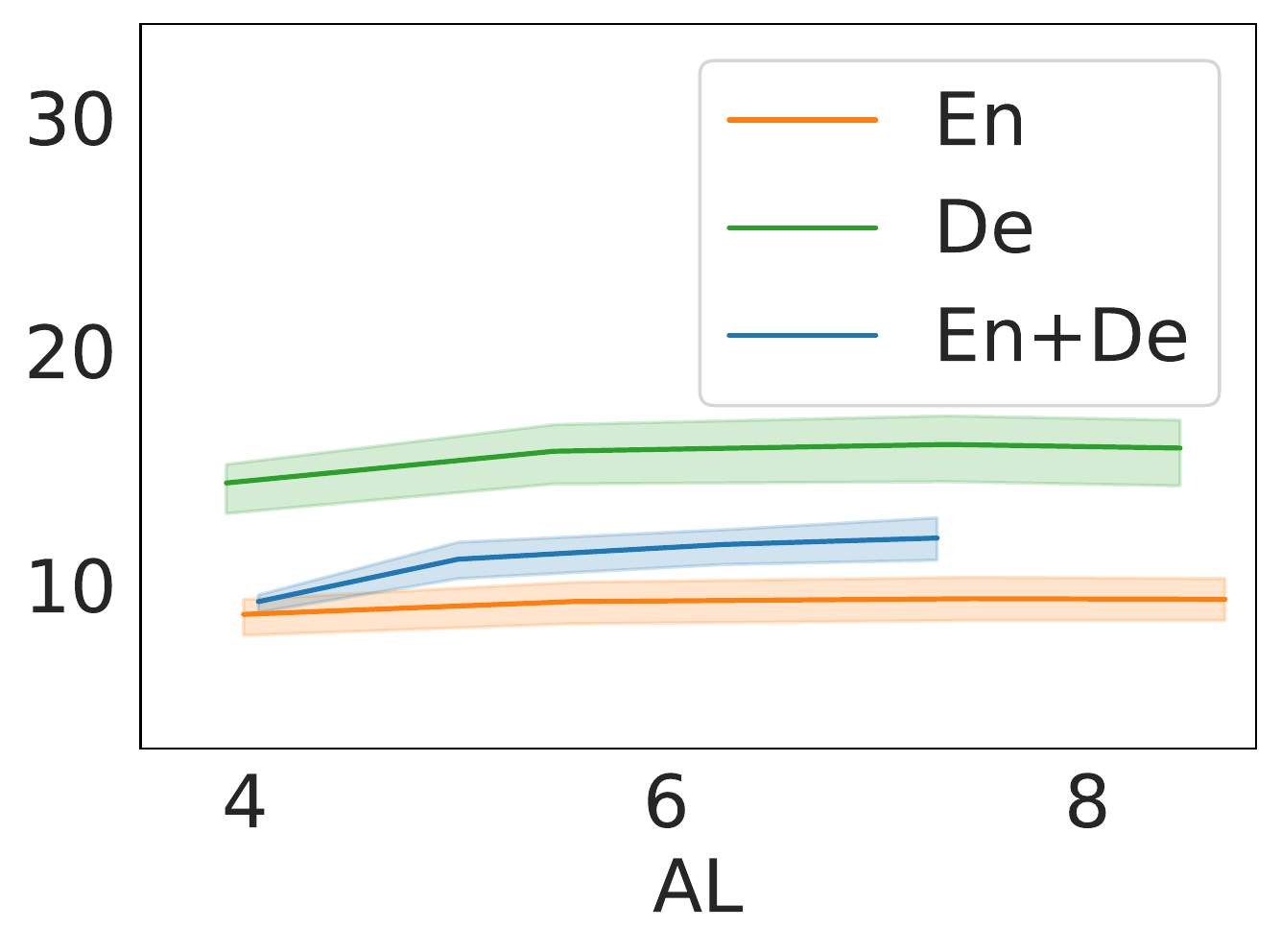} \\
\end{tabular}